\documentclass[final,3p,times]{elsarticle}
\usepackage{amssymb}

\usepackage{graphicx}
\usepackage{comment}
\usepackage{algorithm}
\usepackage[noend]{algpseudocode}
\usepackage{algorithmicx}
\usepackage{multirow}
\usepackage{soul}
\usepackage{color,xcolor}
\usepackage{xpatch}
\usepackage{subfigure} 

\usepackage{amsmath,amssymb,amsfonts}
\usepackage{textcomp}
\usepackage{makecell}
\usepackage{multirow}
\usepackage{booktabs}
\usepackage{tabularx}
\usepackage{float}
\usepackage{overpic}
\usepackage[section]{placeins}

\usepackage{xcolor}
\usepackage{xpatch}

\makeatletter
\def\changeBibColor#1{%
	\in@{#1}{zhou2018unet++,azad2024medical,siddique2021u} 
	\ifin@\color{red}\else\normalcolor\fi
}

\xpatchcmd\@bibitem
  {\item}
  {\changeBibColor{#1}\item}
  {}{\fail}
 
\xpatchcmd\@lbibitem
  {\item}
  {\changeBibColor{#2}\item}
  {}{\fail}
\makeatother

\begin{document}

\begin{frontmatter}


\title{An Unsupervised Framework for Dynamic Health Indicator Construction and Its Application in Rolling Bearing Prognostics}



\author[mymainaddress]{Tongda Sun}
\author[mymainaddress]{Chen Yin}
\author[mysecaddress]{Huailiang Zheng}
\author[mymainaddress]{Yining Dong\corref{mycorrespondingauthor}}
\ead{yinidong@cityu.edu.hk}
\cortext[mycorrespondingauthor]{Corresponding author}
\address[mymainaddress]{School of Data Science and Hong Kong Institute for Data Science, City University of Hong Kong, Hong Kong}
\address[mysecaddress]{College of Mechanical and Electrical Engineering, Harbin Engineering University, Harbin 150001, China}

\soulregister{\cite}7 
\soulregister{\citep}7 
\soulregister{\citet}7 
\soulregister{\ref}7 
\soulregister{\pageref}7 

\sethlcolor{yellow}
\begin{abstract}

Health indicator (HI) plays a key role in degradation assessment and prognostics of rolling bearings. Although various HI construction methods have been investigated, most of them rely on expert knowledge for feature extraction and overlook capturing dynamic information hidden in sequential degradation processes, which limits the ability of the constructed HI for degradation trend representation and prognostics. To address these concerns, a novel dynamic HI that considers HI-level temporal dependence is constructed through an unsupervised framework. Specifically, a degradation feature learning module composed of a skip-connection-based autoencoder first maps raw signals to a representative degradation feature space (DFS) to automatically extract essential degradation features without the need for expert knowledge. Subsequently, in this DFS, a new HI-generating module embedded with an inner HI-prediction block is proposed for dynamic HI construction, where the temporal dependence between past and current HI states is guaranteed and modeled explicitly. On this basis, the dynamic HI captures the inherent dynamic contents of the degradation process, ensuring its effectiveness for degradation tendency modeling and future degradation prognostics. The experiment results on two bearing lifecycle datasets demonstrate that the proposed HI construction method outperforms comparison methods, and the constructed dynamic HI is superior for prognostic tasks.

\end{abstract}

\begin{keyword}
{Dynamic health indicator; Skip-connection-based autoencoder; Degradation prognostics; Unsupervised learning.}
\end{keyword}

\end{frontmatter}

\section{Introduction}
\label{sec:introduction}

As one of the most important rotating components, rolling bearings are widely used in various industrial scenarios. However, the harsh and complex operating conditions can lead to unexpected failures, resulting in both accidental breakdowns and economic losses \cite{zhou2022construction}. 
\textcolor{black}{Predictive maintenance approaches are essential to address this issue, as they not only ensure the reliability of rolling bearings but also enhance safety and productivity in industrial systems \cite{zio2022prognostics, yin2025physics}.} A common strategy involves constructing a health indicator (HI) for degradation representation, prediction, and remaining useful life (RUL) estimation \cite{ni2022data}. Therefore, a well-constructed HI can substantially improve the performance of degradation prognostics \cite{wang2017prognostics}.

HIs can be classified into two categories based on their construction methodologies: Physics HIs (PHIs) and Virtual HIs (VHIs) \cite{lei2018machinery}. PHIs are typically derived from monitoring signals using statistical or signal processing techniques. In this context, various methods such as the root mean square (RMS) \cite{meng2021health}, kurtosis \cite{li2015improved}, sparsity measures\cite{hou2021investigations}, and entropy-based methods \cite{kumar2022state, aremu2019relative} have been extensively investigated. Although these PHIs have specific physical interpretations, several limitations hinder their application. First, PHIs are generally designed for specific tasks, which heavily depend on expert knowledge, resulting in poor generalization ability. Second, PHIs struggle to provide a comprehensive understanding of the intricate degradation processes in industrial settings. 
For instance, while RMS is capable of illustrating the monotonic degradation trend, it lacks sensitivity towards incipient faults. Conversely, kurtosis demonstrates a higher sensitivity to incipient faults, but it does not effectively reflect the degradation trend.

VHIs are mainly constructed by fusing multiple PHIs to contain more degradation information \cite{lei2018machinery}. 
Traditional dimension reduction approaches, such as dynamic principal component analysis (DPCA) \cite{buchaiah2022bearing} and independent component analysis (ICA) \cite{yang2022novel}, as well as advanced feature fusion methods, have been explored for this purpose.
For example, Wu et al. \cite{wu2018degradation} preprocessed vibration signals into several statistical, intrinsic energy, and fault frequency features and then fused them into an HI using dynamic principal component analysis (DPCA). 
Li et al. \cite{li2022remaining} constructed an HI by calculating the Mahalanobis distance of time and time-frequency domain degradation features. Li et al. \cite{li2023feature} selected monotonicity-sensitive features from time and frequency features and formulated these features into an HI by calculating the feature degradation ratio.

Instead of directly analyzing and fusing PHIs, unsupervised deep learning (DL) approaches extract higher level and more complex features for comprehensive VHI construction. Among them, autoencoder and its variants are most commonly adopted as the backbone for feature learning \cite{xu2022constructing, qin2023new,wang2024deep,mao2024swdae}. For example,
Ma et al. \cite{ma2022health} applied a self-attention convolutional autoencoder to derive low-dimensional features. These features were then fused into an HI based on the weighted similarity between baseline and monitoring data features. Guo et al. \cite{guo2022unsupervised} used the relative similarity of features learned by a multiscale convolutional autoencoder network to build HI. 
In several cases, the results of the autoencoder are directly used as the HI. 
In Ref. \cite{gonzalez2022health}, the latent reconstruction error given by the deep autoencoder and variational autoencoder (VAE) is used as the HI. Xu et al. \cite{xu2023novel} proposed an enhanced stacked autoencoder based on an exponent weight moving average to construct HI from raw signals. Chen et al. \cite{chen2021health} proposed a novel quadratic function-based deep convolutional autoencoder to extract HI. Additionally, Qin et al. \cite{qin2021unsupervised} extracted several time, frequency, and time-frequency features and then used a VAE to learn a latent variable as HI from them. To sum up, the HI plays a key role in machinery degradation prognostics. The quality of HI can directly affect the performance of degradation assessment and prediction.

\begin{figure*}[!t]
\centerline{\includegraphics[width=0.6\textwidth]{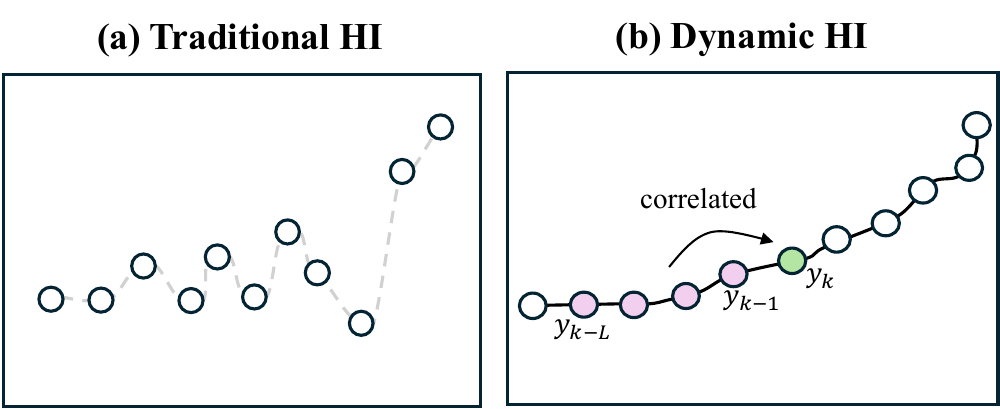}}
\caption{The comparison of traditional HI and the proposed dynamic HI. \textcolor{black}{(a) Traditional HI fails to ensure the temporal dependence between current and past HI states, which may lead to inaccurate degradation modeling and unreliable prognostics.} (b) The dynamic HI captures and guarantees the HI-level temporal dependence, where current HI is explicitly correlated with its past values, enhancing the representability and predictability of the HI. This in turn improves the prognostic performance of the HI.}
\label{dynamic HI compar}
\end{figure*}

Although the abovementioned methods have shown promising results for HI construction, there are still several challenges that need to be addressed. On the one hand, most construction methods for PHI and VHI fused by multiple PHIs rely on expert knowledge for feature engineering, which is difficult to obtain in practical scenarios and introduces human bias. 
\textcolor{black}{On the other hand, although some VHI methods based on DL can construct HI from raw signals, they often approach this task from a static perspective. These methods derive HI based solely on current observations but ignore the influence of previous degradation states. As a result, they fail to capture the dynamic nature of system degradation, which ultimately constrains the effectiveness of HIs in prognostic applications.}
\textcolor{black}{Several methods attempted to construct HIs with improved prognostic ability by exploring the temporal information within the progressive degradation process. For instance, Li et al. \cite{li2023canonical} proposed a canonical correlation analysis (CCA) based HI construction method, where the degradation features are learned by an autoencoder and the temporal dependence between the health and faulty and health state features is calculated by CCA as the HI. Cheng et al. \cite{cheng2021convolutional} extracted intrinsic energy features of degradation through complete ensemble empirical mode decomposition with adaptive noise (CEEMDAN) and then constructed HI by analyzing the temporal dynamics of these features through dynamic principal component analysis (DPCA). Chen et al. \cite{chen2022health} developed an autoencoder for degradation features learning, where the long short-term memory (LSTM) layers are integrated to capture temporal information of the input samples. Although these methods capture the temporal information of the degradation process to facilitate the HI construction, they only focus on modeling the sample-level or feature-level temporal dependence but ignore the HI-level temporal dependence. Considering that degradation is a progressive process, as shown in Fig. \ref{dynamic HI compar}, the HI-level temporal dependence is crucial for constructing a HI with strong representability and predictability.}
Overlooking HI-level temporal dependence limits the characteristic ability of the constructed HI for degradation modeling, which may consequently result in unreliable prognostics.

\textcolor{black}{To tackle these challenges, an unsupervised framework is proposed for dynamic HI construction. By explicitly considering the HI-level temporal dependence, dynamic HI achieves superior representability and predictability compared to traditional HIs. Consequently, its performance in bearing degradation tendency characterization and prognostics is enhanced.} The framework consists of two sequential modules. First, a degradation feature learning module is designed to map raw signals into a low-dimensional degradation feature space, where a skip-connection-based autoencoder (SkipAE) is proposed as the backbone to extract comprehensive features by considering multi-level and multi-scale information. \textcolor{black}{Then, a dynamic HI-generating module integrated with an inner HI-prediction block is developed to construct HI from the learned degradation features. The inner HI-prediction model explicitly models and guarantees the HI-level temporal dependence, aligning the HI more closely with the true degradation process. Moreover, to ensure the monotonic tendency of the HI, a monotonic loss is further introduced into the dynamic HI-generating module as a constraint term.} The main contributions of this work are summarized as follows.

\begin{enumerate}
\item A degradation feature learning module composed of a SkipAE is proposed to learn more meaningful degradation features from raw signals, getting rid of the need for manual feature engineering and preserving multi-level diagnostic information in features.

\item A novel HI-generating module integrated with an inner HI-prediction block is proposed to construct dynamic HI. Compared to traditional HI, dynamic HI explicitly guarantees the HI-level temporal dependence, enhancing its ability for degradation tendency characterization and prognostics.


\item \textcolor{black}{An unsupervised framework is proposed for automatic dynamic HI construction to ensure its representability and predictability of the degradation process, while extensive experiments on two benchmarks are conducted to verify its superior performance in rolling bearing prognostics.}

\end{enumerate}


The remainder of this article is organized as follows. The procedure and theoretical background of the proposed method are introduced in section \ref{sec:Methods}. Section \ref{sec:Applications} presents and discusses the results of comparative experiments and ablation studies on two experiments. Finally, section \ref{sec:Conclusions} concludes this paper.

\section{Methodology}
\label{sec:Methods}
\subsection{\textcolor{black}{Framework for Dynamic HI Construction}}
\begin{figure*}[!t]
\centerline{\includegraphics[width=0.8\textwidth]{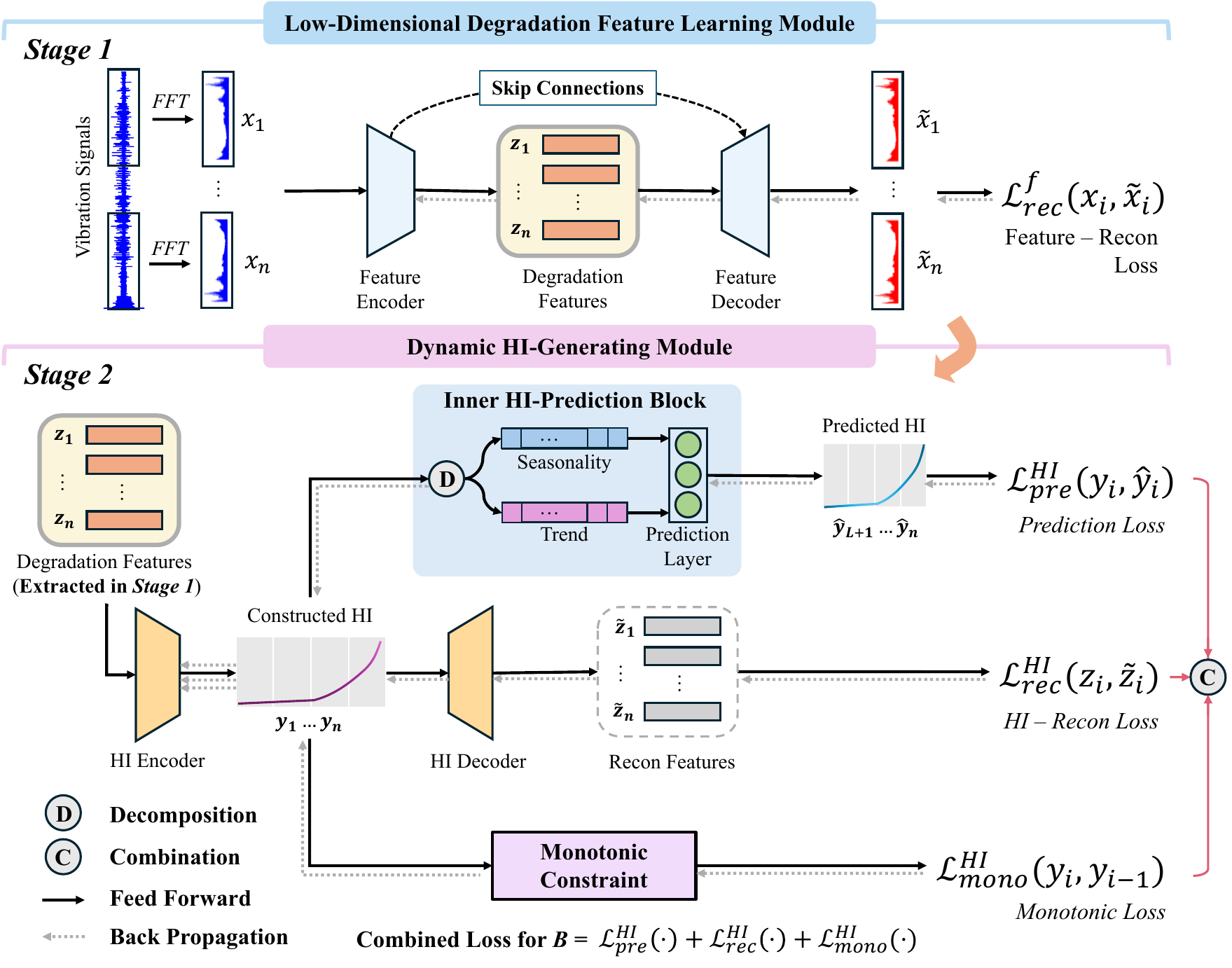}}
\caption{\textcolor{black}{Framework for dynamic HI construction. It consists of two stages. Stage 1: Extracting low-dimensional degradation features through the degradation feature learning module. Stage 2: Constructing dynamic HI from the degradation features extracted in Stage 1 through the dynamic HI-generating module, where HI-level temporal dependence is guaranteed by the integrated inner HI-prediction block.}}
\label{fig1}
\end{figure*}

\textcolor{black}{
In this paper, an unsupervised framework is proposed for dynamic HI construction. As shown in Fig. \ref{fig1}, the framework mainly consists of two stages. In stage 1, to deal with the background noise and the curse of dimensionality, the low-dimensional degradation feature learning is conducted through a skip-connection-based autoencoder (SkipAE).
Through skip connections, this autoencoder maps run-to-failure degradation data $\{x_k\}_{1}^{n},x_k\in\mathbb{R}^{m}$ into a representative degradation feature space with multi-scale and multi-level information, ensuring the automatic learning of critical degradation features. Subsequently, in stage 2, to guarantee the HI-level temporal dependence, which is important for degradation prognostics but often overlooked by existing methods, a dynamic HI-generating module is proposed for dynamic HI construction. Specifically, the degradation features $\{z_k\}_{1}^{n},z_k\in\mathbb{R}^{l}$ extracted in stage 1 are further compressed into an HI series $\{y_k\}_{1}^{n},y_k\in\mathbb{R}^{1}$ by the HI encoder. In the meanwhile, as high predictability typically indicates strong temporal dependence, an inner HI-prediction block is integrated to explicitly model and guarantee the HI-level temporal dependence by minimizing the prediction loss between $y_k$ and $\hat{y}_k$, which is forecasted from the past states $\{y_i\}_{i=k-L}^{k-1}$. Furthermore, a monotonic loss is incorporated to align the dynamic HI with the natural degradation trend. The HI-level temporal dependence and the monotonicity ensure that the constructed dynamic HI effectively captures the dynamic information of the degradation process, enhancing its ability for degradation trend characterization and prognostics.
}


\subsection{Degradation Features Learning Module based on a Skip-connection-based Autoencoder}

\textcolor{black}{Due to the background noise and the curse of dimensionality, it is difficult to directly construct HI from raw signals. Therefore, the representative and degradation-sensitive features should be extracted first for HI construction. However, current HI construction methods typically use traditional autoencoders for feature learning, which can result in information loss during reconstruction and limit the feature representativeness. To address this issue, a SkipAE is proposed in this section. }

\begin{figure}[!t]
\centerline{\includegraphics[width=0.7\columnwidth]{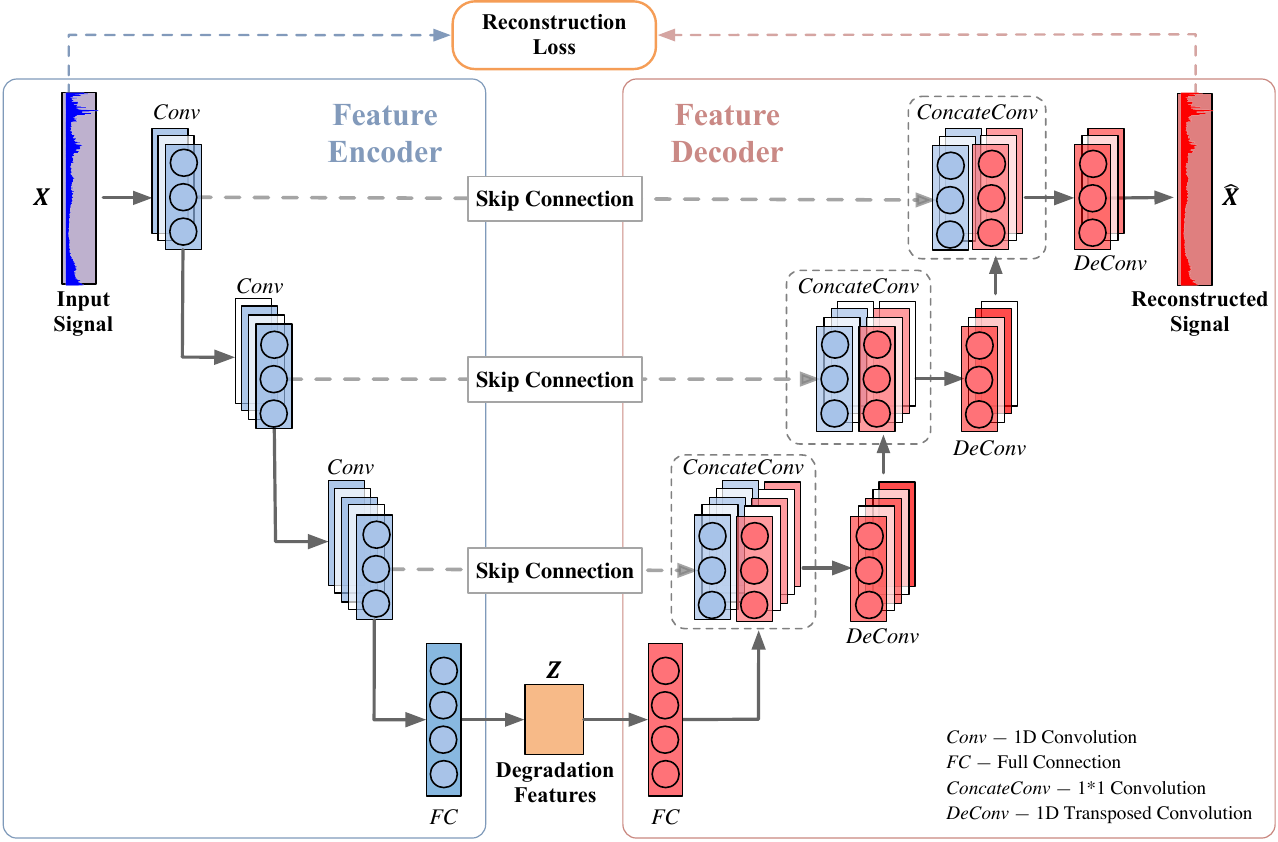}}
\caption{\textcolor{black}{Structure of the skip-connection based autoencoder.}}
\label{fig2}
\end{figure}

\textcolor{black}{
Fig. \ref{fig2} shows the basic structure of SkipAE. In this structure, various layers of the encoder capture feature maps at different levels and scales. Skip connections directly link the output of these layers to their corresponding layers in the decoder, preserving multi-level and multi-scale information in the learned critical degradation features. This approach reduces the useful information loss during the reconstruction process and enhances the representativeness of the learned features. Additionally, skip connections provide direct pathways for gradients to flow back to earlier layers, mitigating the vanishing gradient problem and maintaining a stable and efficient learning process \cite{li2018visualizing}. 
}

\textcolor{black}{Specifically, as illustrated in Fig. \ref{fig2}, each decoder layer receives two feature maps: one from the corresponding encoder layer and one from the preceding decoder layer. These feature maps are then sequentially processed through a concatenation operation followed by a deconvolution operation. Let $h_j$ denote the output of the $j$-th layer of the decoder path and let $f_{en}^j$ denote the output of the $j$-th layer of the encoder path, the operation of the skip connection is defined as
\begin{equation}
h_j = DeConv(ConcatConv(h_{j-1}\oplus f_{en}^j)),\label{eq2}
\end{equation}
where $\oplus$ represents the channel-wise concatenation, and $ConcatConv(\cdot)$ denotes the unit convolution to integrate the concatenated feature maps while retaining the spatial dimensions. $DeConv(\cdot)$ is a one-dimensional transposed convolution operation.}


\textcolor{black}{Let $x_i$ denote the input signal. The reconstruction process is defined as
\begin{equation}
\begin{aligned}
z_i=f_{en}(x_i), \\
\hat{x}_i=f_{de}(z_i),\label{eq1}
\end{aligned}
\end{equation}
where $f_{en}(\cdot)$ and $f_{de}(\cdot)$ represent the encoding and decoding operation, respectively. $\hat{x}_i$ indicates the reconstructed signal. $z_i$ denotes the extracted degradation features, which are the output of the feature encoder and will be used in stage 2 for subsequent dynamic HI construction.}

The loss function, which aims to minimize the difference between the input data $x_i$ and output data $\hat{x}_i$, is formulated as
\begin{equation}
    L_{rec}=\frac{1}{n}\sum_{i=1}^n \left \| x_i - \hat{x}_i \right \|^2,\label{eq3}
\end{equation}
and the optimal values of the parameters of encoder $\theta_{en}$ and decoder $\theta_{de}$ can be solved as
\begin{equation}
  {\theta}_{en}^*, {\theta}_{de}^* = \mathop{\arg\min}\limits_{\theta_{en}, \theta_{de}} L_{rec}(\theta_{en}, \theta_{de}).\label{eq4}
\end{equation}
In practice, the stochastic gradient descent algorithm is applied to update the parameters as
\begin{equation}
    \theta_{en} \leftarrow \theta_{en} -  \alpha \frac{\partial L_{rec}}{\partial \theta_{en}};
    \theta_{de} \leftarrow \theta_{de} -  \alpha \frac{\partial L_{rec}}{\partial \theta_{de}},\label{eq5}
\end{equation}
where $\alpha$ denotes the learning rate.

\subsection{Dynamic HI-generating module integrated with an inner HI-prediction block}
\begin{figure}[!t]
\centerline{\includegraphics[width=0.5\columnwidth]{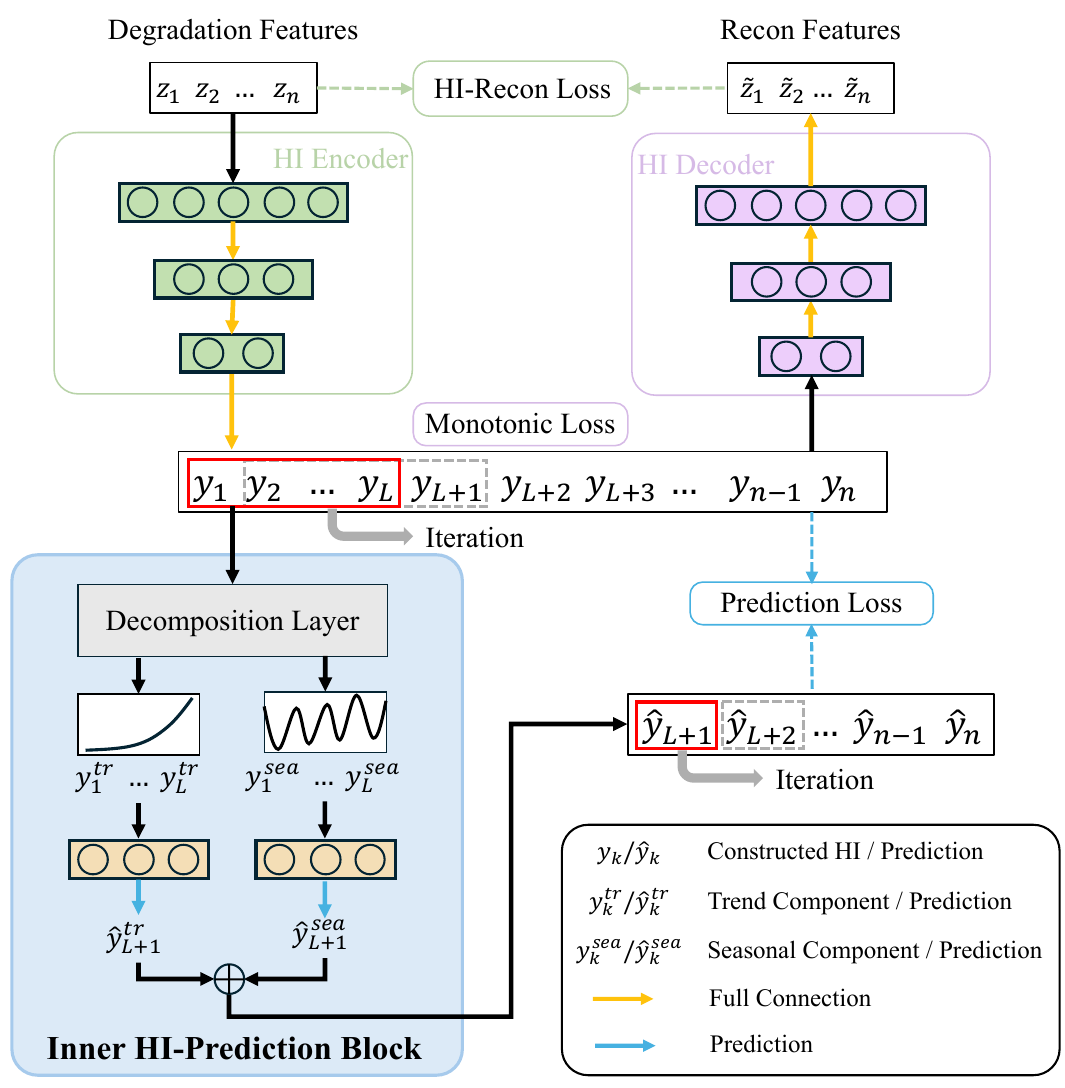}}
\caption{\textcolor{black}{Schematic diagram of the dynamic HI-generating module integrated with an inner HI-prediction block.}}
\label{fig3}
\end{figure}

HI is designed to quantitively characterize the degradation trend of machinery. \textcolor{black}{Given the progressive nature of the degradation process, HI-level temporal dependence is crucial for ensuring the effectiveness of HI in degradation prognostics. However, it is often overlooked by current HI construction methods, which may limit the characteristic ability of the constructed HI for degradation modeling and consequently result in unreliable prognostics.}

\textcolor{black}{To address this issue, a novel dynamic HI-generating module integrated with an inner HI-prediction block is proposed for dynamic HI construction. This approach explicitly guarantees the HI-level temporal dependence, aligning the HI more closely with the bearing degradation process. Fig. \ref{fig3} presents the schematic diagram of this module. The degradation features learned by the SkipAE are mapped into the HI series by an HI encoder. In the meanwhile, as high predictability typically indicates strong temporal dependence, the integrated inner HI-prediction block explicitly models and guarantees the HI-level temporal dependence, allowing the HI to retain as much dynamic information as possible.}

Specifically, the degradation features learned by SkipAE are mapped into the HI series by an HI encoder. Let $\{z_i\}_{1}^{n}$ denote the degradation features. The HI $y_i$ is constructed as
\begin{equation}
    y_i=f_{en}^{HI}(z_i),\label{eq6}
\end{equation}
where $f_{en}^{HI}(\cdot)$ denotes the computation of the HI encoder and $y_i\in\mathbb{R}$. Considering the prognostics may face the problem of data scarcity, this operation is conducted by a lightweight full connection autoencoder. The HI reconstruction loss can be calculated as
\begin{equation}
    L_{rec}^{HI}=\frac{1}{n}\sum_{i=1}^{n} \left \| z_i - \hat{z}_i \right \|^2,\label{eq7}
\end{equation}
where $\{\hat{z}_i\}_{1}^{n}$ denote the reconconstructed degradation features.

\textcolor{black}{As high predictability typically indicates strong temporal dependence, the inner HI-prediction block is integrated to model the HI-level temporal dependence by measuring the difference between $y_k$ and its predicted value $\hat{y}_k$. This block includes a decomposition layer and two full connection prediction layers. 
Specifically, for a specific $y_k$ at time $k$, its value can be forecasted from the past L values $\Gamma_k = \{y_i\}_{i=k-L}^{k-1}$, $\Gamma_k\in\mathbb{R}^{L\times{1}}$.}
To enhance the prediction performance, the input $\Gamma_k$ is first decomposed into trend and seasonal components as
\begin{equation}
\begin{aligned}
    &\Gamma_k^{tr} = AvgPool(\Gamma_k)\\
    &\Gamma_k^{sea} = \Gamma_k-\Gamma_k^{tr},\label{eq9}
\end{aligned}
\end{equation}
where $AvgPool(\cdot)$ denotes the moving average operation, and $\Gamma_k^{tr}, \Gamma_k^{sea}\in\mathbb{R}^{L\times1}$ denote the trend and seasonal components, respectively.

\textcolor{black}{Subsequently, the trend and seasonal components are predicted separately by two prediction layers, and their prediction results are combined to obtain the predicted HI $\hat{y}_k$ as}
\begin{equation}
\begin{aligned}
    &\hat{y}_k^{tr} = W_{tr}\Gamma_k^{tr}\\
    &\hat{y}_k^{sea} = W_{sea}\Gamma_k^{sea}\\
    &\hat{y}_k = \hat{y}_k^{tr} + \hat{y}_k^{sea},\label{eq10}
\end{aligned}
\end{equation}
where $W_{tr}, W_{sea}\in\mathbb{R}^{1\times{L}}$ denote the parameters of prediction layers for trend and seasonal components, respectively. The prediction loss is calculated as
\begin{equation}
    L_{pre}^{HI}=\frac{1}{n-L}\sum_{k=L+1}^{n} \left \| y_k - \hat{y}_k \right \|^2.\label{eq11}
\end{equation}
\textcolor{black}{This prediction loss measures the difference between $y_k$ and $\hat{y}_k$, which reflects the HI-level temporal dependence. By embedding this loss into the HI construction process, the HI can be optimized to maintain a strong dependence with its previous states, thereby reflecting the dynamic nature of the degradation process. Consequently, the constructed dynamic HI achieves superior representability and predictability for rolling bearing prognostics.}

To further ensure the degradation trend of the constructed HI, a monotonic loss is introduced according to the research of Qin et al. \cite{qin2021unsupervised}, and it can be represented as
\begin{equation}
    L_{mono}^{HI}=\frac{1}{n}\sum_{i=1}^n \left \| y_i - y_{i-1} - r \right \|^2,\label{eq8}
\end{equation}
where $r$ is a relatively large constant to encourage $y_i > y_{i-1}$. In this paper, $r$ is set as 10 via a batch of experiments.

The total loss is formulated as a weighted combination of reconstruction loss, prediction loss, and monotonic loss as
\begin{equation}
    L_{combined} = L_{rec}^{HI} + \lambda L_{pre}^{HI} + \gamma L_{mono}^{HI},\label{eq12}
\end{equation}
where $\lambda$ and $\gamma$ are two coefficients controlling the importance of the representativeness, predictability, and monotonicity of the constructed HI. In the case studies, $\lambda$ is set as 0.1, and $\gamma$ is set as 1 via numerous experiments.

\textcolor{black}{The optimization is conducted by the stochastic gradient
descent algorithm. During the training, the parameters of the HI encoder, HI decoder, and the inner HI-prediction block are updated in each epoch. The HI series, which is used as the target to formulate the prediction loss, is updated every $\tau$ epoch, which is set as 3 via extensive experiments.} The implementation of dynamic HI construction is summarized in Algorithm \ref{algorithm1}.

\begin{algorithm}[H]
    \caption{The Training Algorithm of the Dynamic HI Construction}
    \hspace*{0.02in} {\bf Input:}
        Degradation features $\{z_i\}_{i=1}^{n}$; auxiliary target updating interval $\tau$; early stop patience $p$; maximum epoch $max\_epcoh$. \\
    \hspace*{0.02in} {\bf Output:} 
        Parameters of the HI encoder $\theta_{en}^{HI}$; Parameters of the HI decoder $\theta_{de}^{HI}$; Parameters of the inner HI-prediction block $\theta_{pr}^{HI}$; Dynamic HI $\{y_k\}_{k=1}^{n}$.  
    \begin{algorithmic}[1]
        \State {Pre-train the HI encoder and decoder.}
        \State {Initilize $\{y_k\}_{k=1}^{n}$ via (\ref{eq6}) with the pre-trained HI encoder.}
        \For{$epcoh\in{\left\{1,...,max\_epcoh\right\}}$} 
        \If{$iter\%\tau==0$}
        \State {Calculate and update $\{y_k\}_{k=1}^{n}$ for the inner HI-prediction block by (\ref{eq6}).} 
        \EndIf
        \State {Calculate $L_{rec}^{HI}$, $L_{mono}^{HI}$, and $L_{pre}^{HI}$ by (\ref{eq7}), (\ref{eq8}), and (\ref{eq11}).}
        \State {Calculate $L_{combined}$ by (\ref{eq12}).}
        \If{Meet the early stop patience $p$}
        \State {Stop training.}
        \EndIf
        \State {Update $\theta_{ge}^{HI}$, $\theta_{pr}^{HI}$ by stochastic gradient descent.}
        \EndFor
        \State {Construct dynamic HI $\{y_k\}_{k=1}^{n}$ with well-trained dynamic HI-generating module.}
    \end{algorithmic}
    \label{algorithm1}
\end{algorithm}
    
\section{Applications}
\label{sec:Applications}
In this section, the effectiveness of the proposed method is verified using two datasets: the IEEE PHM 2012 dataset and a bearing lifecycle dataset we collected. Two prognostic tasks are conducted for comparison: the HI construction and the degradation prognosis for the rolling bearing. The experimental platform is a Dell workstation with an Intel Core i9-12900 processor and an Nvidia GeForce RTX 3090 graphics card.
\subsection{Data Description}
\subsubsection{IEEE PHM 2012 Dataset}
The IEEE PHM 2012 Prognostic Challenge datasets \cite{nectoux2012pronostia} were collected from a PRONOSTIA platform. Seventeen bearing (Type: NSK 6804DD) degradation datasets were collected under three operating conditions. Vibration signals in horizontal and vertical directions were measured by two miniature accelerometers every 10 seconds with a sampling frequency of 25.6 kHz, and each sample contains 2560 points. The horizontal signals are adopted in the experiment since the vertical signals contribute little to the characterization of the degradation \cite{wu2018degradation}.
\subsubsection{HIT-B Dataset}
\begin{figure}[!h]
\centerline{\includegraphics[width=0.4\columnwidth]{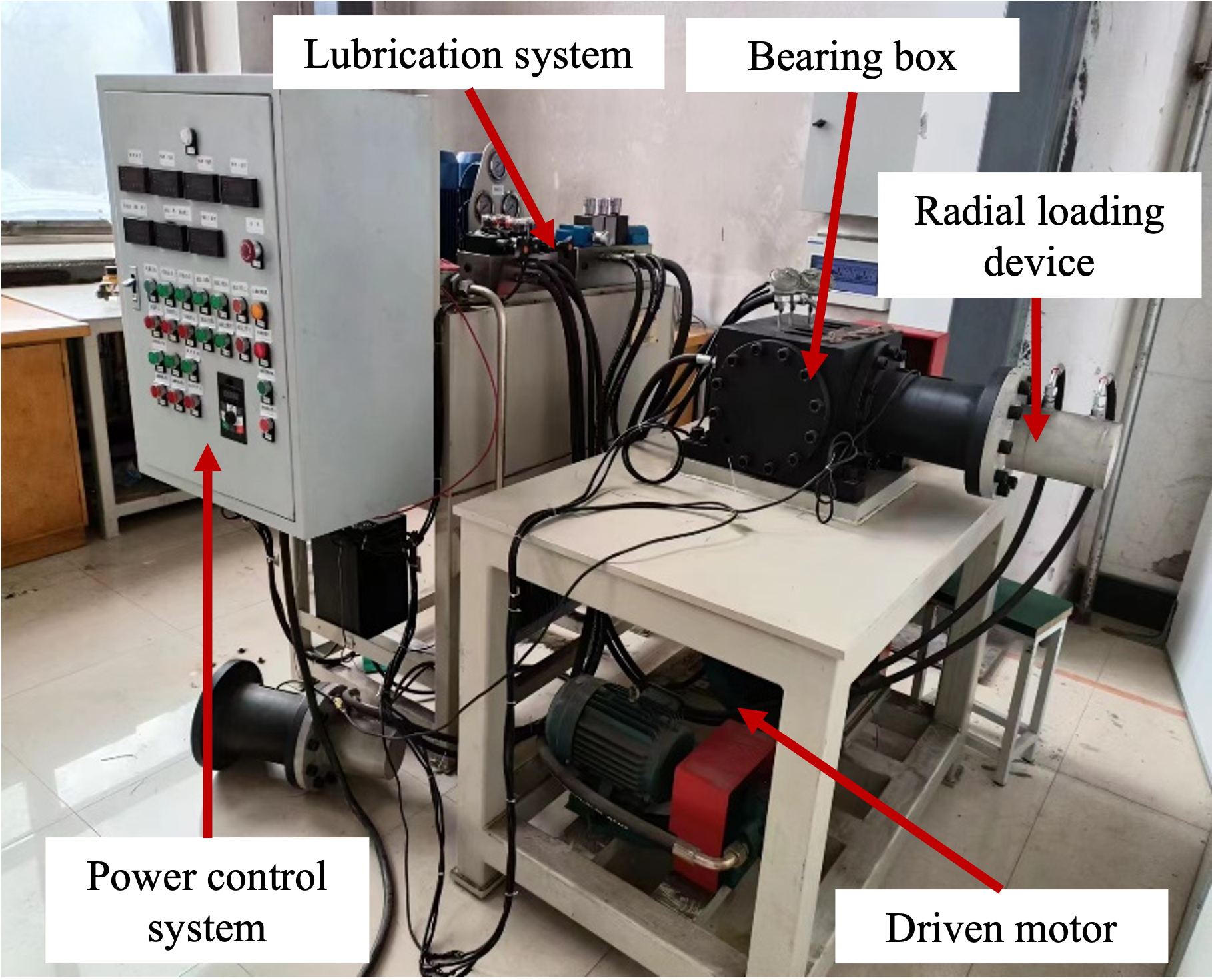}}
\caption{Accelerated test bench for HIT-B dataset.}
\label{hitbench}
\end{figure}

An accelerated test is conducted to collect a run-to-failure bearing (Type: 6211) dataset. The test bench is shown in Fig. \ref{hitbench}, which consists of a bearing box, radial loading device, driven motor, lubrication, and power control systems. The vibration signals were collected with a sampling frequency of 24kHz, a sampling period of 1 s, and a sampling interval of 30 min. The inner fault occurred at the end of the test.

\subsection{HI Construction of Rolling Bearings}
\label{HIcon}
\subsubsection{Implementation Details}
\label{HIcondetails}
The hyperparameters of the proposed method are summarized in Table. \ref{table1}. 
\textcolor{black}{For the SkipAE, the feature encoder consists of three one-dimensional convolution layers, where the channel size increases layer by layer (8-16-32) to learn complex features. The feature decoder is symmetric with the encoder, consisting of three layers with channel size decreases layer by layer (32-16-8). For the dynamic HI-generating module, considering the degradation data of one bearing is limited, a lightweight autoencoder is selected as the backbone, where an HI encoder with three full connection layers is used to map the degradation features into the one-dimensional dynamic HI. To ensure that the decomposition and prediction layers can capture sufficient degradation trend information while maintaining computational efficiency, the inner HI-prediction block consists of a one-dimensional average pooling layer for HI decomposition with a window size of 9 and two full connection layers for HI prediction with the input length of 20.}

\begin{figure}[!h]
\centering
\vspace{-3.0mm}
\subfigure[]{
\begin{minipage}[t]{0.27\linewidth}
    \centering
    \includegraphics[width=1.8in]{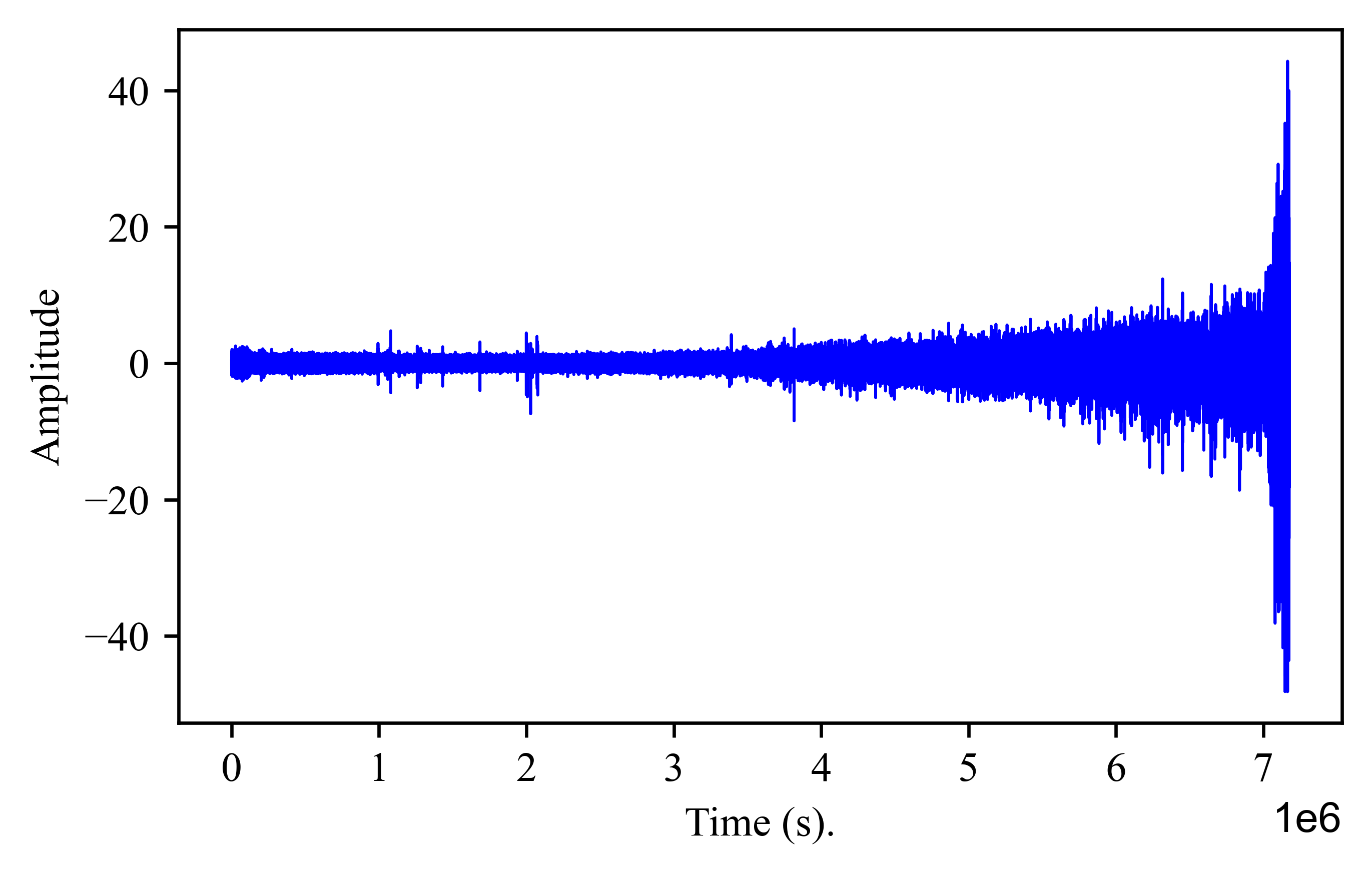}
\end{minipage}%
}
\subfigure[]{
\begin{minipage}[t]{0.27\linewidth}
    \centering
    \includegraphics[width=1.8in]{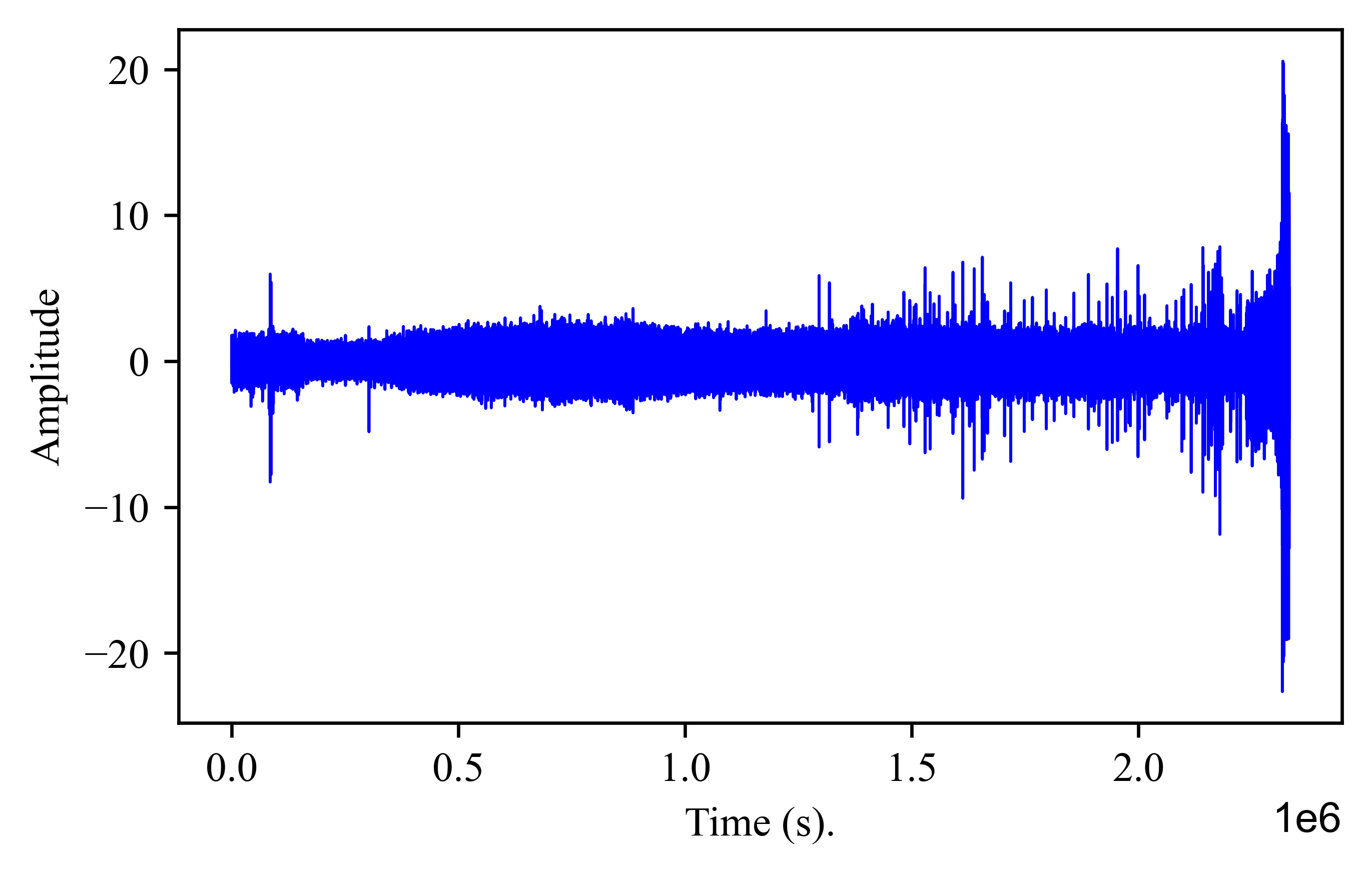}
\end{minipage}%
}
\subfigure[]{
\begin{minipage}[t]{0.27\linewidth}
    \centering
    \includegraphics[width=1.8in]{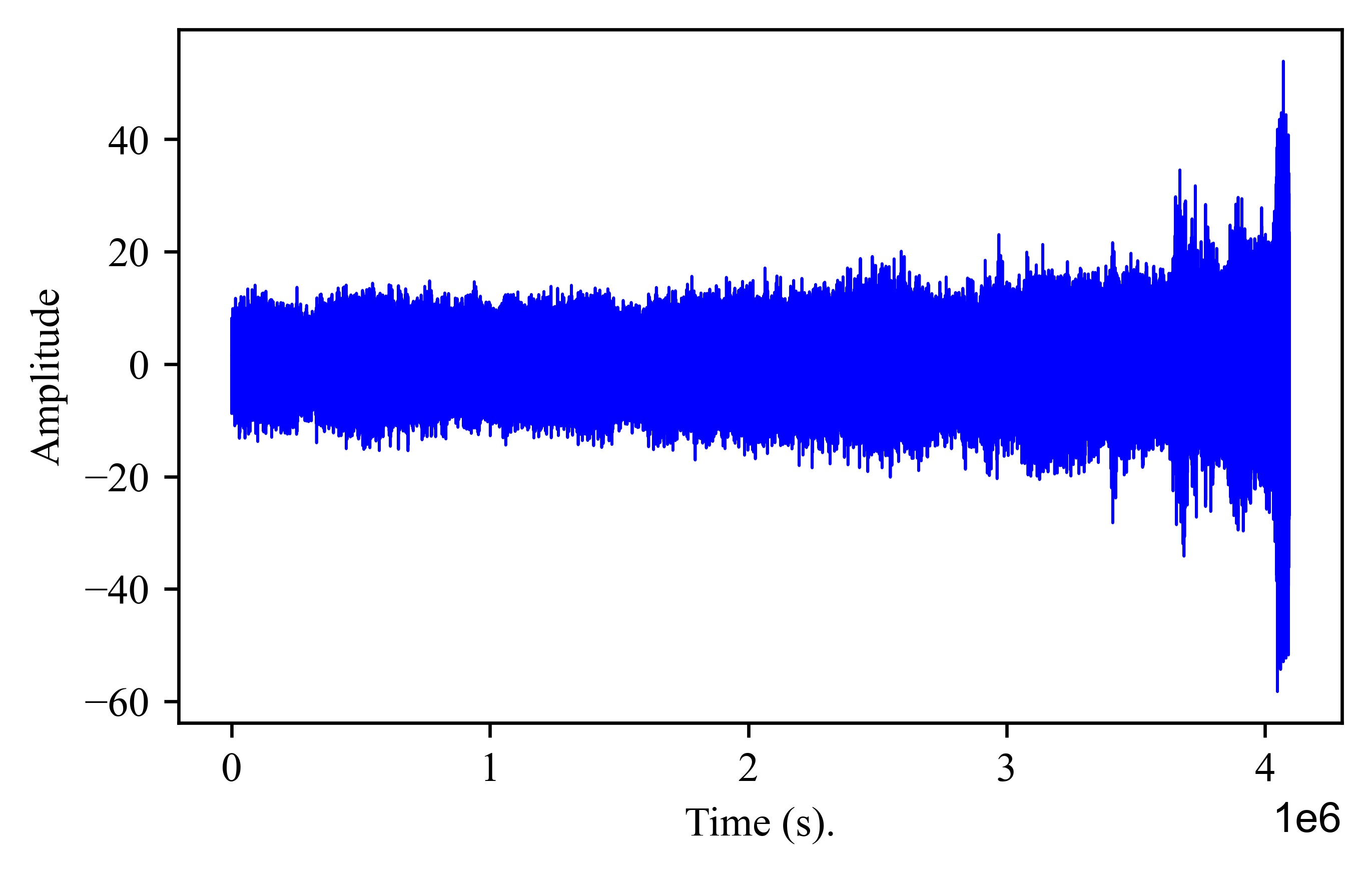}
\end{minipage}%
}
\centering
\caption{Life cycle time domain waveforms of three test bearings. (a) P-Bearing1\_1, (b) P-Bearing2\_1, (c) H-Bearing.}
\label{waveforms}
\end{figure}

The input of the SkipAE is frequency signals with a length of 1280. The dimension of the degradation features is set as 32. The Adam optimizer is adopted during the entire training process. In the training of SkipAE and the dynamic HI construction network, the training epoch, learning rate, and batch size are set as 1000, 0.001, and 64, respectively. The early stopping technique with 15 epochs is applied in the training of the dynamic HI construction network. Three distinct research tasks are proposed, as shown in Table. \ref{table2}, to assess the effectiveness and robustness of the proposed method. The P-Bearing and H-Bearing denote the bearings of the IEEE PHM 2012 dataset and HIT-B dataset, respectively. Tasks 1 and 2 are conducted on the same machine with known working conditions. Specifically, for Task 1, six bearings on condition 1 are used as the training set, and one bearing is used for testing. For Task 2, the bearings under condition 2 are used to test the method under different operating conditions. Task 3 is performed on different machines with unseen working conditions. The P-Bearings under condition 1 are used for training, and the H-Bearing is used for testing. The time domain waveforms of these three test bearings are presented in Fig. \ref{waveforms}.

\begin{table}[!t]
\centering
\caption{Parameter Settings of the Proposed Model}
\label{table1}
\setlength{\tabcolsep}{3pt}
\scalebox{0.75}{
\begin{tabular}{l|l|l|l}
\hline
\quad & Layer & KernelSize@Channel, Stride & OutputSize\\
\hline
\multirowcell{12}{Skip-connection- \\ based Autoencoder \\ (SkipAE)} & $Conv$ &$1\times10@8$, 2 & $8\times636$ \\
& $Conv$ & $1\times10@16$, 2 & $16\times314$ \\
& $Conv$ & $1\times10@32$, 2 & $32\times153$ \\
& $Flatten$ & $\backslash$ & $4896$ \\
& $FC$ & $\backslash$ & $32$ \\
& $FC$ & $\backslash$ & $4896$ \\
& $ConcatConv$ & $1\times1@32$, 1 & $32\times153$ \\
& $DeConv$ & $1\times10@16$, 2 & $16\times314$ \\
& $ConcatConv$ & $1\times1@16$, 1 & $16\times314$ \\
& $DeConv$ & $1\times10@8$, 2 & $8\times636$ \\
& $ConcatCon$v & $1\times1@8$, 1 & $8\times636$ \\
& $DeConv$ & $1\times10@1$, 2 & $1\times1280$ \\
\hline
\multirowcell{6}{HI-generating \\ Module} & $FC$ &$\backslash$ & 16\\
& $FC$ & $\backslash$ & 8 \\
& $FC$ & $\backslash$ & 1 \\
& $FC$ & $\backslash$ & 8 \\
& $FC$ & $\backslash$ & 16 \\
& $FC$ & $\backslash$ & 32 \\
\hline
\multirowcell{3}{\textcolor{black}{inner HI-prediction} \\ \textcolor{black}{block}} & $AvgPool1d$ &$1\times9@\backslash$, 1 & $1\times20$\\
& $FC$ & $\backslash$ & $1\times{1}$ \\ & \textcolor{black}{$FC$} & \textcolor{black}{$\backslash$} & \textcolor{black}{$1\times{1}$} \\
\hline
\multicolumn{4}{p{330pt}}{Batch normalization \cite{ioffe1502normalization} and LeakyRelu \cite{dahl2013improving} function are applied for all layers of the SkipAE. Relu function is used for all layers of the dynamic HI-generating module.}\\
\multicolumn{4}{p{330pt}}{Conv=1D Convolution, FC=Full Connection, ConcatConv=$1\times1$ Convolution, DeConv=1D Transposed Convolution.}
\end{tabular}}
\vspace{-3.0mm}
\end{table}

\begin{table}[!h]
\centering
\vspace{-3.0mm}
\caption{Research Tasks for Construction and Degradation Trend Prediction of Bearing HI}
\label{table2}
\setlength{\tabcolsep}{5pt}
\scalebox{0.85}{
\begin{tabular}{l|l|l|l}
\hline
\makecell[l]{Tasks} & \makecell[l]{Task 1} & \makecell[l]{Task 2}& \makecell[l]{Task 3}\\
\hline
\multirow{6}*{Training Set} & P-Bearing1\_2 & P-Bearing2\_2 & P-Bearing1\_2 \\
~ & P-Bearing1\_3 & P-Bearing2\_3 & P-Bearing1\_3\\
~ & P-Bearing1\_4 & P-Bearing2\_4 & P-Bearing1\_4\\
~ & P-Bearing1\_5 & P-Bearing2\_5 & P-Bearing1\_5\\
~ & P-Bearing1\_6 & P-Bearing2\_6 & P-Bearing1\_6\\
~ & P-Bearing1\_7 & P-Bearing2\_7 & P-Bearing1\_7\\
\hline
\multirow{1}*{Test Set} & P-Bearing1\_1 & P-Bearing2\_1 & H-Bearing\\
\hline
\end{tabular}}
\end{table}

\subsubsection{Evaluation Metrics}
To quantitatively evaluate the HI construction methods, Monotonicity (Mon), Trendability (Tred), Robustness(Rob), and Hybrid Scale (HS) \cite{lei2018machinery} are selected as evaluation metrics. Higher values indicate better performance. 
Specifically, \textcolor{black}{Mon is selected to evaluate whether the HI can reflect the monotonic degradation trend of the equipment, as the degradation processes of machinery are irreversible in the real industrial environment.} Mon is defined as
\begin{equation}
\begin{aligned}
    Mon(y) = \frac{\left |\sum_{k=1}^{K-1}\mathbb{I}(y_{k+1}^{tr} - y_{k}^{tr}>0)-\sum_{k=1}^{K-1}\mathbb{I}(y_{k}^{tr} - y_{k+1}^{tr}>0)\right |}{K-1},\label{eq13}
\end{aligned}
\end{equation}
where $\mathbb{I}(\cdot)$ is an indicator function, and $y_{k}^{tr}$ presents the trend value of HI at time $k$ calculated by smoothing methods following the traditions in HI performance assessment \cite{lei2018machinery}. 

\textcolor{black}{Tred is selected to evaluate whether the trend of the HI correlates with operating time, as components are more likely to degrade gradually with increased operating time.} Tred can be calculated as
\begin{equation}
\begin{aligned}
    Tred(y) = \frac{\left |K\left (\sum_{k=1}^{K}y_{k}^{tr}t_k \right )-\sum_{k=1}^{K}y_{k}^{tr}\sum_{k=1}^{K}t_k\right |}{\sqrt{\left [K\sum_{k=1}^{K}(y_{k}^{tr})^2-\left (\sum_{k=1}^{K}y_{k}^{tr}\right )^2 \right ]\left [K\sum_{k=1}^{K}(t_{k})^2-\left (\sum_{k=1}^{K}t_{k}\right )^2 \right]}}.\label{eq14}
\end{aligned}
\end{equation}

\textcolor{black}{Rob is chosen to assess whether the HI is robust to interferences and exhibits a smooth degradation trend, given that measurement noise, the stochastic nature of degradation processes, and variations in operational conditions are unavoidable.} Rob is represented as
\begin{equation}
    Rob(y) = \frac{1}{K}\sum_{k=1}^{K}exp\left (-\left |\frac{y_{k}-y_{k}^{tr}}{y_k}\right |\right ).\label{eq15}
\end{equation}

Finally, to comprehensively assess the performance of an HI, an HS index is proposed as
\begin{equation}
    HS(y) = \frac{Mon(y)+Tred(y)+Rob(y)}{3}.\label{eq16}
\end{equation}

\subsubsection{Comparison Results of Various Methods}
\label{sec323}
To demonstrate the superiority of the proposed method, various popular unsupervised HI construction methods, including RMS, P-Entropy \cite{kumar2022state}, ISOMAP, KPCA, CEEMDAN \cite{cheng2021convolutional}, VAE \cite{yao2019unsupervised}, SSAE \cite{sun2018deep}, are applied for comparison. These methods cover the mainstream machine learning and deep learning-based approaches. 
For a fair comparison, the network for VAE is the same as the proposed method, and the SSAE is replicated with two stacked linear networks with structures of 1280-400-1280 and 400-1-400. 

\begin{table*}[!b]
\vspace{-5mm}
\centering
\caption{Comparison Results of the Proposed Method with Other Methods in the Construction of Bearing HI.}
\setlength\tabcolsep{3.5pt}
\label{table3}
\scalebox{0.9}{
\begin{tabularx}{0.88\textwidth}{ccccccccccccc}
  \hline
  \multirowcell{2}{Method} & \multicolumn{4}{c}{Task 1} & \multicolumn{4}{c}{Task 2} & \multicolumn{4}{c}{Task 3}\\
  \quad & Mon & Tred & Rob & HS & Mon & Tred & Rob & HS & Mon & Tred & Rob & HS\\
  \hline
  RMS & 0.006 & 0.537 & 0.963 & 0.502 & 0.055 & 0.373 & 0.832 & 0.420 & 0.028 & 0.453 & 0.886 & 0.456 \\
  P-Entropy & 0.027 & 0.280 & 0.553 & 0.287 & 0.062 & 0.263 & 0.598 & 0.308 & 0.036 & 0.395 & 0.823 & 0.418 \\
  ISOMAP & 0.056 & 0.854 & 0.963 & 0.625 & 0.101 & 0.628 & 0.942 & 0.557 & 0.068 & 0.775 & 0.922 & 0.588 \\
  KPCA & 0.026 & $\mathbf{0.886}$ & 0.864 & 0.592 & 0.103 & 0.593 & 0.946 & 0.547 & 0.026 & 0.875 & 0.911 & 0.604 \\
  CEEMDAN & 0.025 & 0.332 & 0.807 & 0.388 & 0.064 & 0.333 & 0.926 & 0.441 & 0.038 & 0.368 & 0.910 & 0.439 \\
  VAE & 0.028 & 0.651 & 0.975 & 0.551 & 0.097 & 0.556 & 0.981 & 0.545 & 0.057 & 0.852 & $\mathbf{0.973}$ & 0.627 \\
  SSAE & 0.017 & 0.505 & 0.893 & 0.472 & 0.042 & 0.343 & 0.959 & 0.447 & 0.041 & 0.779 & 0.803 & 0.541 \\
  Ours & $\mathbf{0.069}$ & 0.854 & $\mathbf{0.983}$ & $\mathbf{0.635}$ & $\mathbf{0.105}$ & $\mathbf{0.639}$ & $\mathbf{0.984}$ & $\mathbf{0.576}$ & $\mathbf{0.074}$ & $\mathbf{0.899}$ & $\mathbf{0.973}$ & $\mathbf{0.649}$ \\
  \hline
\end{tabularx}}
\end{table*}

\begin{figure}[!h]
\vspace{-3mm}
\centering
\subfigure[]{
\begin{minipage}[t]{0.18\linewidth}
    \centering
    \includegraphics[width=1.2in]{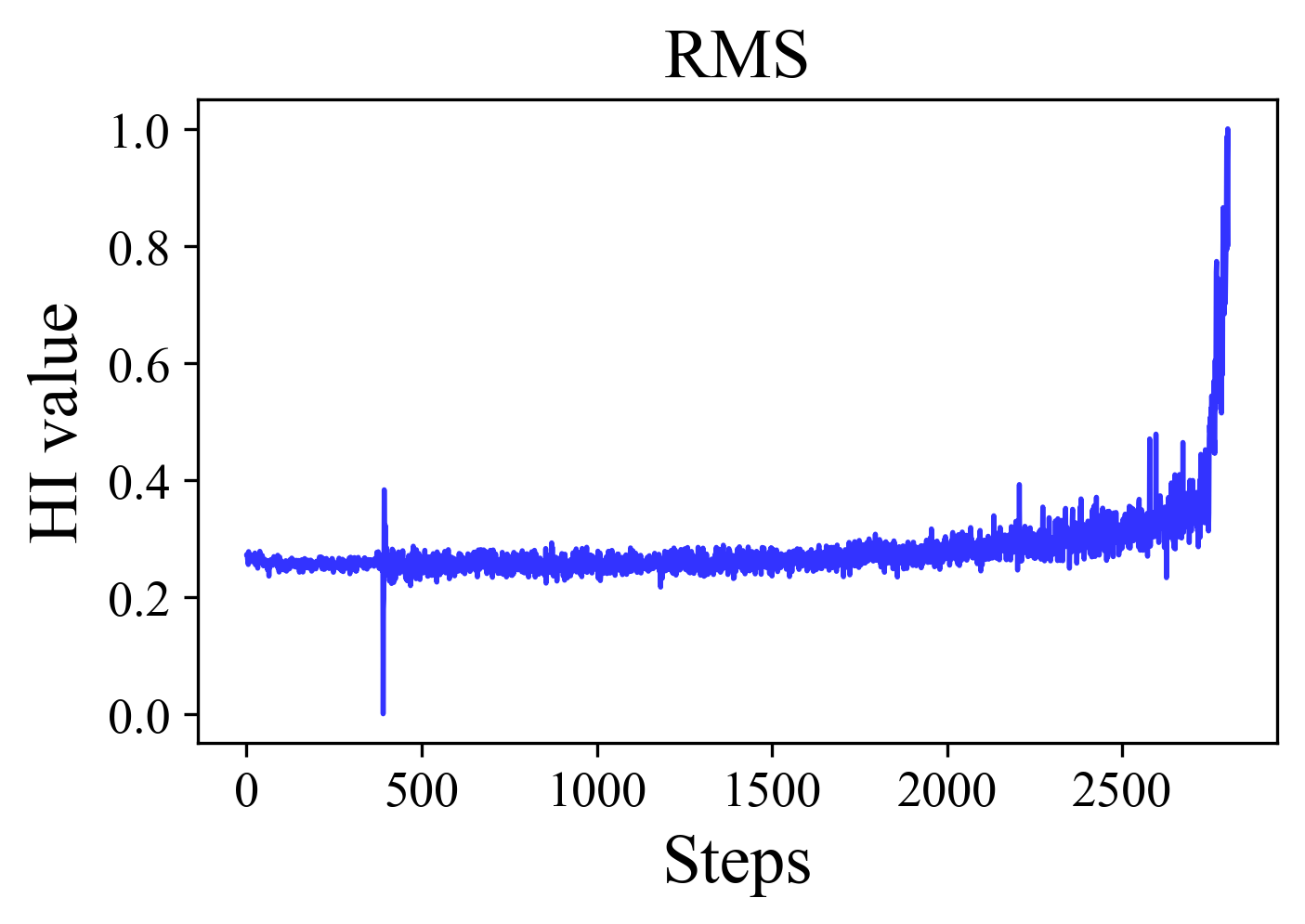}
\end{minipage}%
}%
\subfigure[]{
\begin{minipage}[t]{0.18\linewidth}
    \centering
    \includegraphics[width=1.2in]{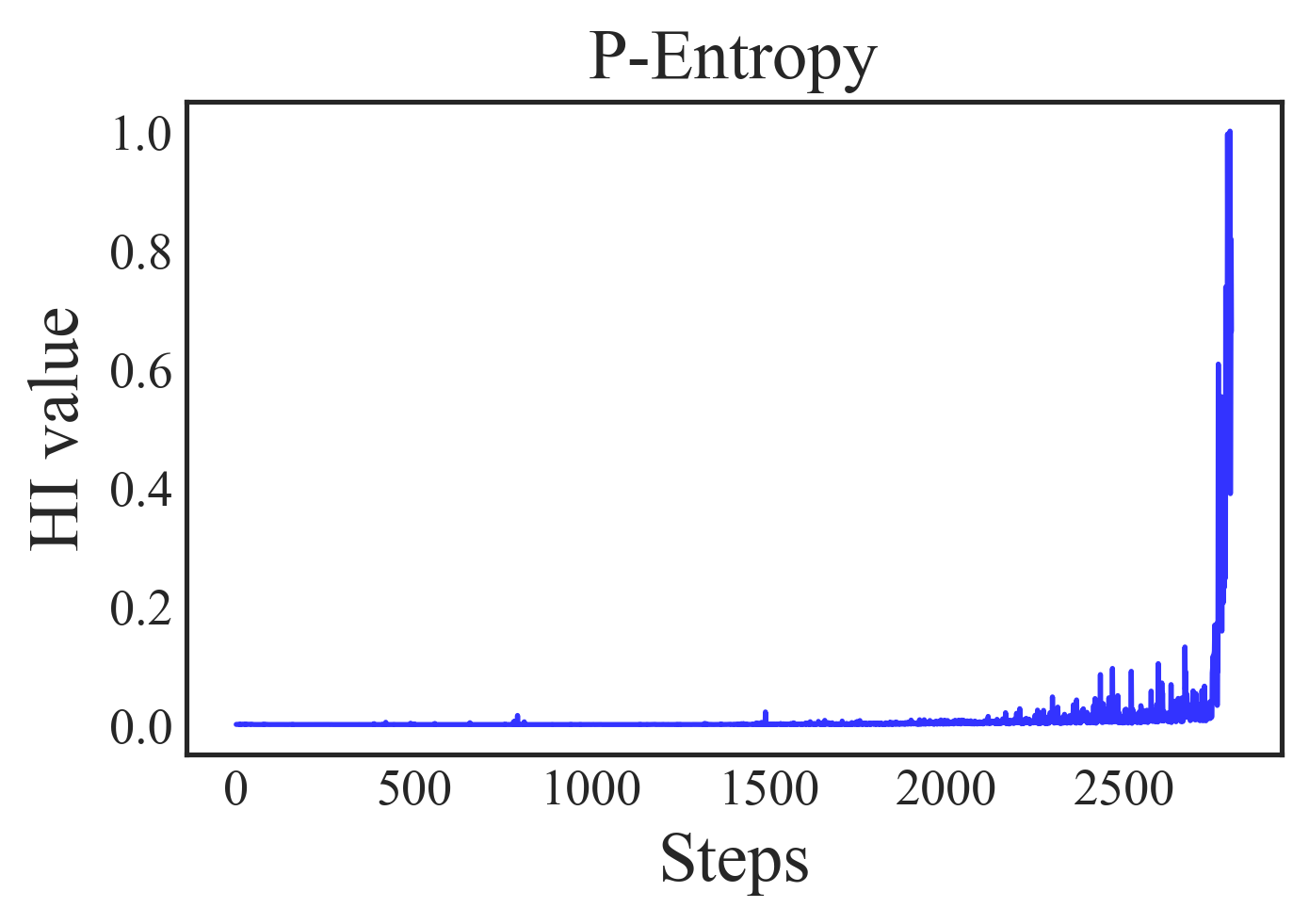}
\end{minipage}%
}
\subfigure[]{
\begin{minipage}[t]{0.18\linewidth}
    \centering
    \includegraphics[width=1.2in]{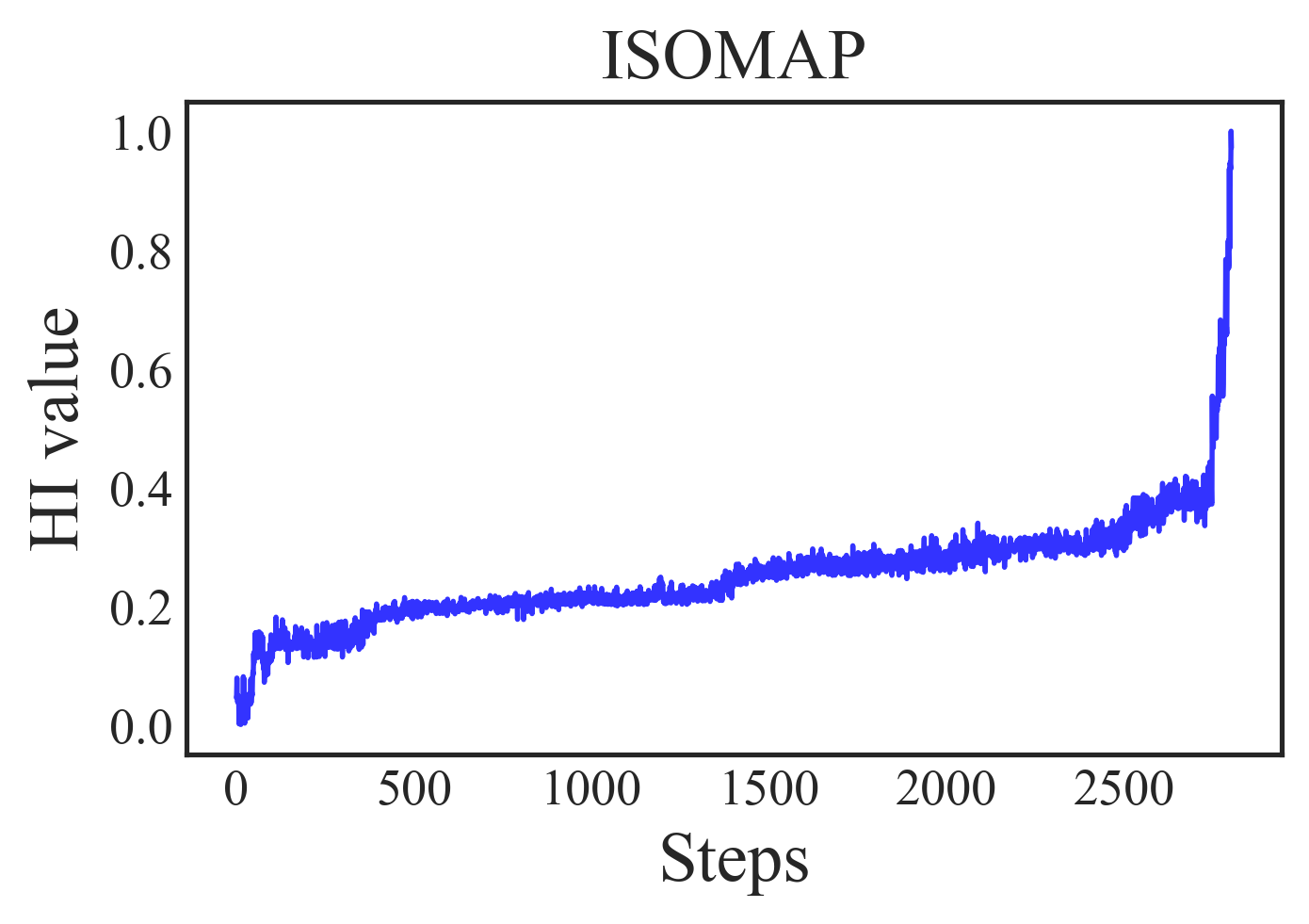}
\end{minipage}%
}
\subfigure[]{
\begin{minipage}[t]{0.18\linewidth}
    \centering
    \includegraphics[width=1.2in]{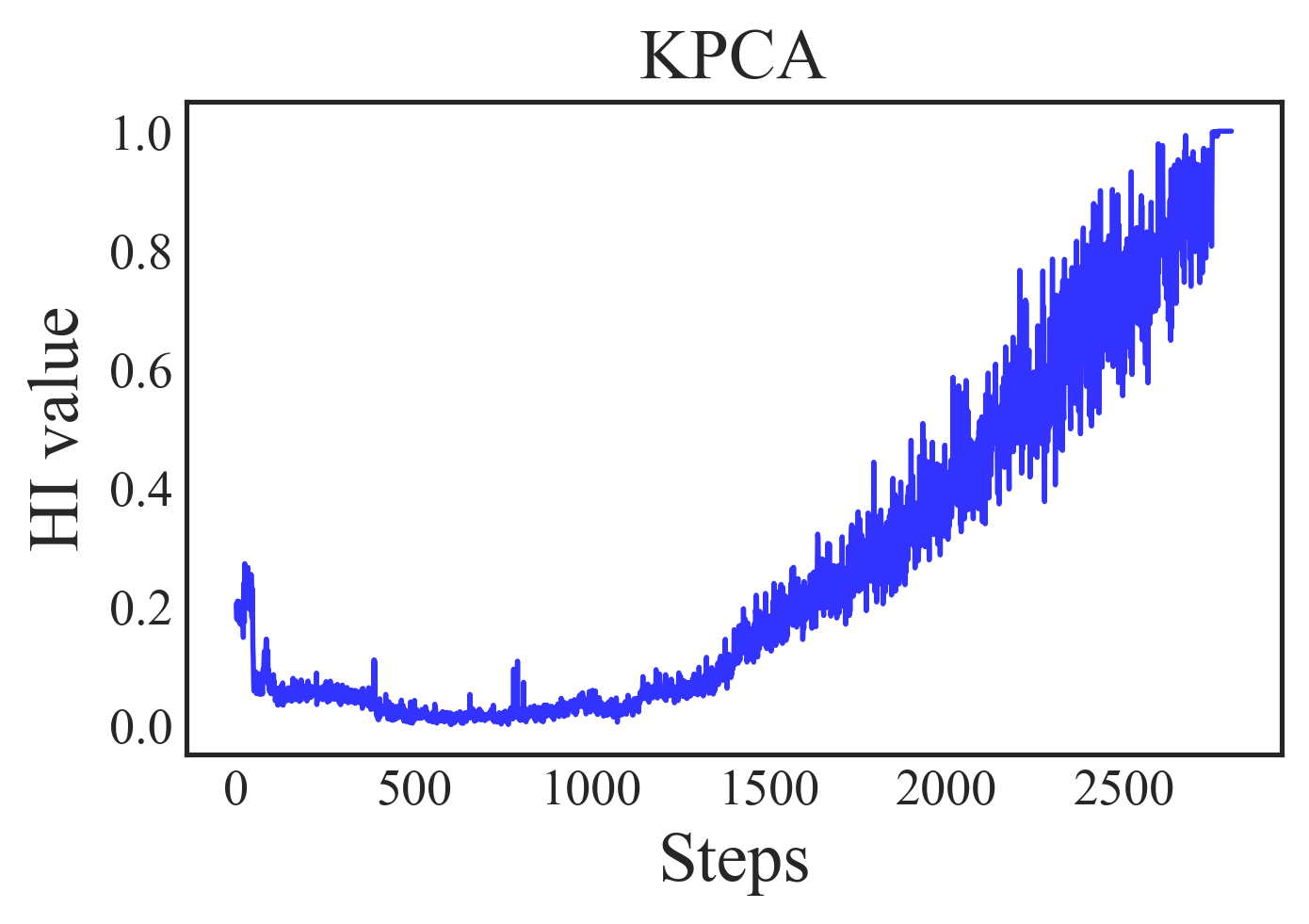}
\end{minipage}%
}\\
\vspace{-3mm}
\subfigure[]{
\begin{minipage}[t]{0.18\linewidth}
    \centering
    \includegraphics[width=1.2in]{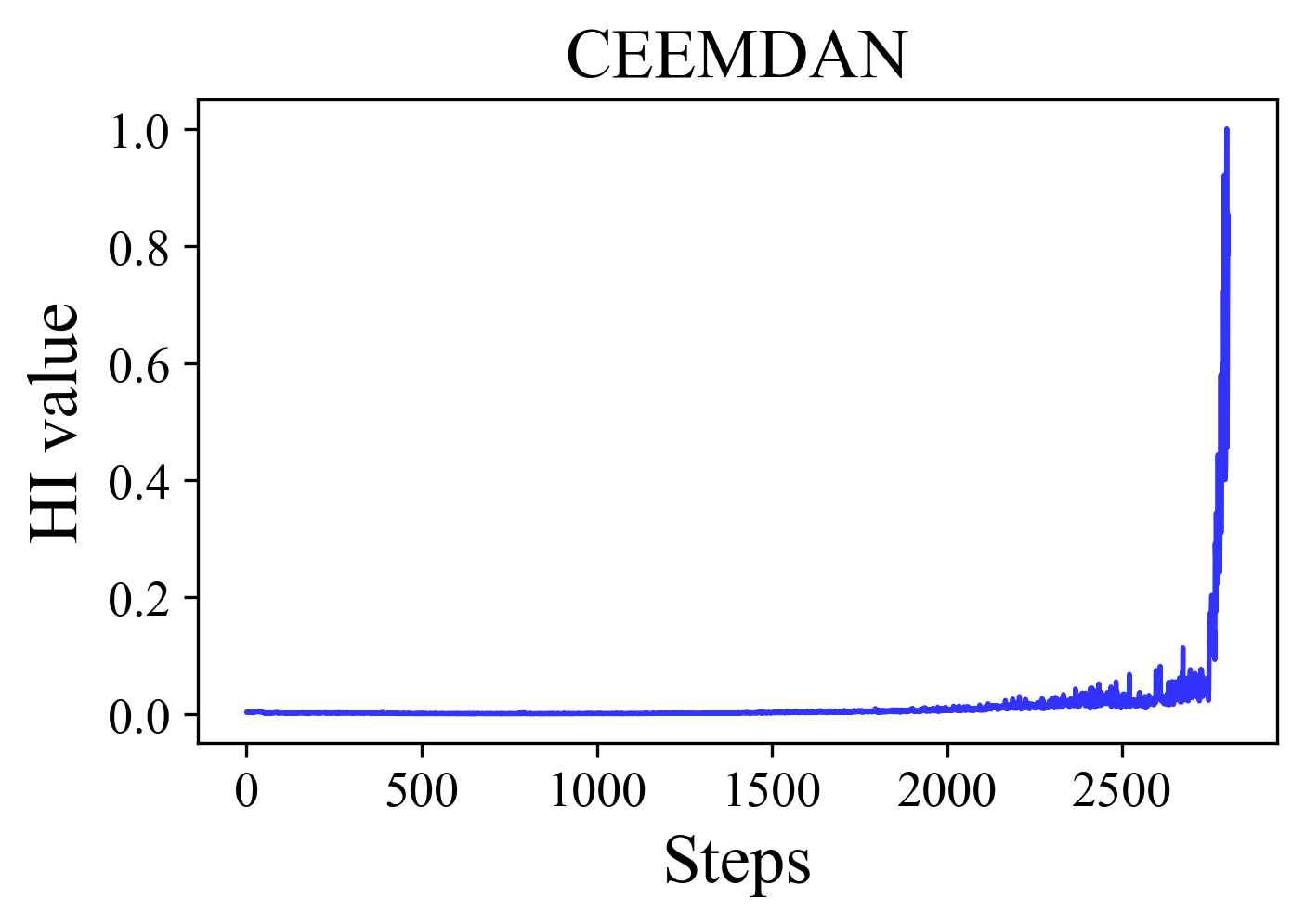}
\end{minipage}%
}%
\subfigure[]{
\begin{minipage}[t]{0.18\linewidth}
    \centering
    \includegraphics[width=1.2in]{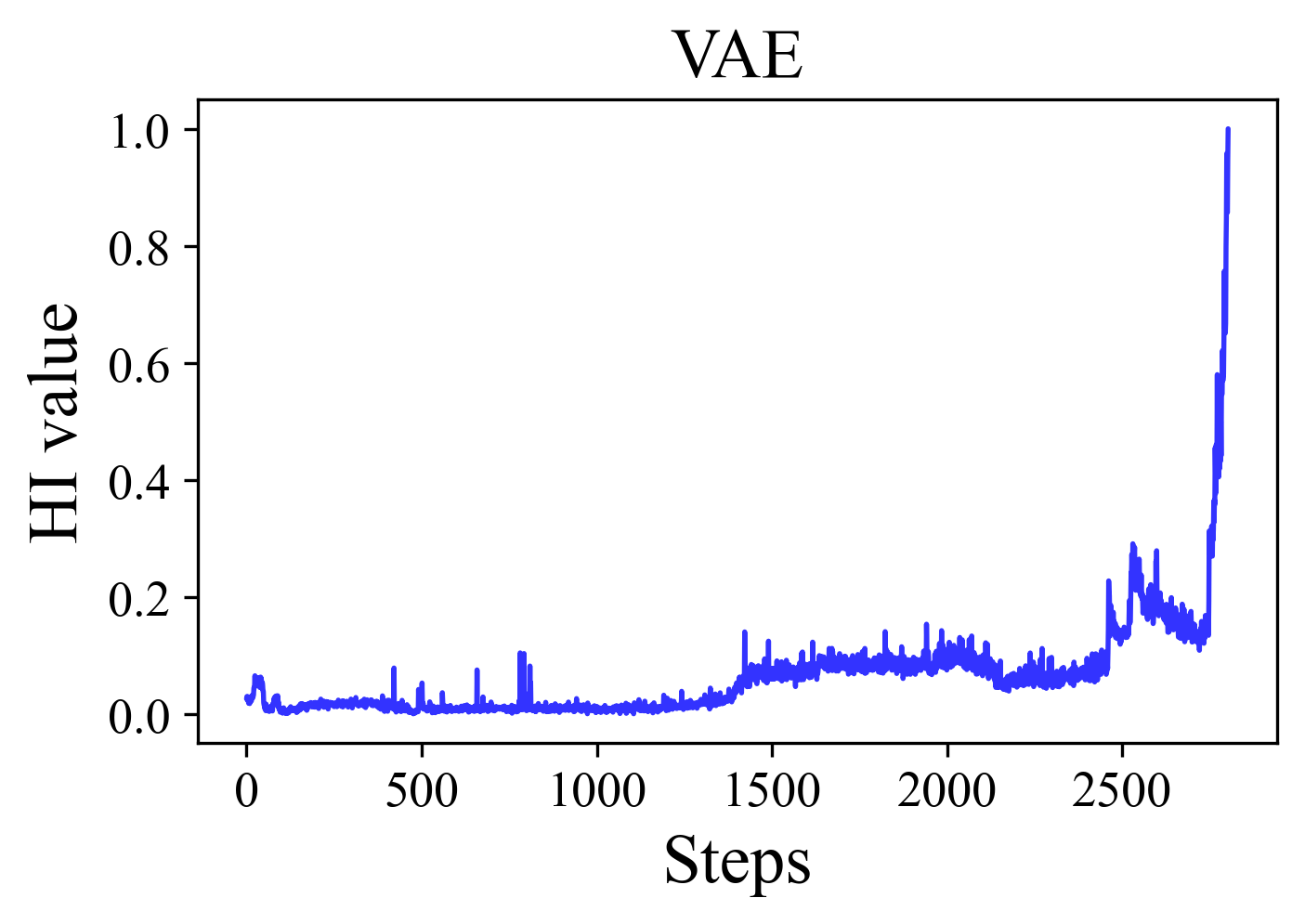}
\end{minipage}%
}
\subfigure[]{
\begin{minipage}[t]{0.18\linewidth}
    \centering
    \includegraphics[width=1.2in]{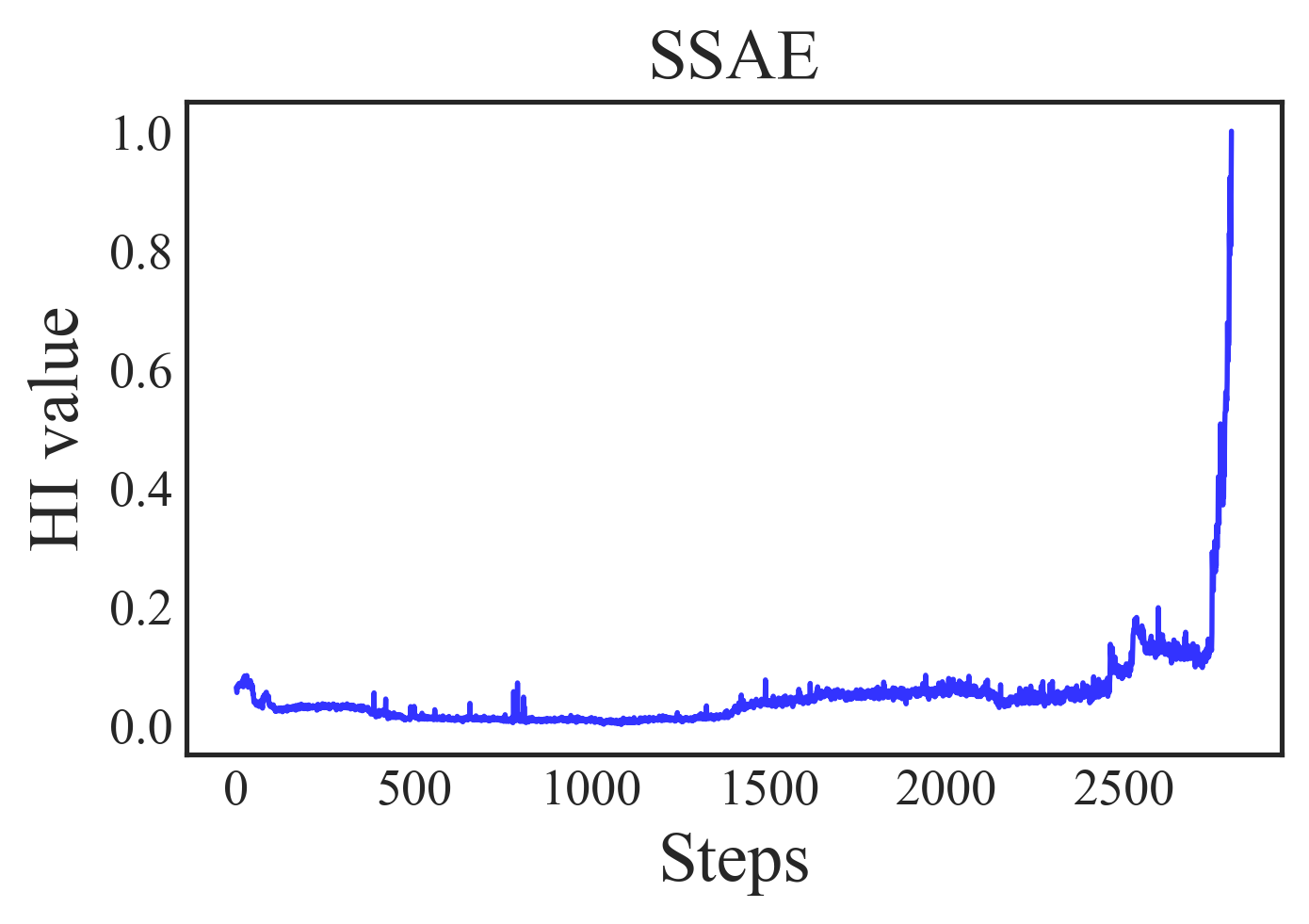}
\end{minipage}%
}
\subfigure[]{
\begin{minipage}[t]{0.18\linewidth}
    \centering
    \includegraphics[width=1.2in]{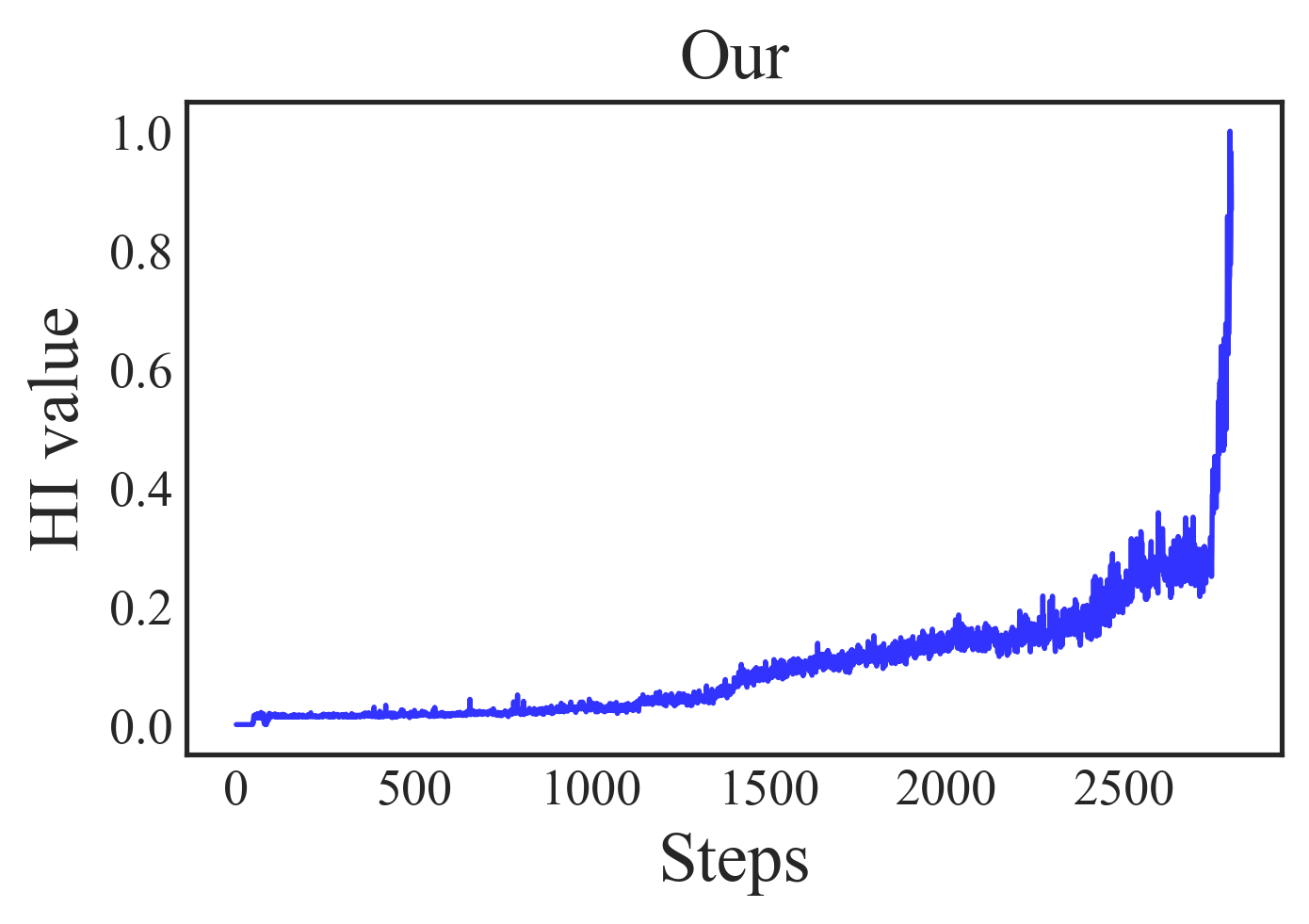}
\end{minipage}%
}
\centering
\vspace{-3mm}
\caption{\textcolor{black}{The constructed HIs for Task 1. (a) RMS, (b) P-Entropy, (c) ISOMAP, (d) KPCA, (e) CEEMDAN, (f) VAE, (g) SSAE, (h) Ours.}}
\label{task1_con}
\end{figure}

\begin{figure}[!h]
\vspace{-5mm}
\centering
\subfigure[]{
\begin{minipage}[t]{0.18\linewidth}
    \centering
    \includegraphics[width=1.2in]{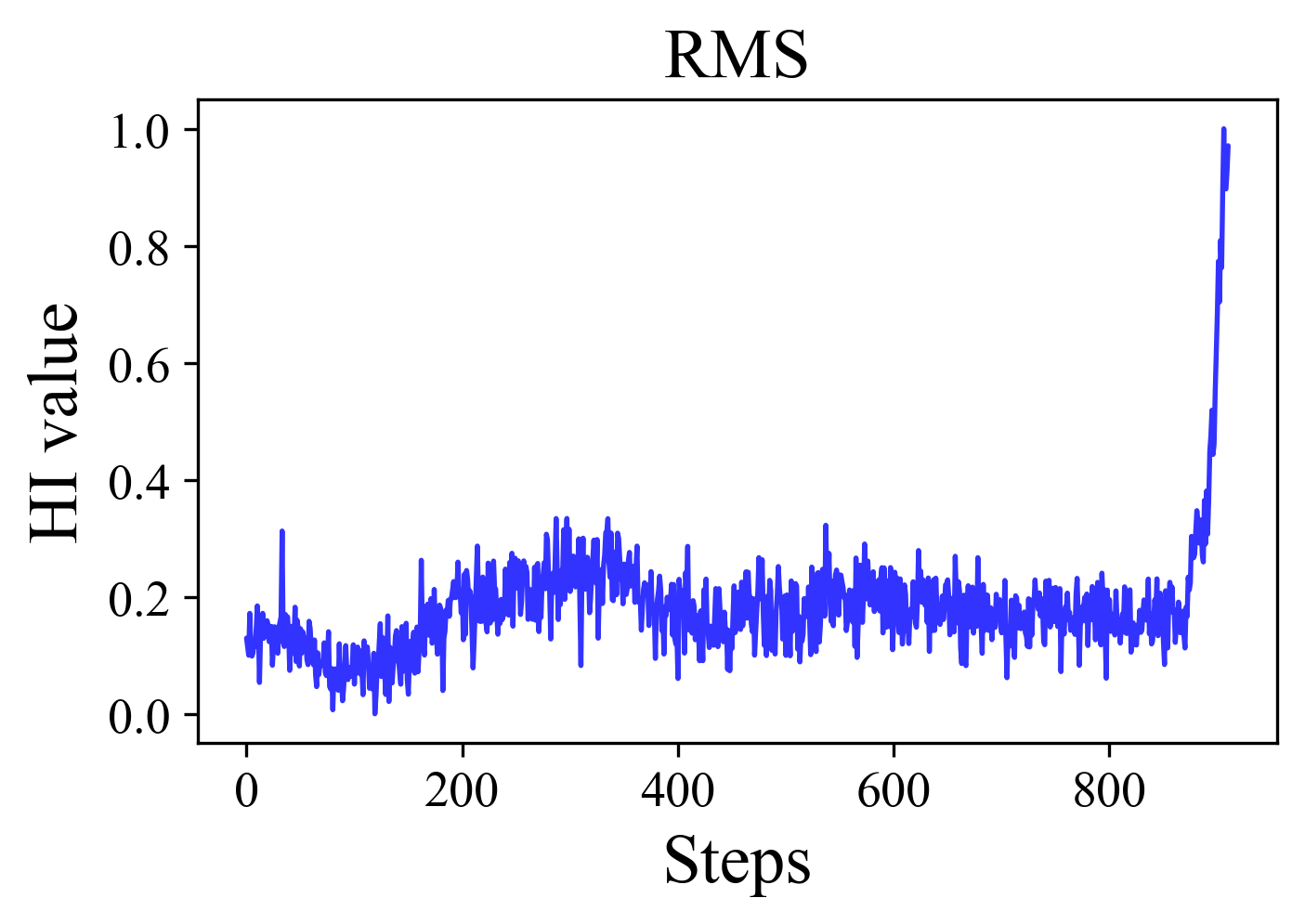}
\end{minipage}%
}%
\subfigure[]{
\begin{minipage}[t]{0.18\linewidth}
    \centering
    \includegraphics[width=1.2in]{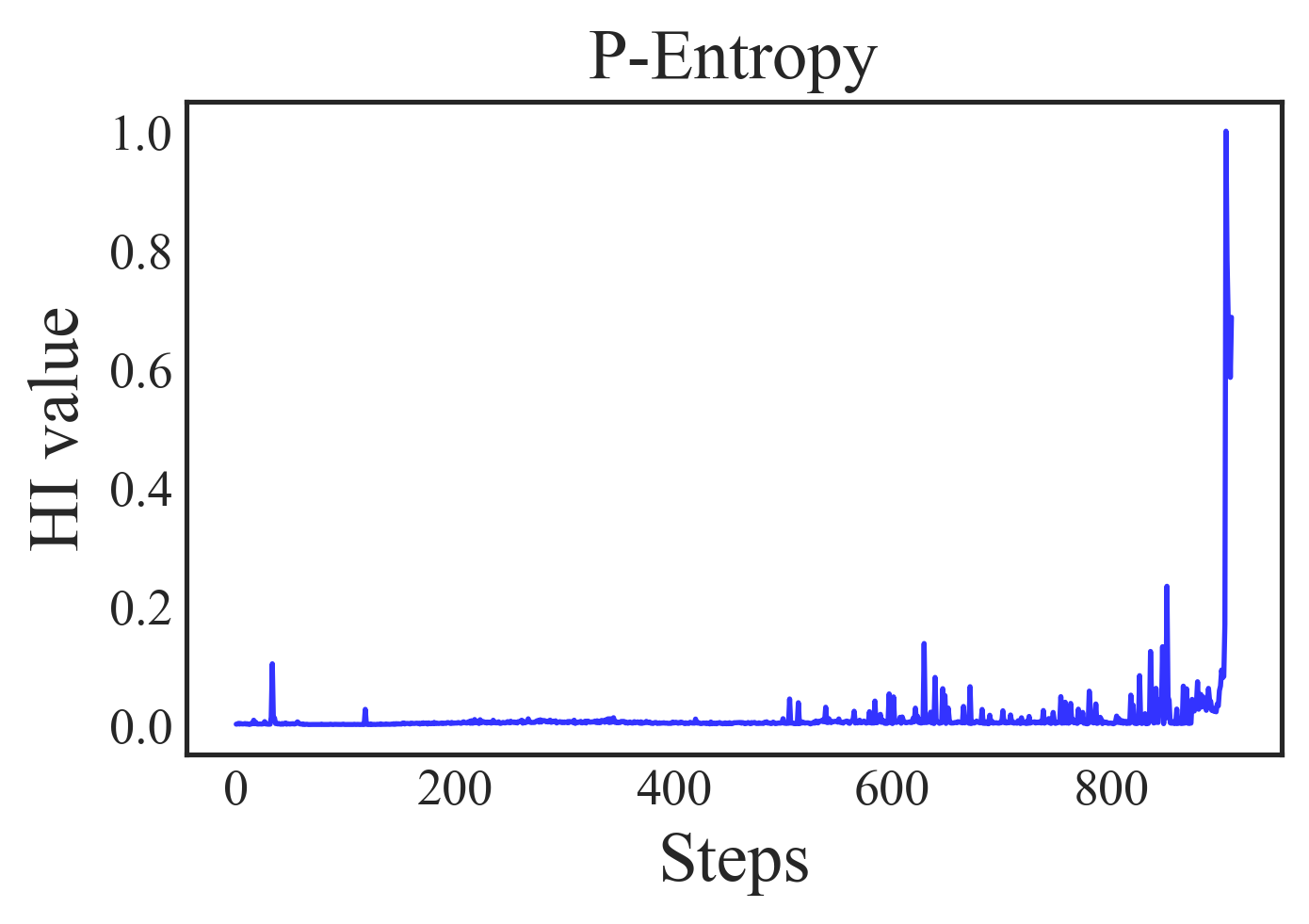}
\end{minipage}%
}
\subfigure[]{
\begin{minipage}[t]{0.18\linewidth}
    \centering
    \includegraphics[width=1.2in]{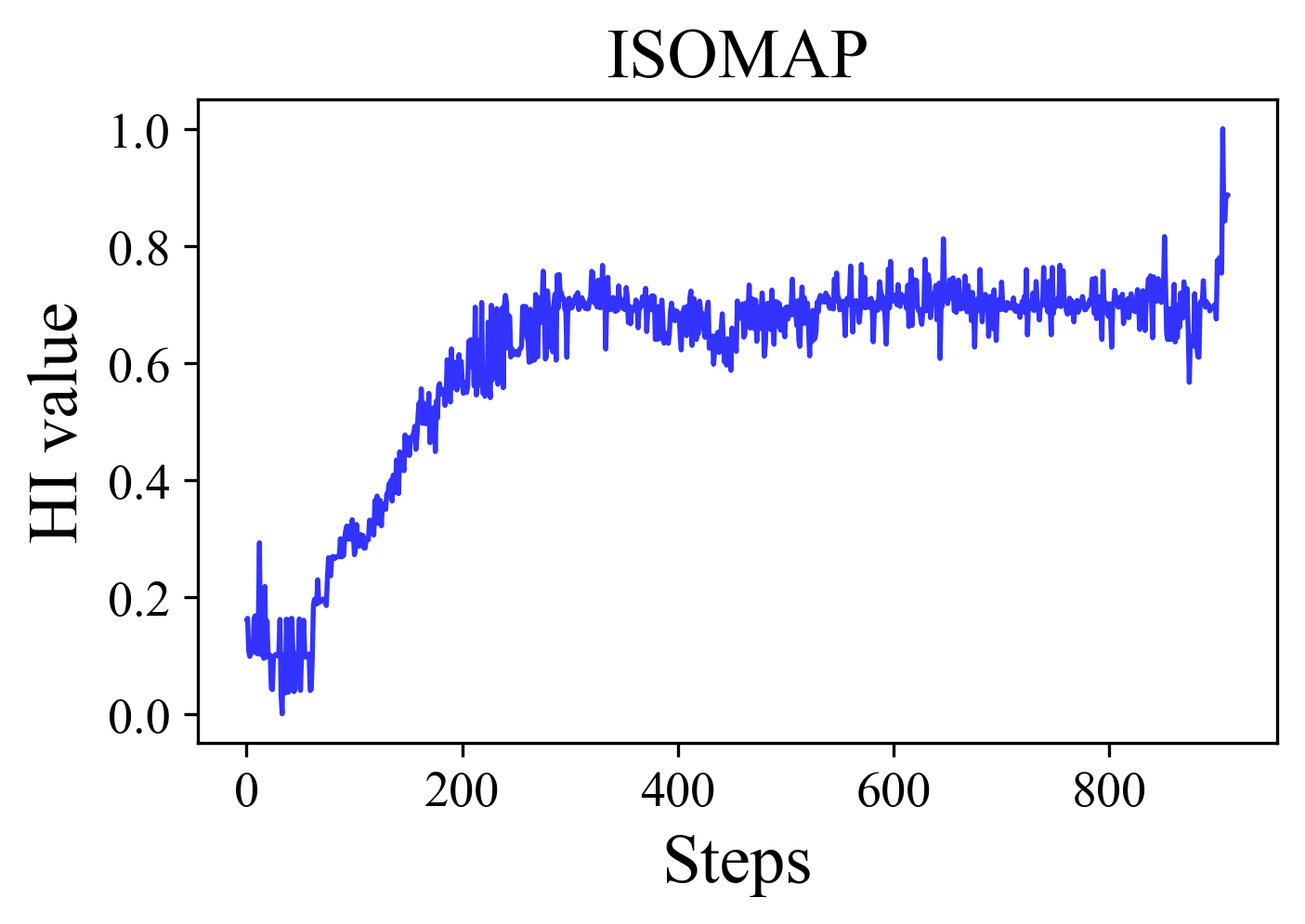}
\end{minipage}%
}
\subfigure[]{
\begin{minipage}[t]{0.18\linewidth}
    \centering
    \includegraphics[width=1.2in]{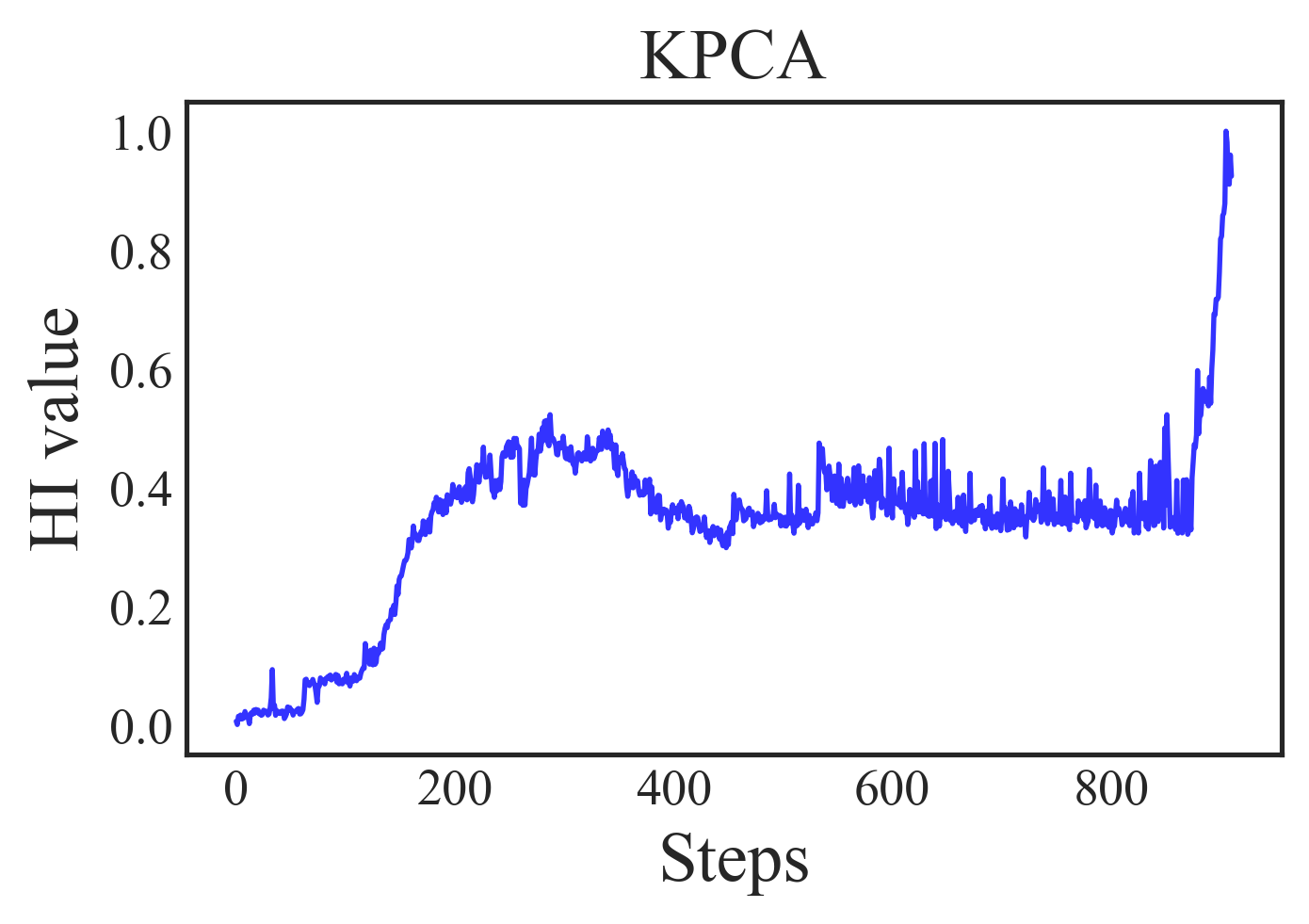}
\end{minipage}%
}\\
\vspace{-3mm}
\subfigure[]{
\begin{minipage}[t]{0.18\linewidth}
    \centering
    \includegraphics[width=1.2in]{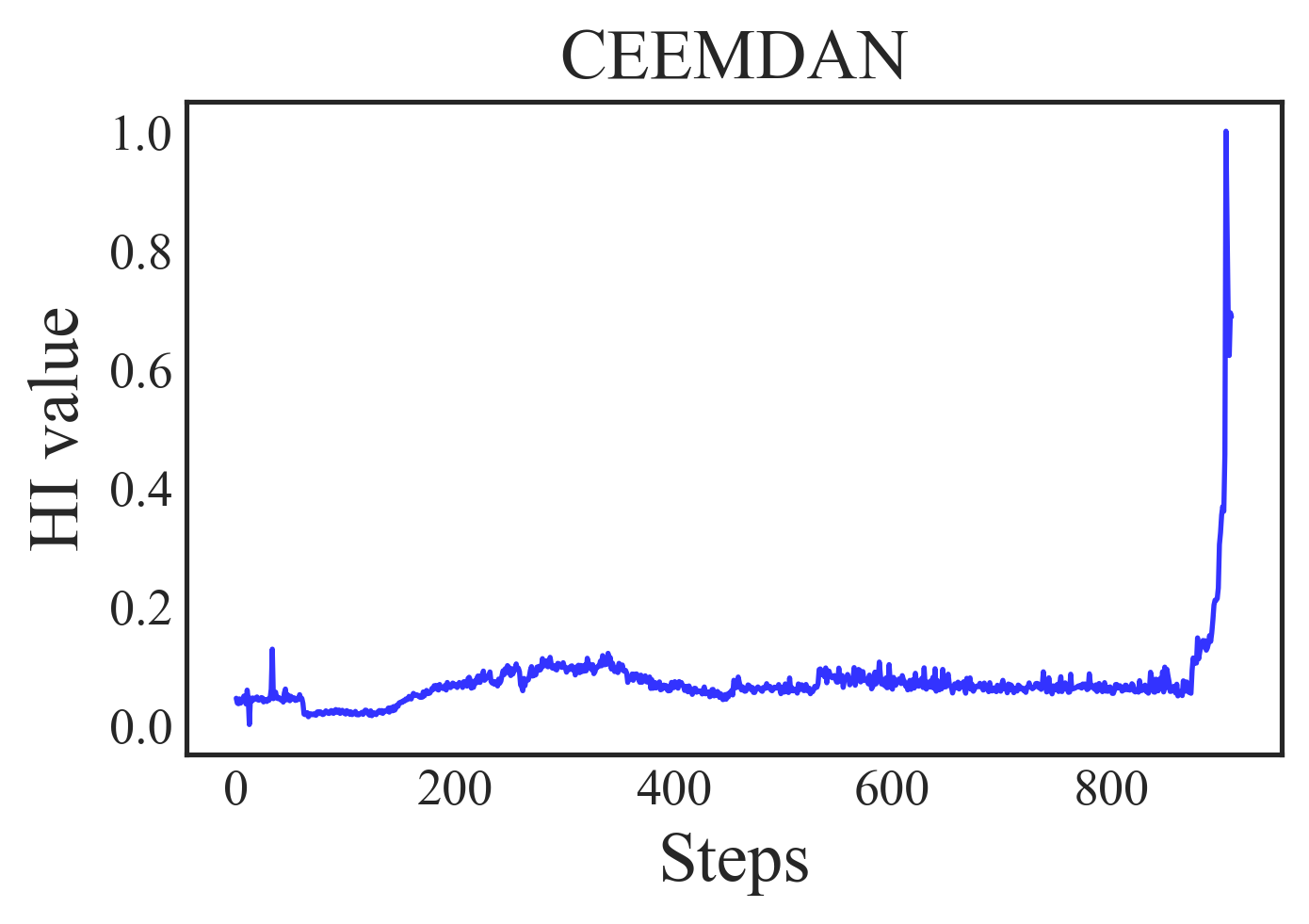}
\end{minipage}%
}%
\subfigure[]{
\begin{minipage}[t]{0.18\linewidth}
    \centering
    \includegraphics[width=1.2in]{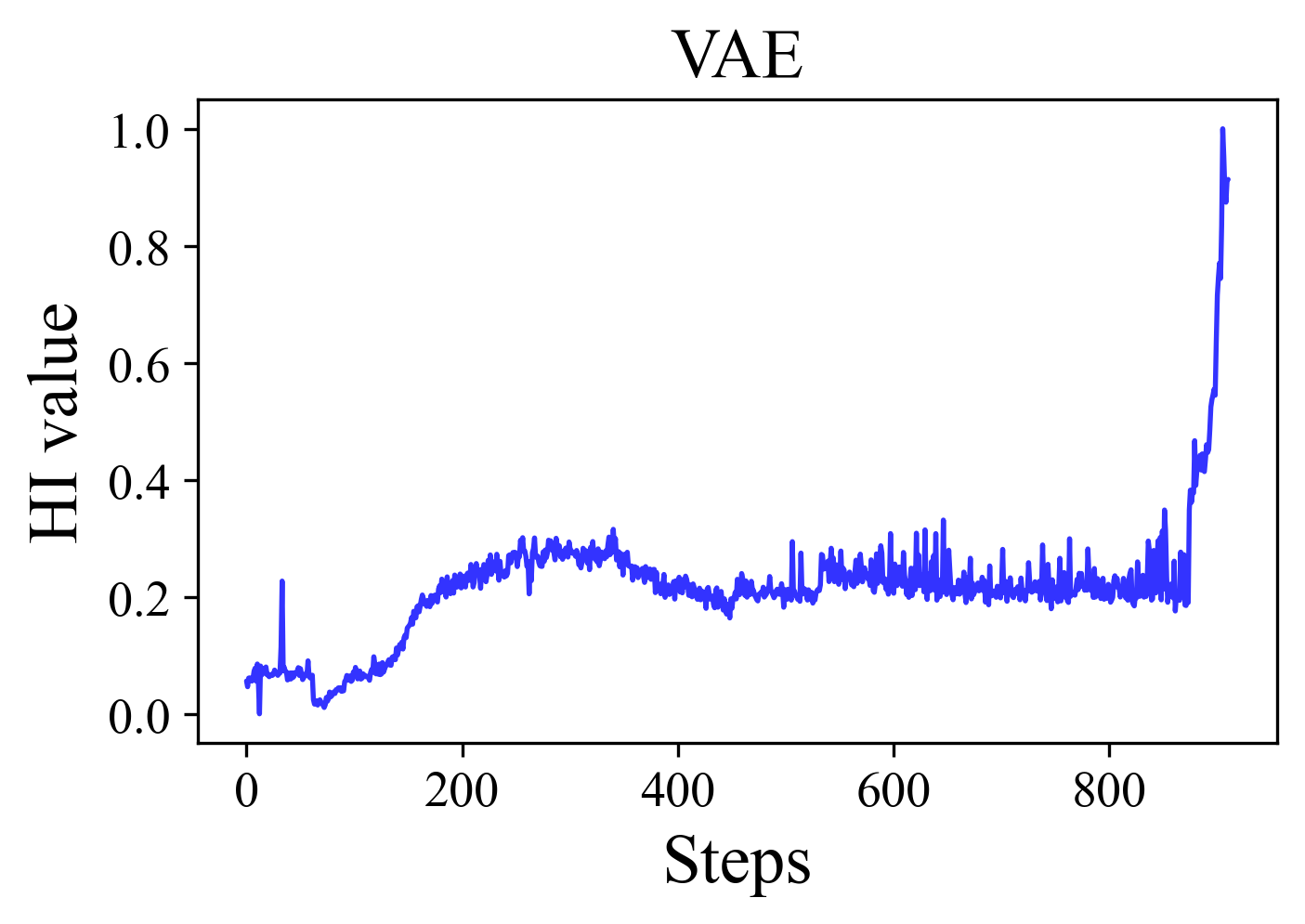}
\end{minipage}%
}
\subfigure[]{
\begin{minipage}[t]{0.18\linewidth}
    \centering
    \includegraphics[width=1.2in]{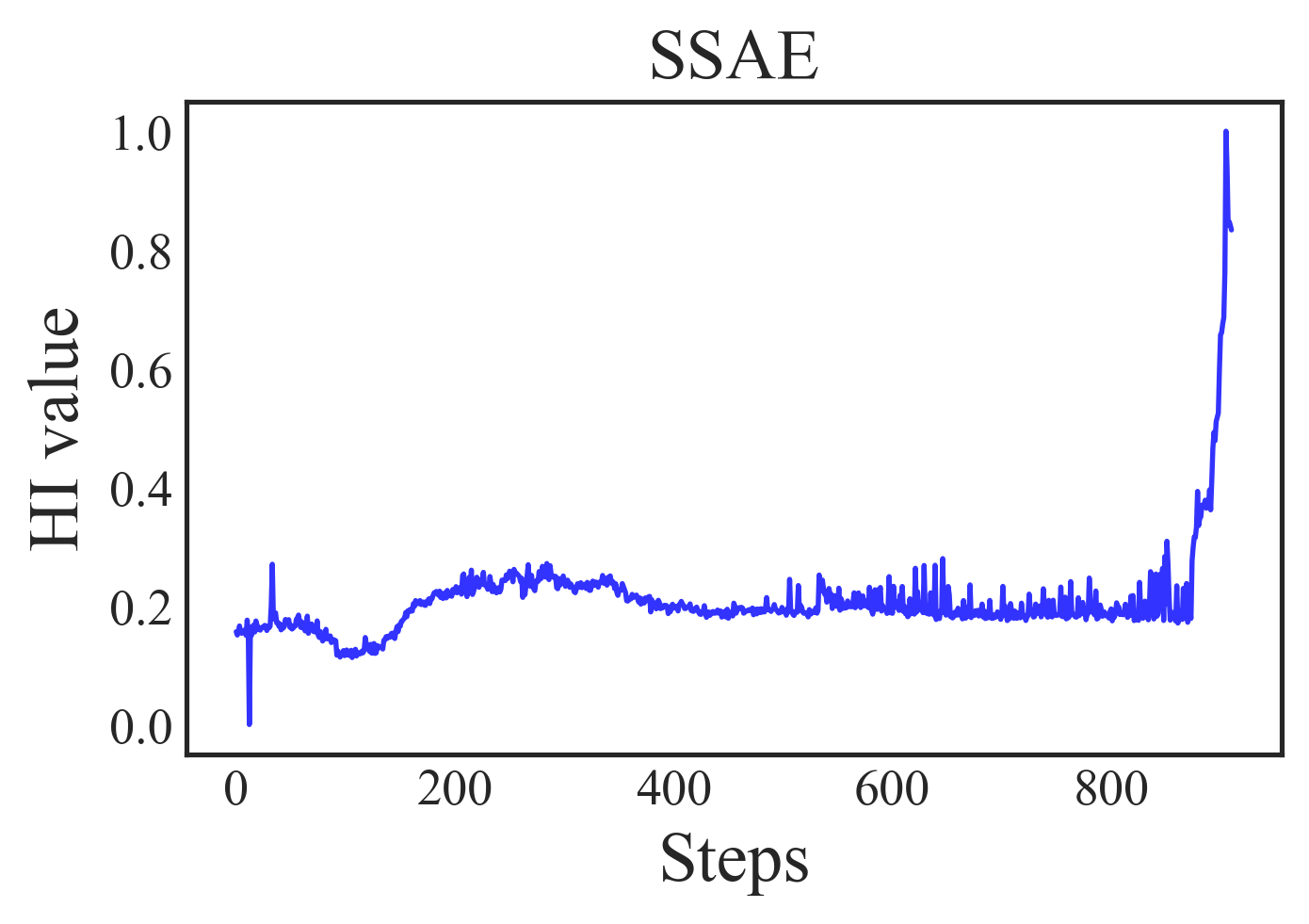}
\end{minipage}%
}
\subfigure[]{
\begin{minipage}[t]{0.18\linewidth}
    \centering
    \includegraphics[width=1.2in]{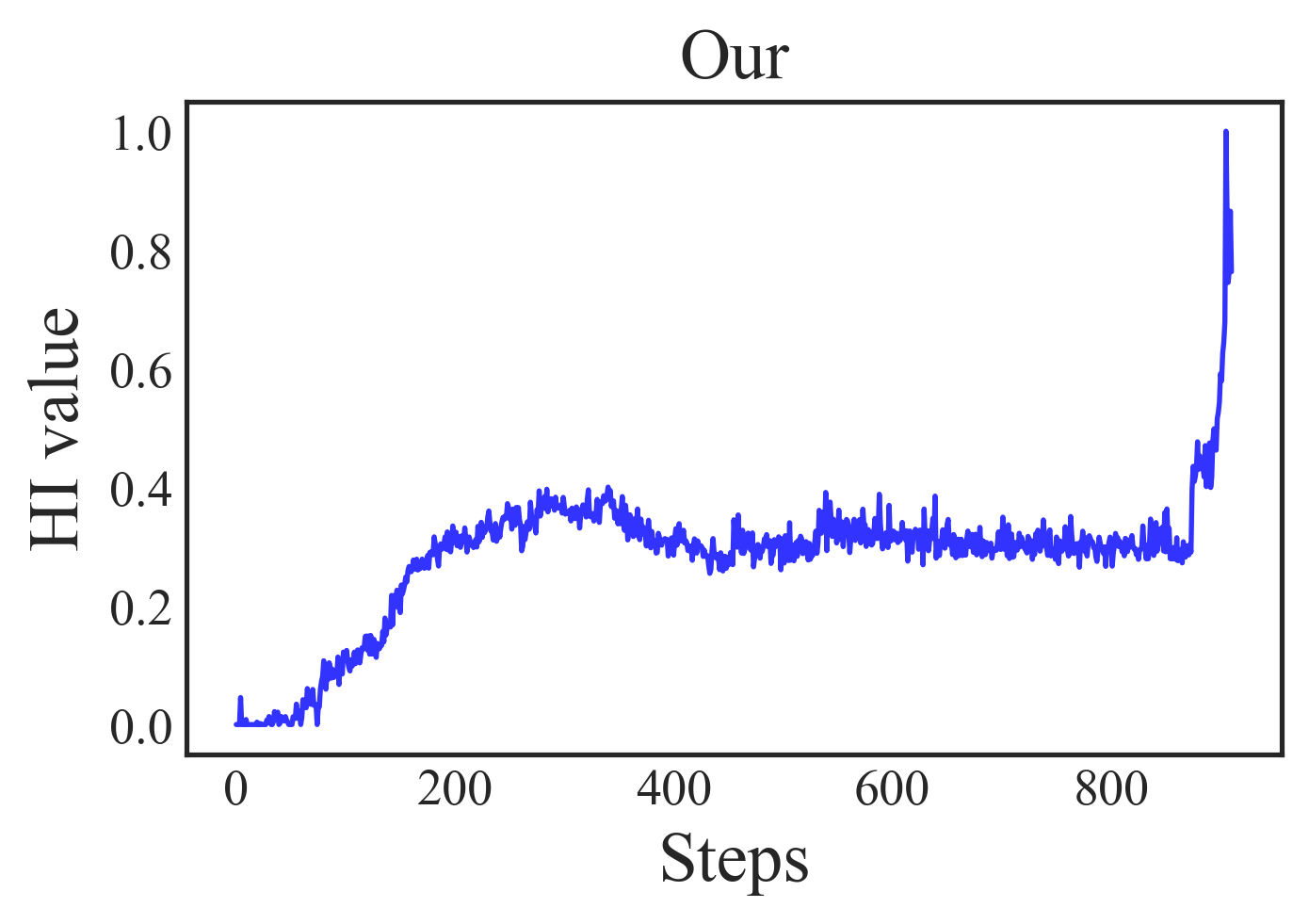}
\end{minipage}%
}
\centering
\vspace{-3mm}
\caption{\textcolor{black}{The constructed HIs for Task 2. (a) RMS, (b) P-Entropy, (c) ISOMAP, (d) KPCA, (e) CEEMDAN, (f) VAE, (g) SSAE, (h) Ours.}}
\label{task2_con}
\end{figure}

\begin{figure}[!h]
\vspace{-5mm}
\centering
\subfigure[]{
\begin{minipage}[t]{0.18\linewidth}
    \centering
    \includegraphics[width=1.2in]{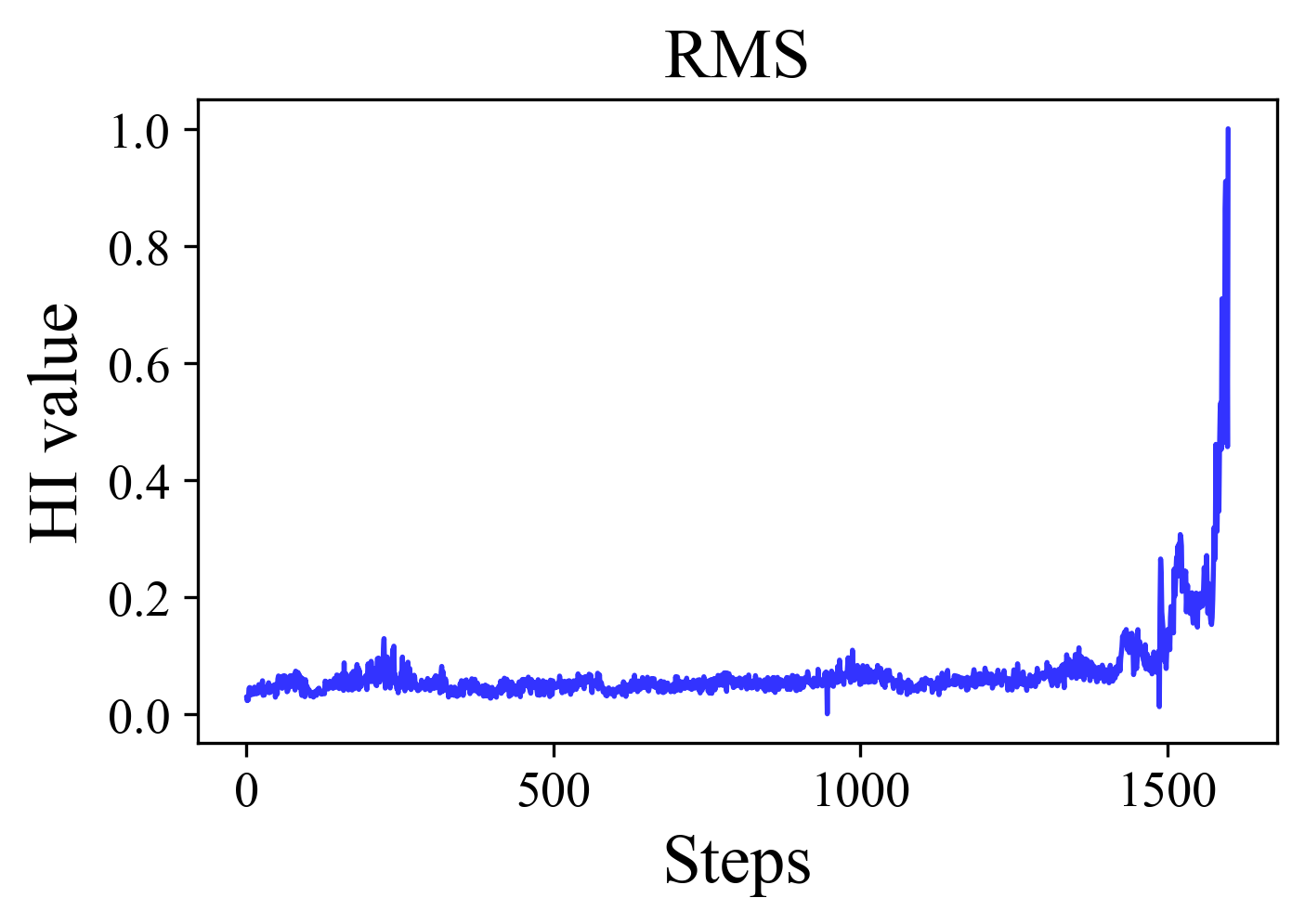}
\end{minipage}%
}%
\subfigure[]{
\begin{minipage}[t]{0.18\linewidth}
    \centering
    \includegraphics[width=1.2in]{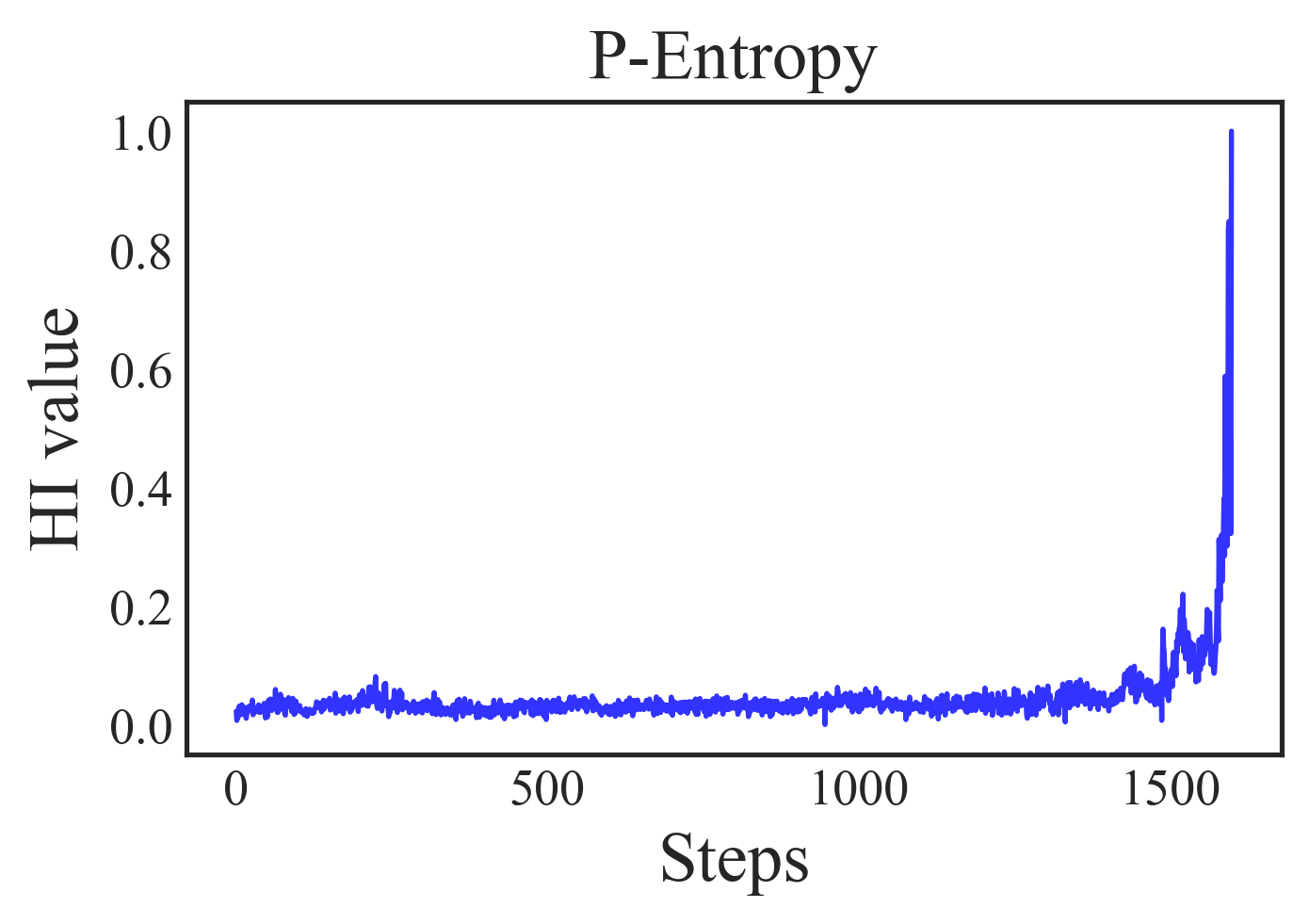}
\end{minipage}%
}
\subfigure[]{
\begin{minipage}[t]{0.18\linewidth}
    \centering
    \includegraphics[width=1.2in]{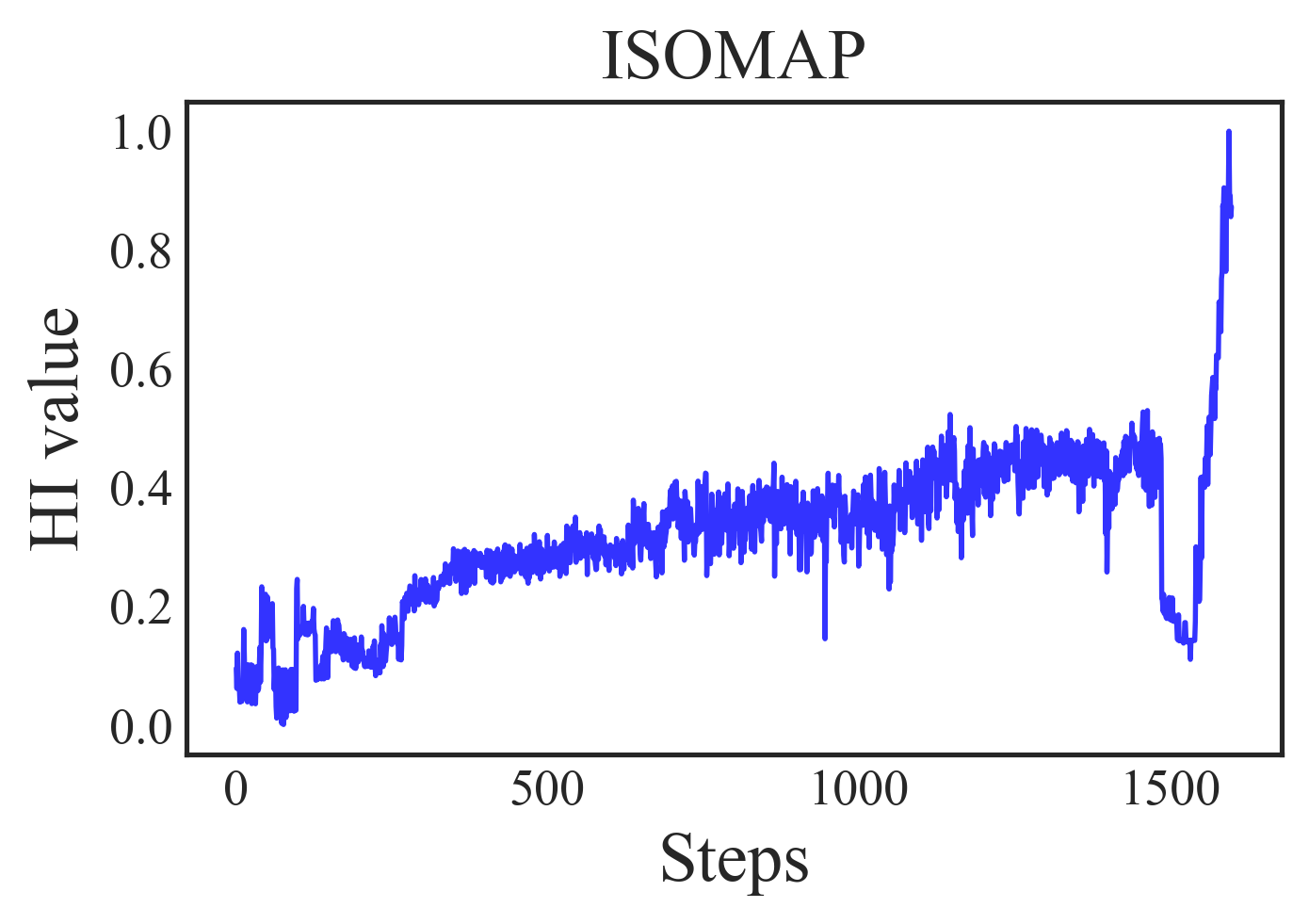}
\end{minipage}%
}
\subfigure[]{
\begin{minipage}[t]{0.18\linewidth}
    \centering
    \includegraphics[width=1.2in]{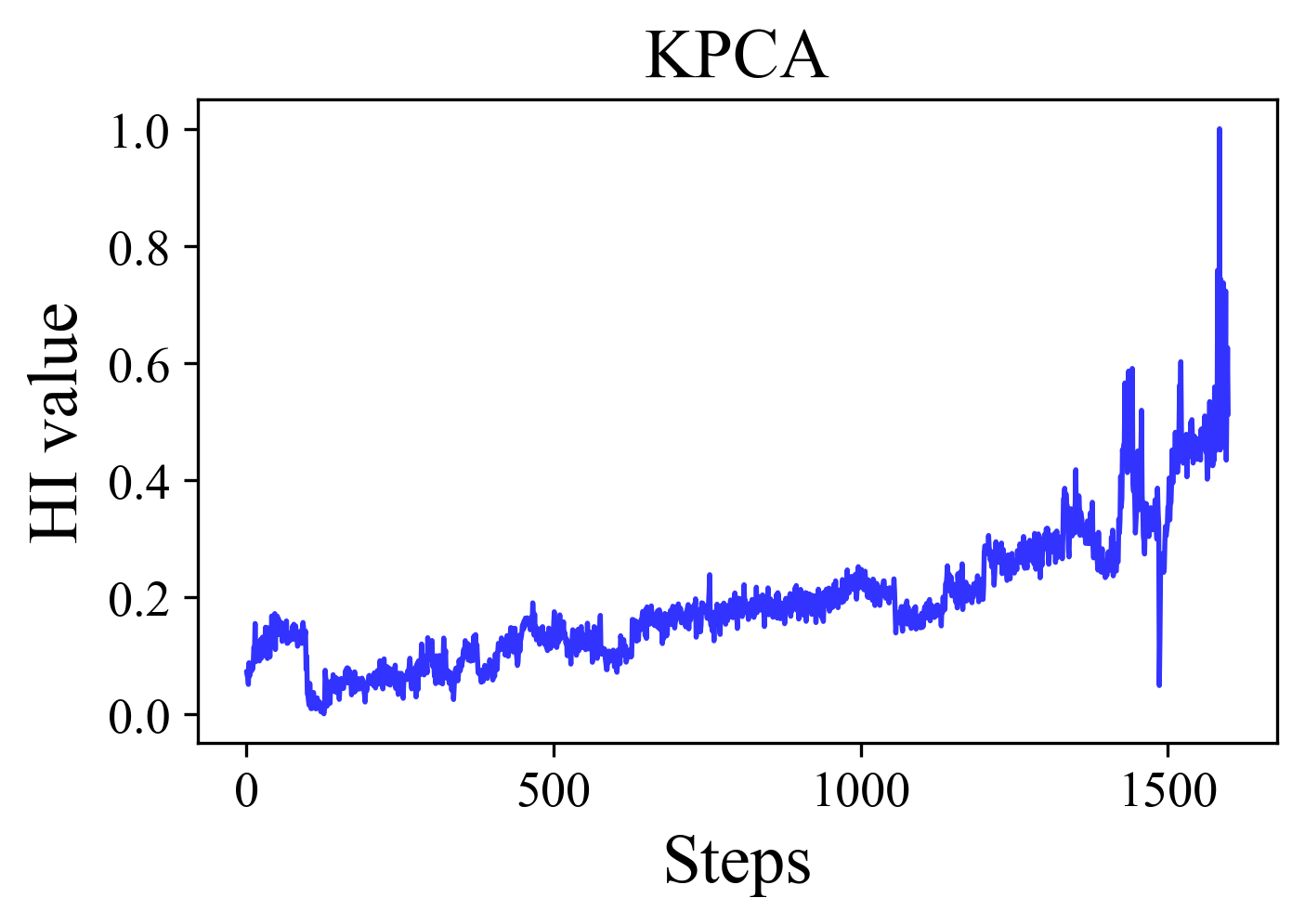}
\end{minipage}%
}\\
\vspace{-3mm}
\subfigure[]{
\begin{minipage}[t]{0.18\linewidth}
    \centering
    \includegraphics[width=1.2in]{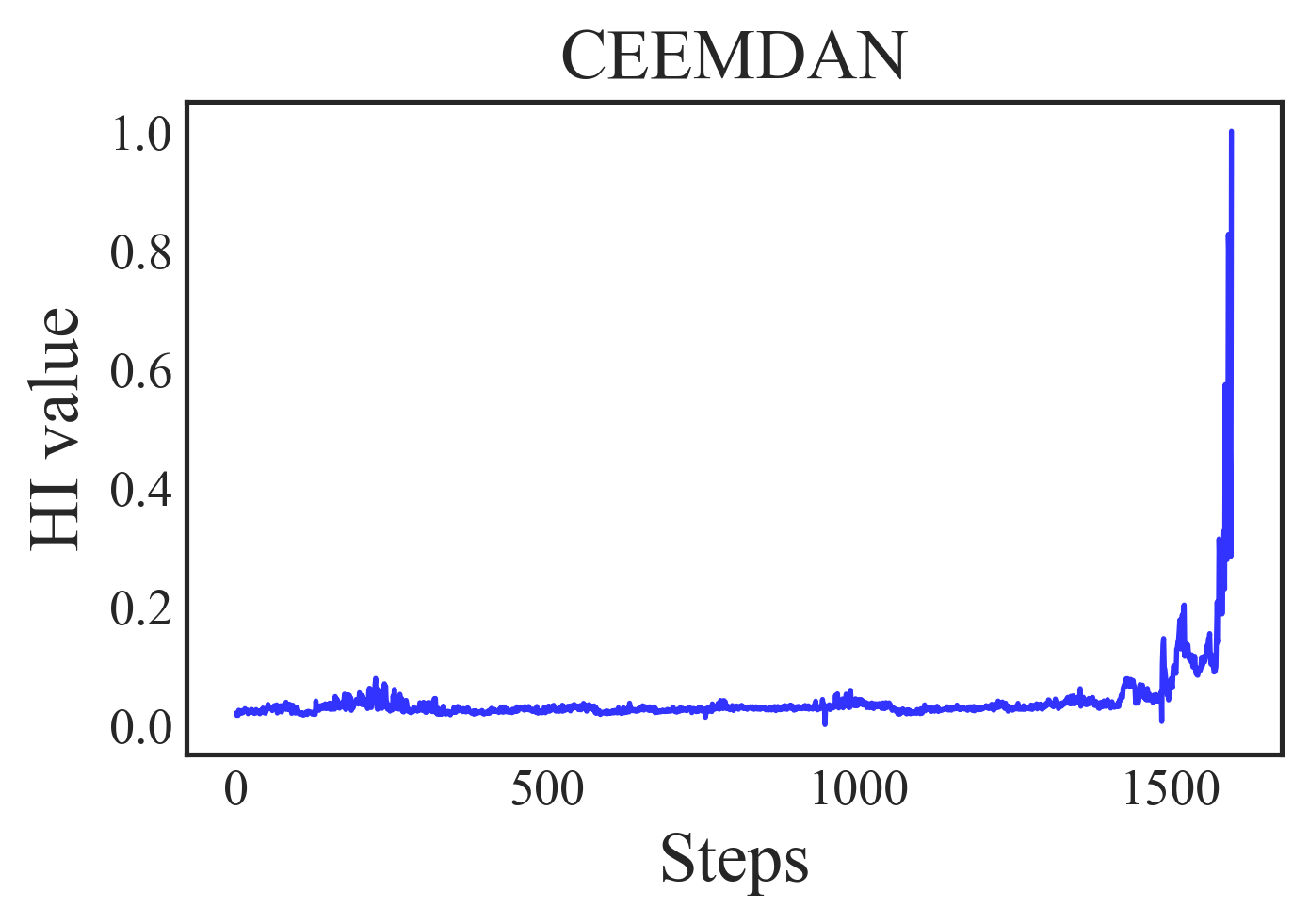}
\end{minipage}%
}%
\subfigure[]{
\begin{minipage}[t]{0.18\linewidth}
    \centering
    \includegraphics[width=1.2in]{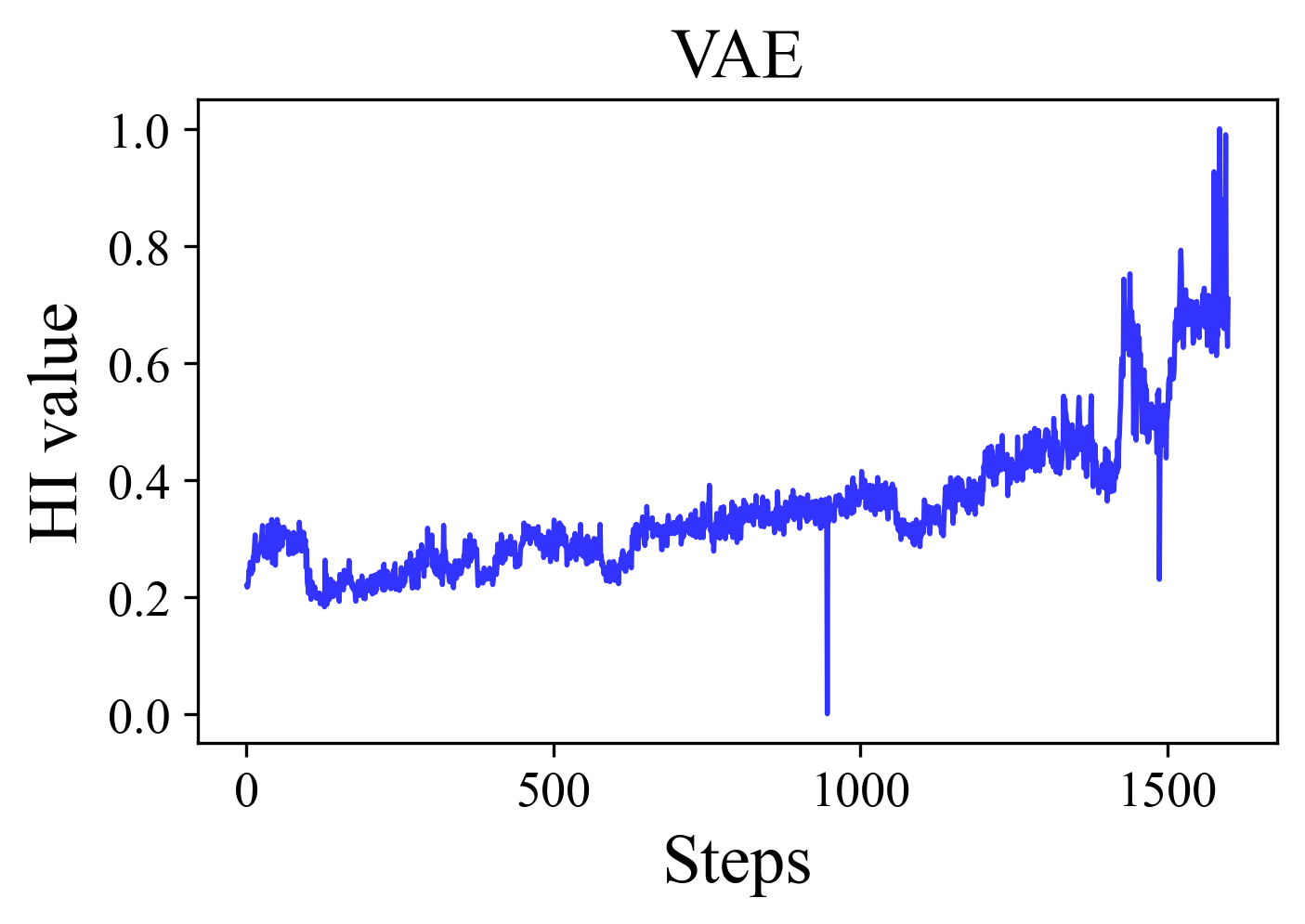}
\end{minipage}%
}
\subfigure[]{
\begin{minipage}[t]{0.18\linewidth}
    \centering
    \includegraphics[width=1.2in]{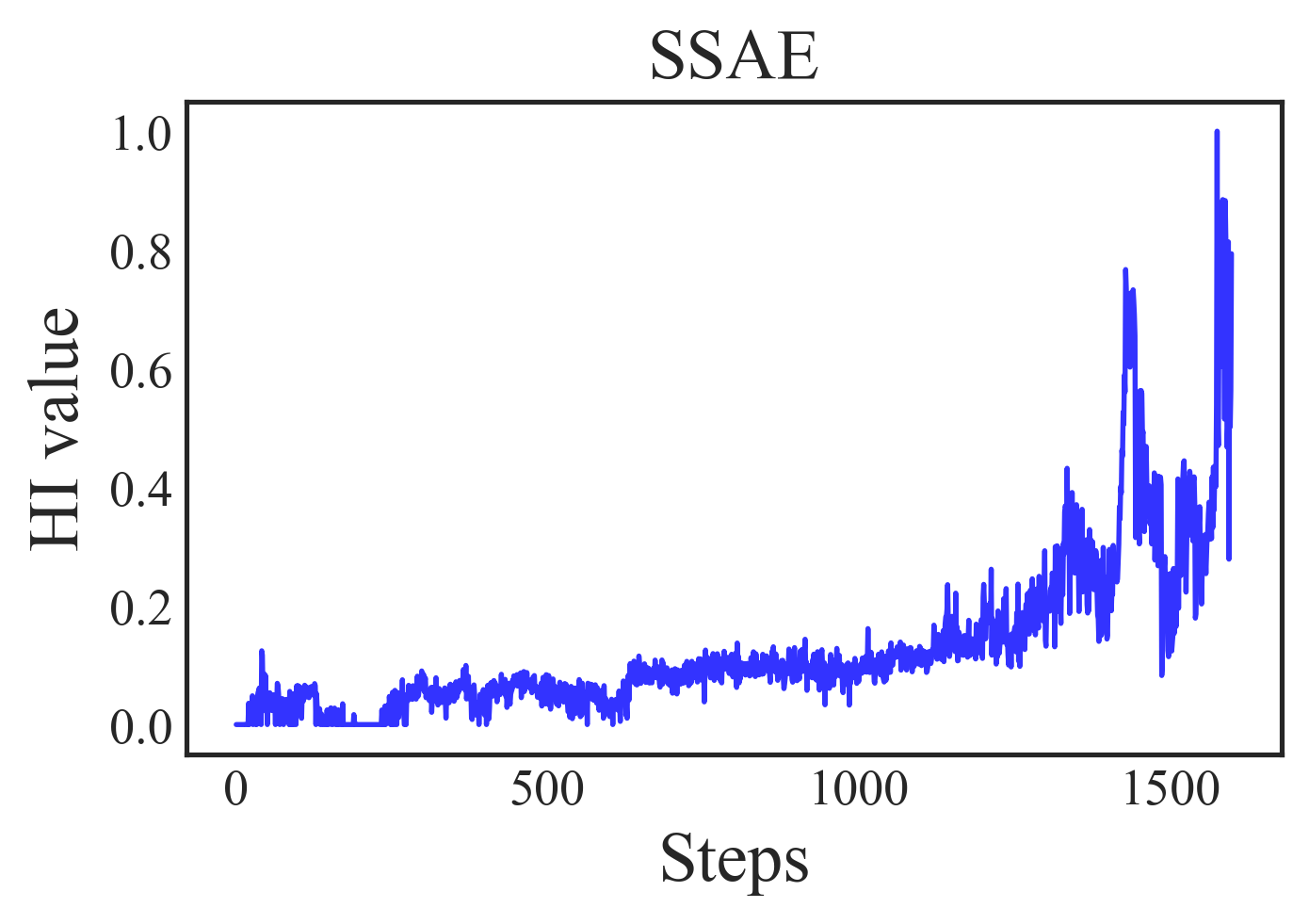}
\end{minipage}%
}
\subfigure[]{
\begin{minipage}[t]{0.18\linewidth}
    \centering
    \includegraphics[width=1.2in]{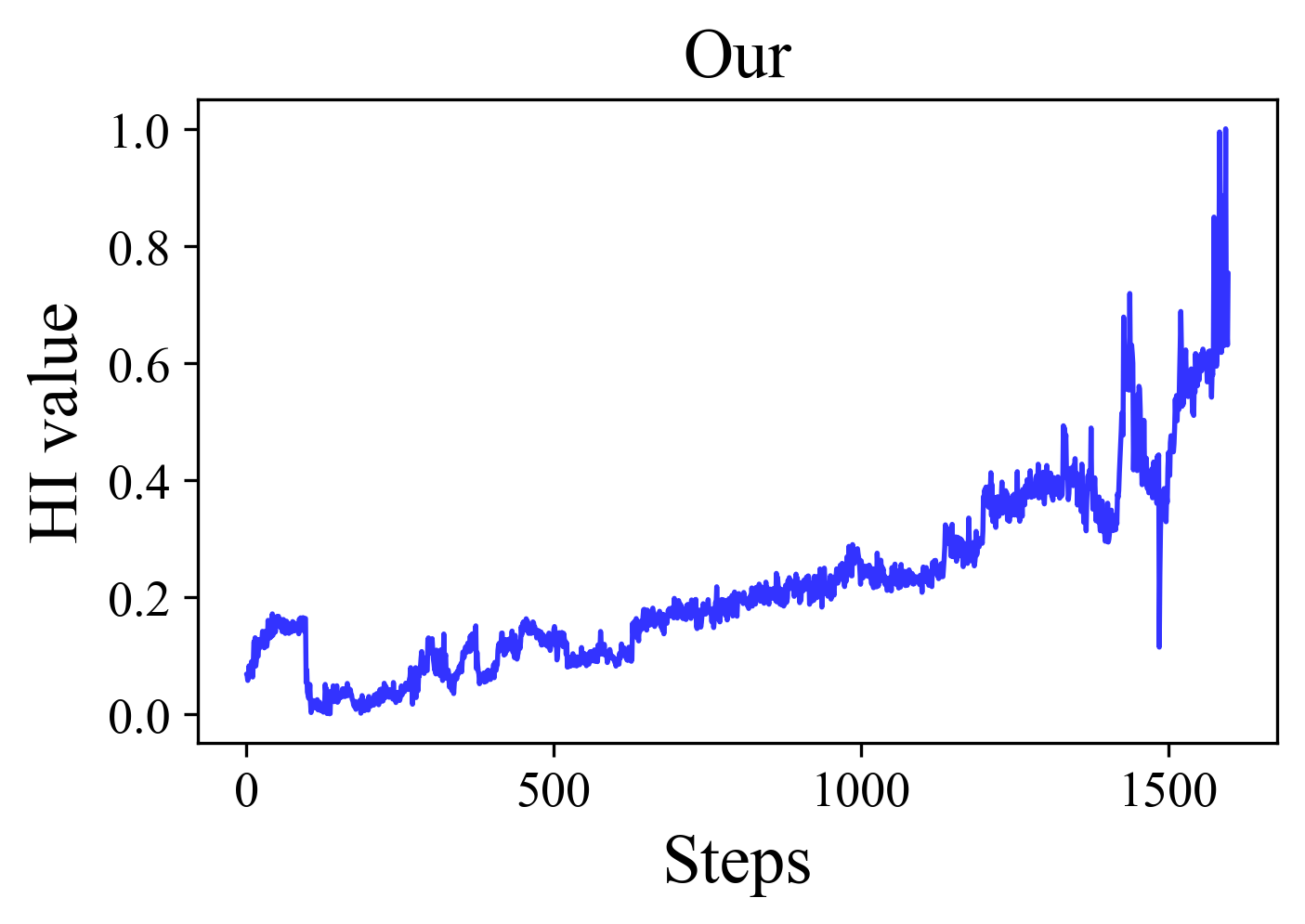}
\end{minipage}%
}
\centering
\vspace{-3mm}
\caption{\textcolor{black}{The constructed HIs for Task 3. (a) RMS, (b) P-Entropy, (c) ISOMAP, (d) KPCA, (e) CEEMDAN, (f) VAE, (g) SSAE, (h) Ours.}}
\label{task3_con}
\end{figure}

\begin{figure}[!h]
\centering
\subfigure[]{
\begin{minipage}[t]{0.32\linewidth}
    \centering
    \includegraphics[width=2.1in]{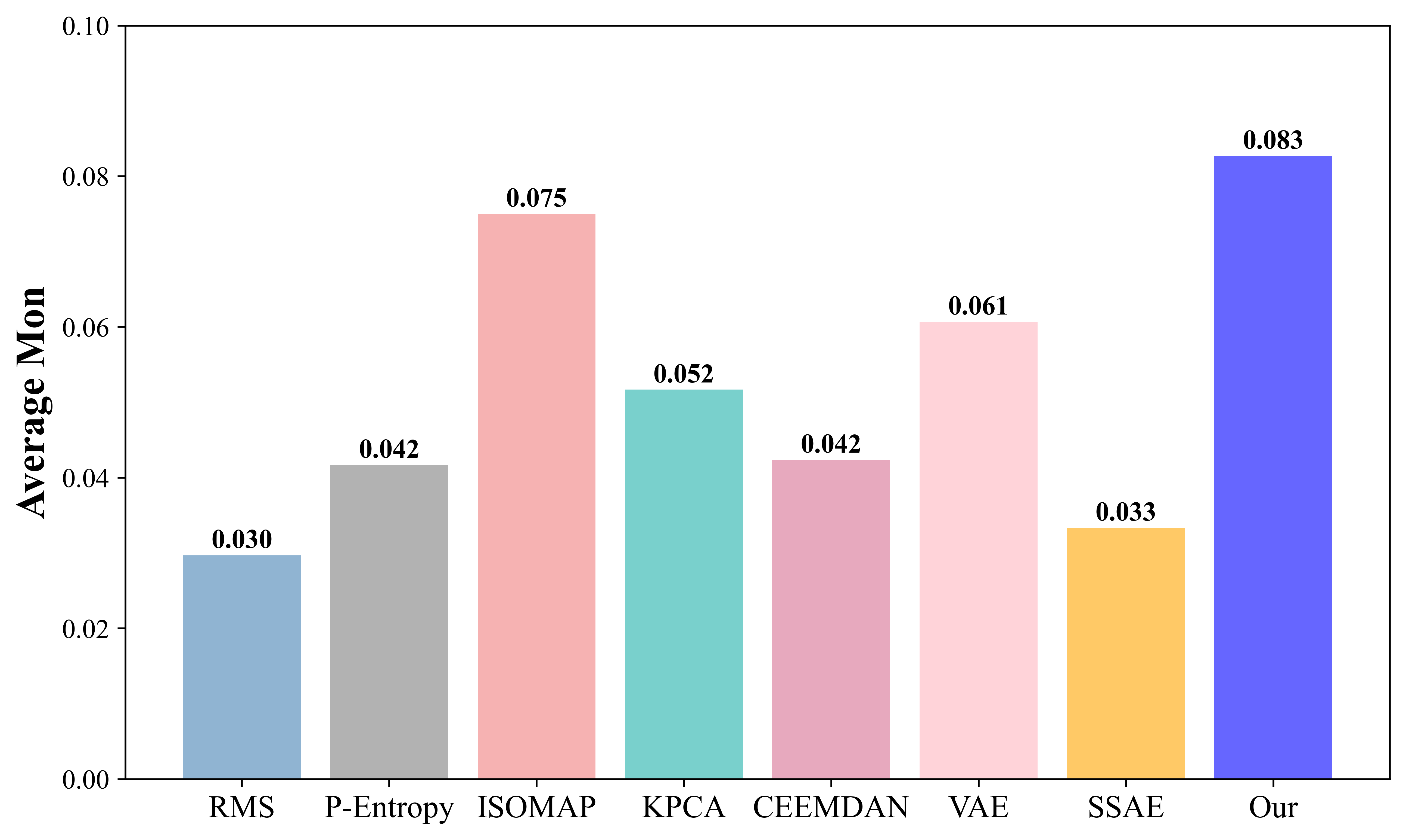}
\end{minipage}%
}
\subfigure[]{
\begin{minipage}[t]{0.32\linewidth}
    \centering
    \includegraphics[width=2.1in]{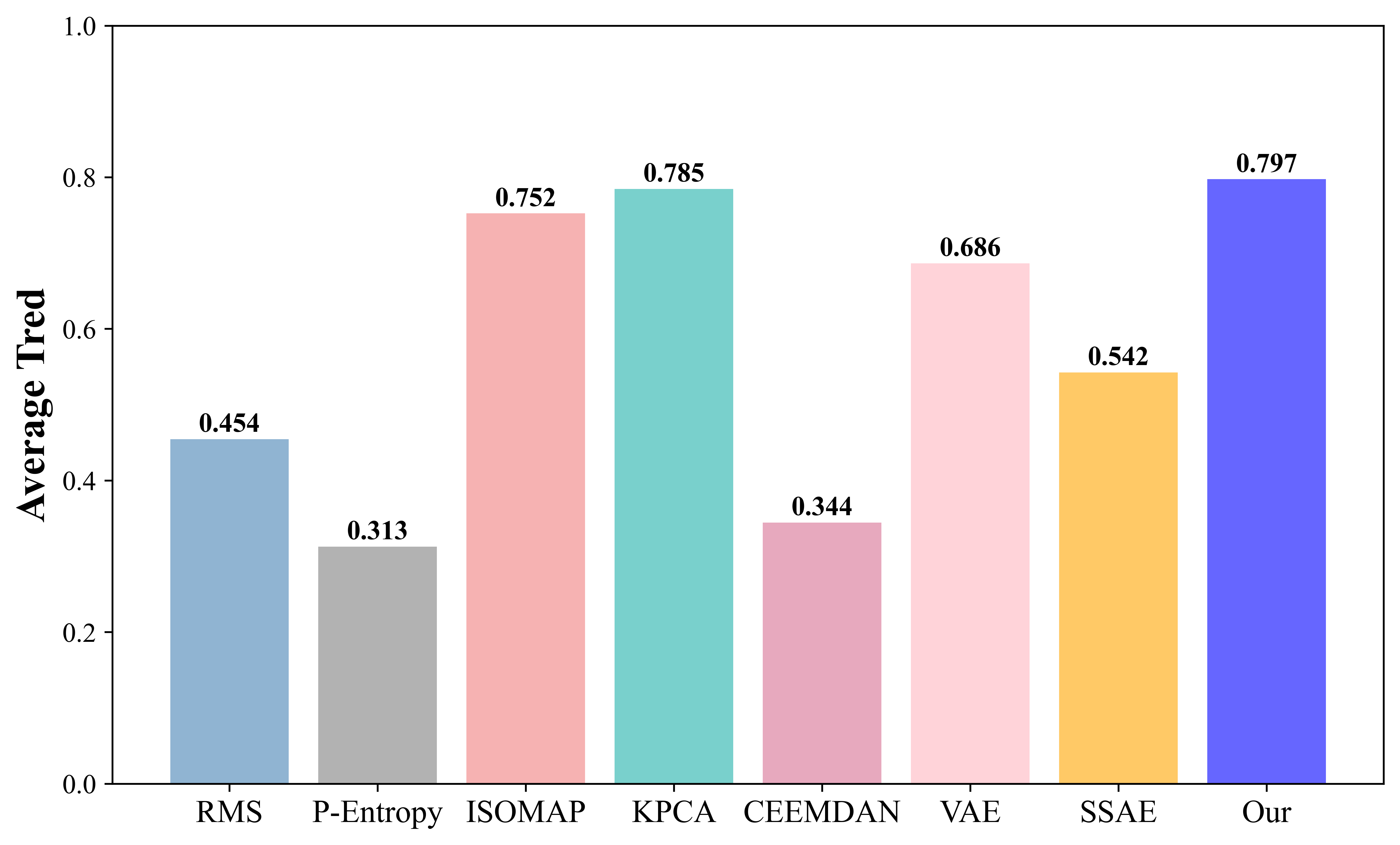}
\end{minipage}%
}\\
\vspace{-3mm}
\subfigure[]{
\begin{minipage}[t]{0.32\linewidth}
    \centering
    \includegraphics[width=2.1in]{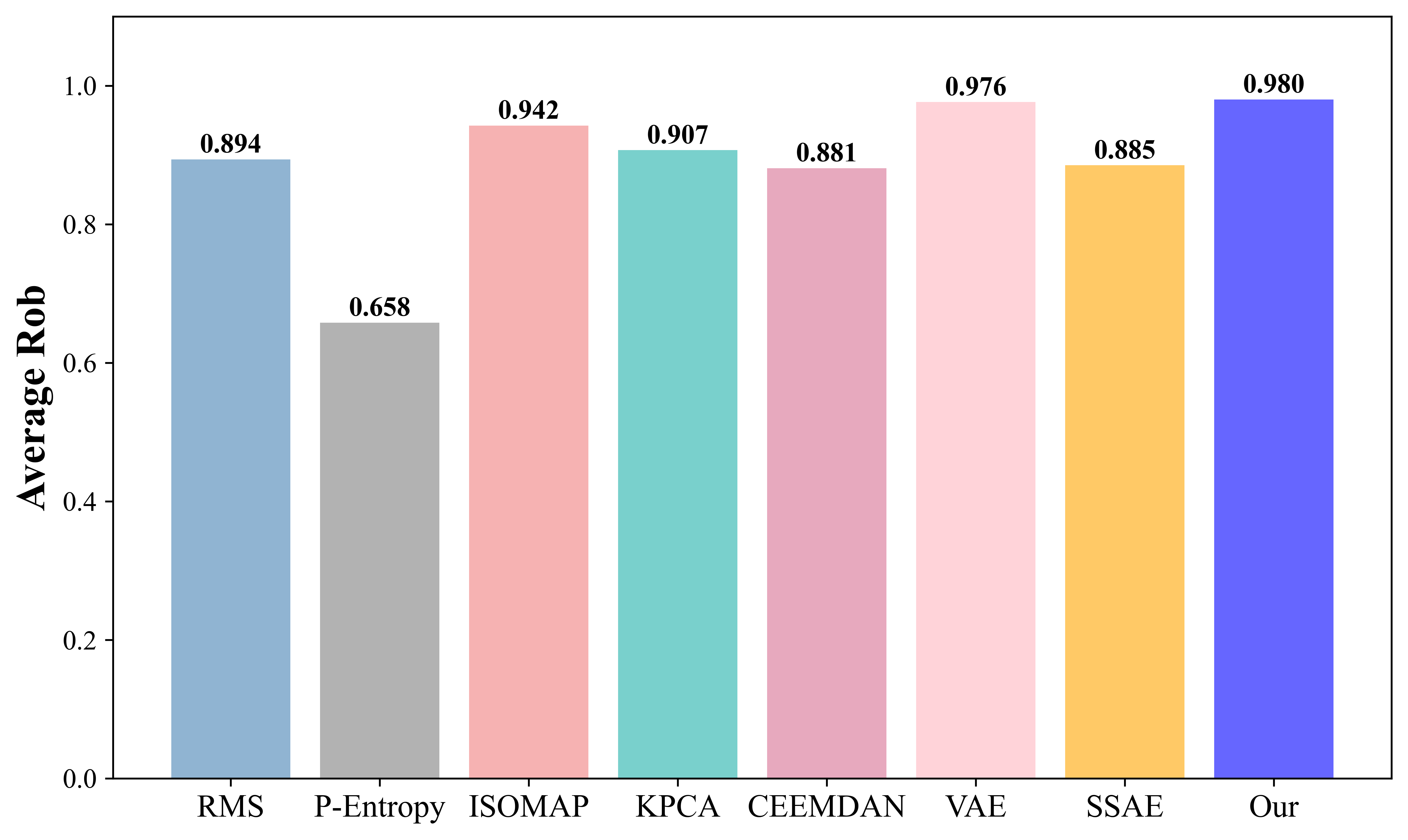}
\end{minipage}%
}
\subfigure[]{
\begin{minipage}[t]{0.32\linewidth}
    \centering
    \includegraphics[width=2.1in]{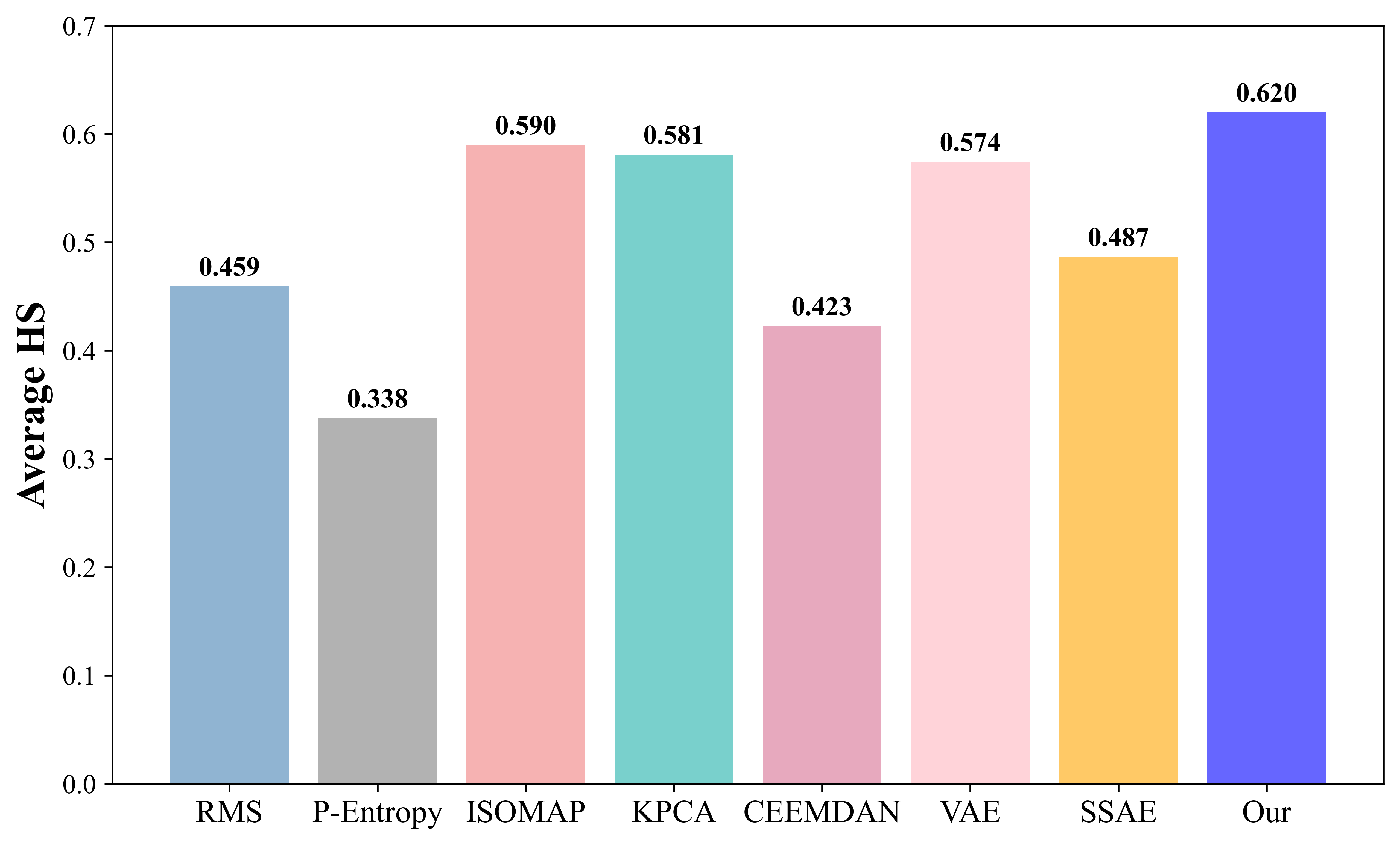}
\end{minipage}%
}
\centering
\vspace{-3mm}
\caption{\textcolor{black}{The averaged performance of HI construction obtained by different methods across all tasks. (a) Mon, (b) Tred, (c) Rob, (d) HS.}}
\label{con_avg}
\end{figure}

\textcolor{black}{Table. \ref{table3} details the comparison results, and Fig. \ref{task1_con} to Fig. \ref{task3_con} visualize the HIs constructed by different methods.} The proposed method achieves the best in almost all metrics among all construction tasks. For Tasks 1 and 2, the proposed method yields the best Mon, Rob, and HS. Although KPCA achieves a slightly higher Tred compared to our method, the HI constructed by KPCA does not accurately reflect the degradation process. This discrepancy is particularly evident after the 1146th step (shown in Fig. \ref{task1_con} (d)), where the HI exhibits a linear increase accompanied by significant fluctuations. \textcolor{black}{Conversely, as shown in Fig. \ref{task1_con} and \ref{task2_con}, the proposed method effectively represents bearing degradation. This indicates that the proposed HI exhibits superior sensitivity to the incipient fault and representability to the degradation trend compared to the comparison methods.} For Task 3, as the distribution of training data is different from that of the testing data, all HIs show more fluctuation than Task 1 and 2, which is shown in Fig. \ref{task3_con}. Obviously, the proposed dynamic HI outperforms other approaches in this task, especially in revealing the monotonicity and trendability of the bearing degradation. This proves the generability of the proposed method on different machines with unseen working conditions. \textcolor{black}{Fig. \ref{con_avg} illustrates the average performance of HIs constructed using various comparison methods across all tasks. Our method consistently achieves the highest scores in all metrics, particularly excelling in Mon and Tred. This significant improvement over other methods demonstrates that our approach effectively constructs an HI that aligns closely with the degradation process of bearings, exhibiting high monotonicity, trendability, and robustness.}

\subsubsection{Ablation Studies and Discussion}
\label{sec3.2.4}
\textcolor{black}{To evaluate the effectiveness of different modules in the proposed framework, comprehensive ablation studies are conducted in this section. Five ablation cases are constructed for comparison under the same experimental settings and parameter selections as the proposed method.}

\begin{enumerate}
    \item No SkipAE: This method eliminates the skip connections from the proposed SkipAE, utilizing a general convolutional neural network (CNN) autoencoder for the degradation feature learning module.
    \item \textcolor{black}{No inner HI-prediction block: This method removes the inner HI-prediction block, where HI is mainly constrained by reconstruction loss and monotonic loss.}
    \item \textcolor{black}{No SkipAE and HI-prediction block: This method removes the skip connection of the proposed SkipAE as well as the inner HI-prediction block. A general CNN autoencoder is leveraged for degradation feature learning and the HI is constructed through the HI-generating module without the HI-level prediction loss.}
    \item \textcolor{black}{No HI-generating module: This method replaces the HI-generating module with the dimension reduction technique, mapping degradation features into a one-dimensional HI through principal component analysis (PCA). Additionally, as the inner HI-prediction block is integrated into the HI-generating module, this method also removes the inner HI-prediction block.}
    \item \textcolor{black}{No SkipAE and HI-generating module: This method eliminates the skip connection from the proposed SkipAE and substitutes the HI-generating module with PCA. The degradation features are learned through a general CNN autoencoder, and HI is subsequently constructed using PCA.}
\end{enumerate}

\begin{figure}[!b]
\centering
\subfigure[]{
\begin{minipage}[t]{0.28\linewidth}
    \centering
    \includegraphics[width=1.8in]{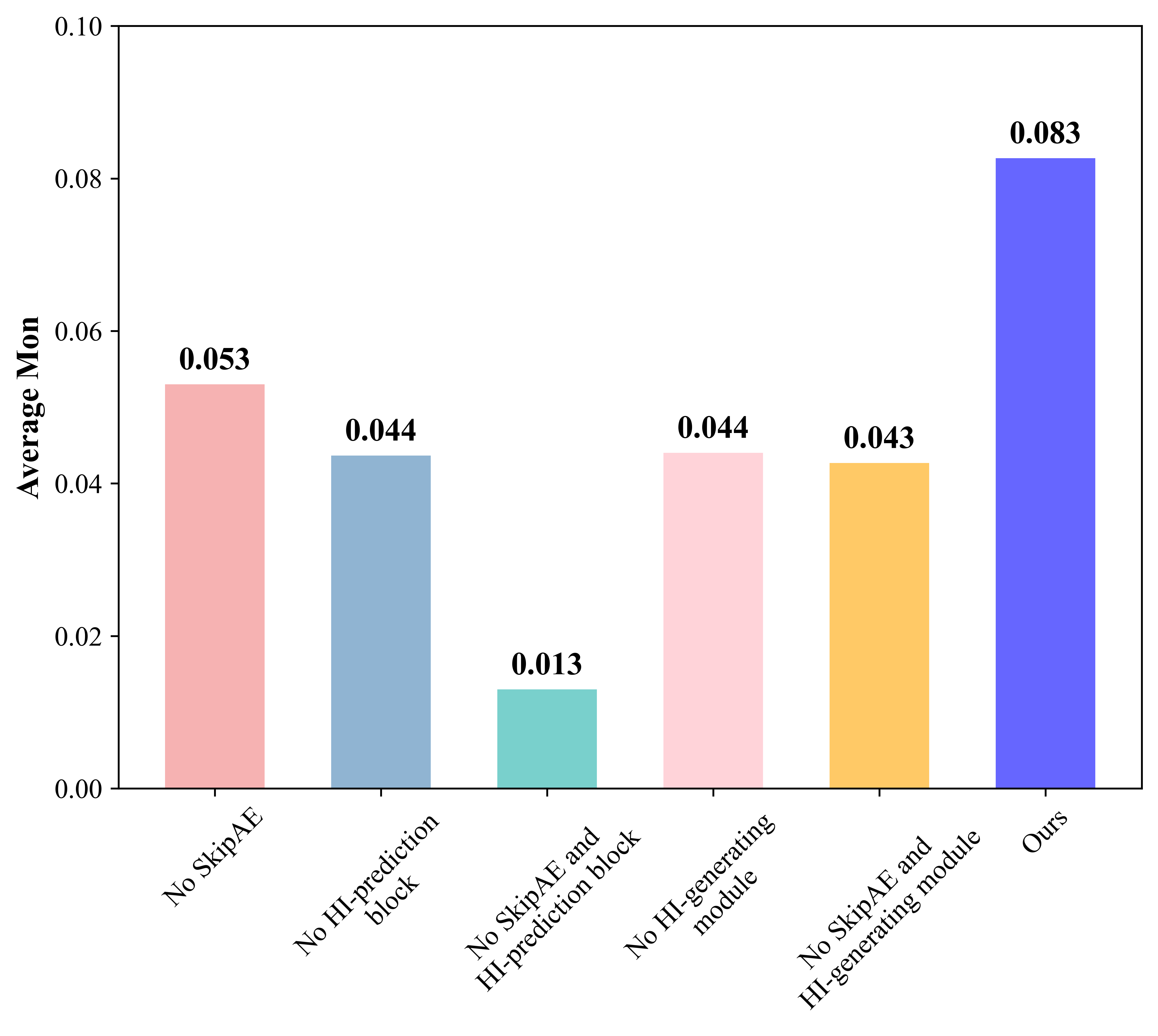}
\end{minipage}%
}
\subfigure[]{
\begin{minipage}[t]{0.28\linewidth}
    \centering
    \includegraphics[width=1.8in]{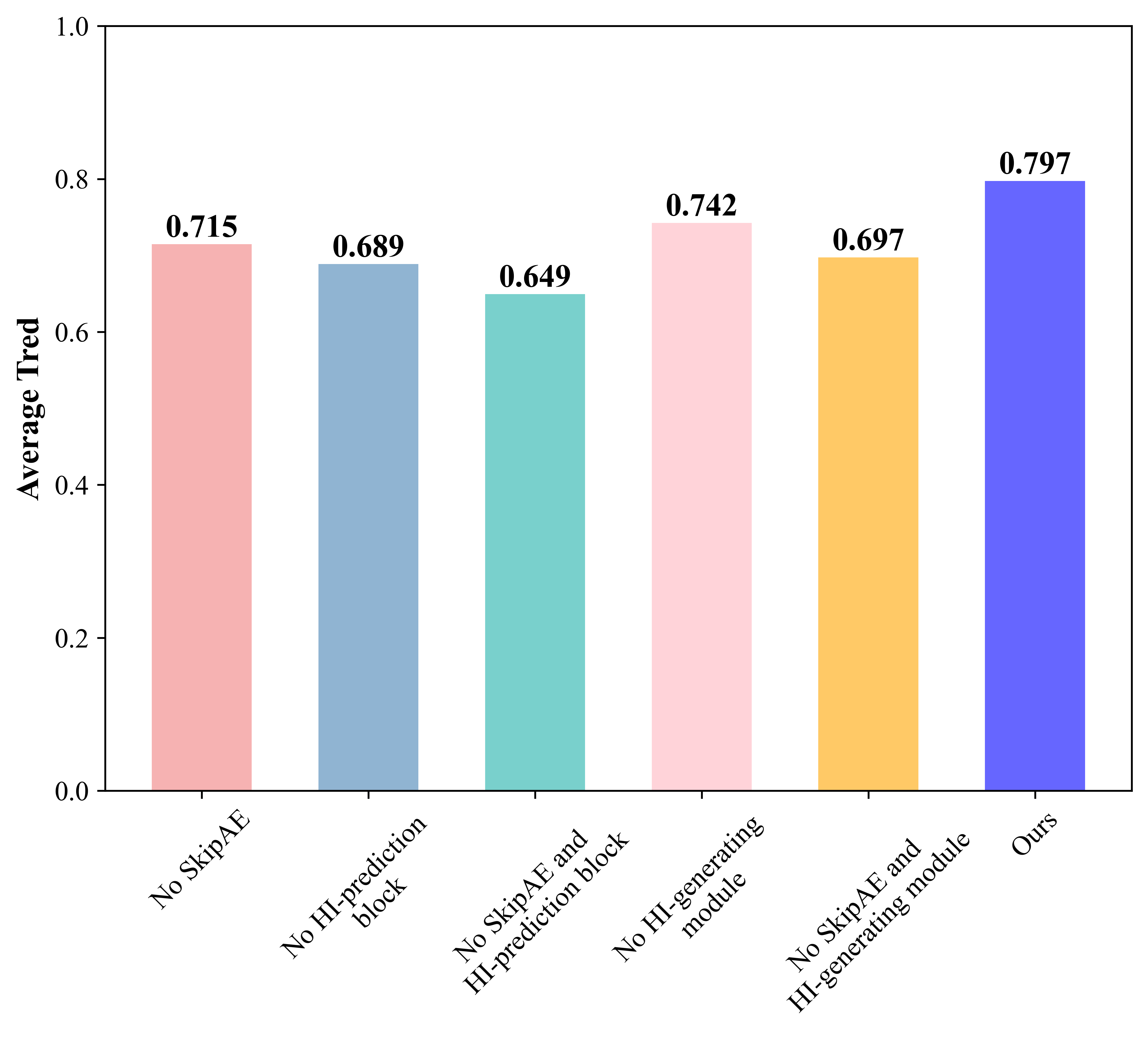}
\end{minipage}%
}\\
\vspace{-3mm}
\subfigure[]{
\begin{minipage}[t]{0.28\linewidth}
    \centering
    \includegraphics[width=1.8in]{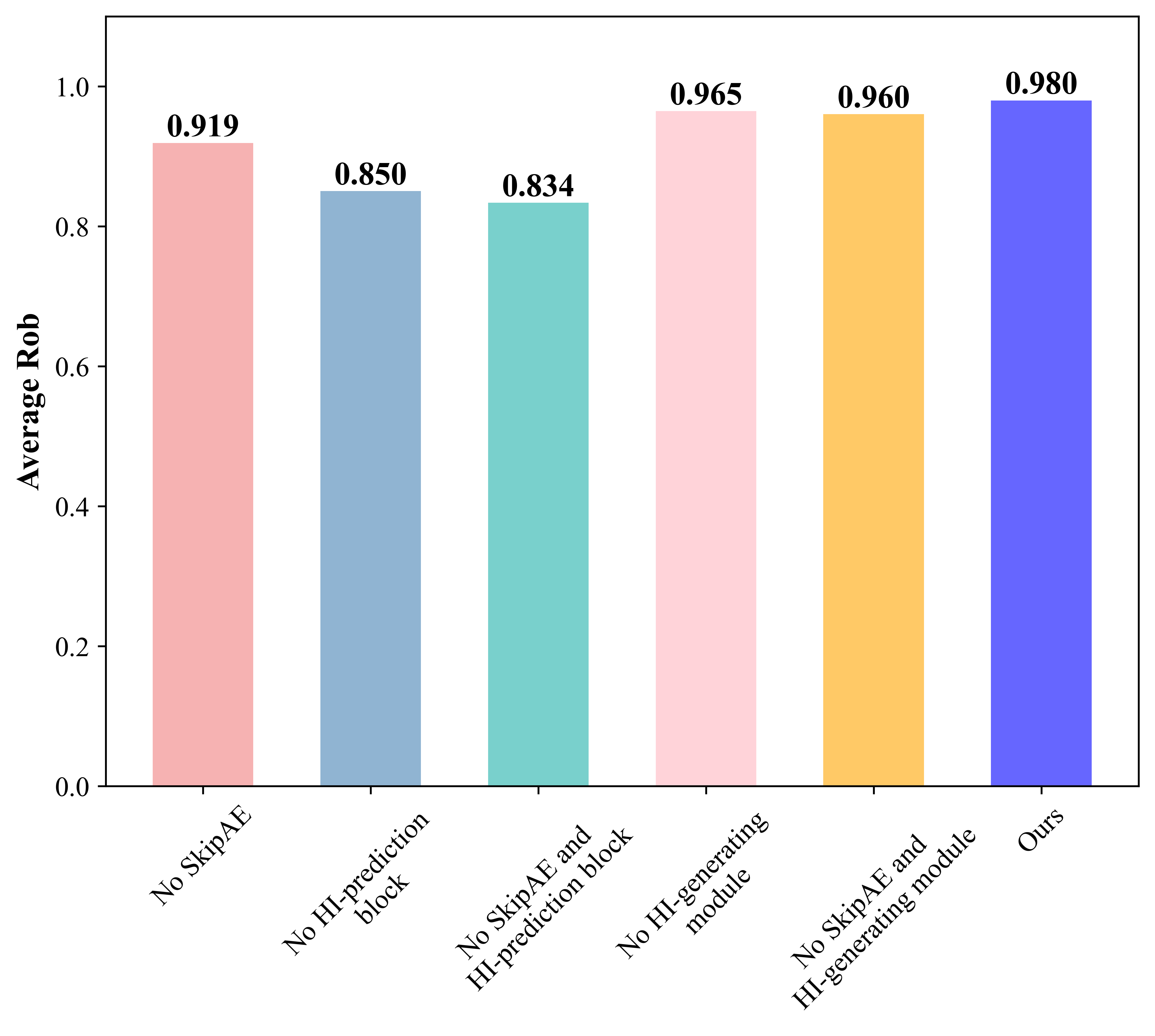}
\end{minipage}%
}
\subfigure[]{
\begin{minipage}[t]{0.28\linewidth}
    \centering
    \includegraphics[width=1.8in]{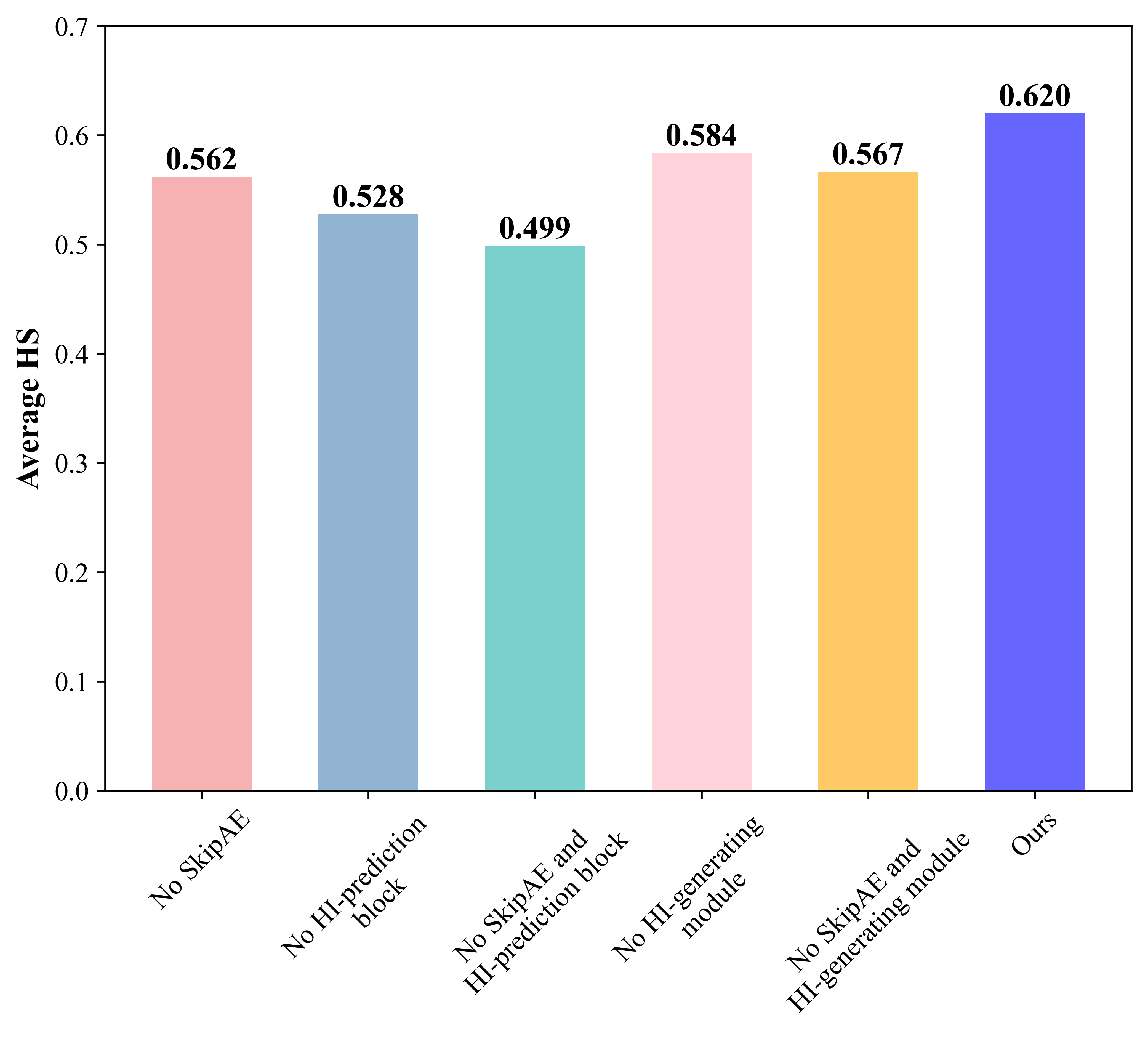}
\end{minipage}%
}
\centering
\vspace{-3mm}
\caption{\textcolor{black}{The average values of various metrics from ablation studies on HI construction across all tasks. (a) Mon, (b) Tred, (c) Rob, (d) HS.}}
\label{ablation_avg}
\end{figure}

Table. \ref{tableAblationinHIconstruction} reports the results of the ablation studies across all research tasks. Overall, these results reveal a decline in performance across nearly all metrics when specific components are removed. Specifically, the effectiveness of the proposed SkipAE is demonstrated by comparing it with the No SkipAE method. After removing the skip connections, the Tred declines across all three tasks, indicating that skip connections contribute to learning more representative degradation features, thereby enhancing the trendability of the HI. Although the removal of skip connections led to slight improvements in Mon and Rob in Task 3 and Task 2, there were significant declines in other tasks. This resulted in an overall decrease in the method's average performance, as shown in Fig. \ref{ablation_avg} (a) and (c). Consequently, skip connections can comprehensively enhance the performance of the HI.

\textcolor{black}{The effectiveness of the inner HI-prediction block is demonstrated through a comparison with the No HI-prediction block method. As shown in Table. \ref{tableAblationinHIconstruction}, all metrics degrade across all tasks when this block is removed, indicating that the inner HI-prediction block significantly enhances the HI's ability for bearing degradation modeling. This is because the inner HI-prediction block ensures HI-level temporal dependence, which helps the constructed HI to contain as much dynamic information as possible, thereby enhancing its representativeness. Consequently, the HS improves from 0.528 to 0.620 when the inner HI-prediction block is integrated into the method.}

\begin{table*}[!t]
\centering
\caption{\textcolor{black}{Results of the Ablation Study in the Construction of Bearing HI.}}
\setlength\tabcolsep{3.5pt}
\label{tableAblationinHIconstruction}
\scalebox{0.8}{
\begin{tabularx}{1.1\textwidth}{ccccccccccccc}
  \hline
  \multirowcell{2}{Method} & \multicolumn{4}{c}{Task 1} & \multicolumn{4}{c}{Task 2} & \multicolumn{4}{c}{Task 3}\\
  \quad & Mon & Tred & Rob & HS & Mon & Tred & Rob & HS & Mon & Tred & Rob & HS\\
  \hline
  No SkipAE & 0.028 & 0.775 & 0.868 & 0.556 & 0.048 & 0.473 & $\mathbf{0.991}$ & 0.504 & $\mathbf{0.083}$ & 0.896 & 0.898 & 0.626 \\
  \textcolor{black}{No inner HI-prediction block} & 0.012 & 0.693 & 0.743 & 0.483 & 0.048 & 0.495 & 0.840 & 0.461 & 0.071 & 0.878 & 0.968 & 0.639 \\
  \textcolor{black}{No SkipAE and HI-prediction block} & \textcolor{black}{0.014} & \textcolor{black}{0.675} & \textcolor{black}{0.722} & \textcolor{black}{0.470} & \textcolor{black}{0.013} & \textcolor{black}{0.413} & \textcolor{black}{0.825} & \textcolor{black}{0.417} & \textcolor{black}{0.012} & \textcolor{black}{0.860} & \textcolor{black}{0.954} & \textcolor{black}{0.609}
\\
  \textcolor{black}{No HI-generating module} & \textcolor{black}{0.027} & \textcolor{black}{0.778} & \textcolor{black}{0.958} & \textcolor{black}{0.588} & \textcolor{black}{0.052} & \textcolor{black}{0.624} & \textcolor{black}{0.966} & \textcolor{black}{0.547} & \textcolor{black}{0.053} & \textcolor{black}{0.825} & \textcolor{black}{0.970} & \textcolor{black}{0.616}
 \\
  \textcolor{black}{No SkipAE and HI-generating module} & \textcolor{black}{0.029} & \textcolor{black}{0.743} & \textcolor{black}{0.936} & \textcolor{black}{0.569} & \textcolor{black}{0.054} & \textcolor{black}{0.543} & \textcolor{black}{0.976} & \textcolor{black}{0.524} & \textcolor{black}{0.045} & \textcolor{black}{0.806} & \textcolor{black}{0.969} & \textcolor{black}{0.607}
 \\
  Ours & $\mathbf{0.069}$ & $\mathbf{0.854}$ & $\mathbf{0.983}$ & $\mathbf{0.635}$ & $\mathbf{0.105}$ & $\mathbf{0.639}$ & 0.984 & $\mathbf{0.576}$ & 0.074 & $\mathbf{0.899}$ & $\mathbf{0.973}$ & $\mathbf{0.649}$\\
  \hline
\end{tabularx}}
\end{table*}

\textcolor{black}{The joint effectiveness of the SkipAE and the inner HI-prediction block is evaluated as well. As shown in Fig. \ref{ablation_avg}, removing these two components results in significant performance degradation, particularly in the averaged value of Mon, which drops from 0.083 to 0.013. Additionally, compared to methods that only remove the SkipAE (No SkipAE) or the inner HI-prediction block (No HI-prediction block), removing both components leads to an overall performance decline. As shown in Fig. \ref{ablation_avg} (d), the HS drops by 0.063 (from 0.562 to 0.499) and 0.029 (from 0.528 to 0.499), respectively. These results indicate that our proposed method effectively integrates the SkipAE and the inner HI-prediction block, enabling them to collaboratively construct a more comprehensive HI.
}

\textcolor{black}{The effectiveness of the proposed dynamic HI-generating module is evaluated by comparing it with the No HI-generating module method. As presented in Table. \ref{tableAblationinHIconstruction}, replacing the HI-generating module with PCA leads to a decrease in all metrics in all tasks, indicating that the proposed dynamic HI-generating module has a better ability to map the low-dimensional features into one-dimensional HI than traditional dimension reduction techniques. As the inner HI-prediction block is integrated into the dynamic HI-generating module, this method removes the inner HI-prediction block as well. Compared to the method that only removes the inner HI-prediction block, the No HI-generating module method achieves better results in the Tred and Rob metrics. This further demonstrates the crucial role of the inner HI-prediction block in dynamic HI construction.
}

\textcolor{black}{
Finally, the effectiveness of both the SkipAE and the dynamic HI-generating module is demonstrated. Compared to our method, all metrics show a decline, indicating that these two components jointly facilitate HI construction. Additionally, compared to the method that only removes the dynamic HI-generating module (No HI-generating module method), its average performance decreases in all metrics, as shown in Fig. \ref{ablation_avg}. This further indicates that the proposed SkipAE contributes to learning more representative degradation features, thereby constructing a comprehensive HI with improved monotonicity, trendability, and robustness.}



\subsection{Degradation Prognosis of Rolling Bearings}
\label{secHIpre}
\subsubsection{Implementation Details}
The ability of dynamic HI in bearing degradation trend prediction is evaluated in this section. \textcolor{black}{The HIs of P-Bearing1\_1, P-Bearing2\_1, and H-Bearing, as constructed in section \ref{sec323}, are used for comparison.} For P-Bearing1\_1 and P-Bearing2\_1, the end-of-life times are set as the last recorded step of the dataset. The last 150 unknown HI points are used as test data $D_{test}$, and the remaining data points are used as training data $D_{train}$. \textcolor{black}{Given the instability of the failure degradation period for H-Bearing (shown in Fig. \ref{task3_con}), the end-of-life time is set at the 1440th step, and the time series (0-1440th steps) is normalized to a range of zero to one.} The last 290 data points data are used as test data $D_{test}$, and the remaining data points are used as training data $D_{train}$.

Unlike traditional RUL prediction methods that require an additional prediction model after HI construction, \textcolor{black}{the integrated inner prediction block can be used as the backbone for future dynamic HI prediction due to its prediction ability according to the study of Zeng et al. \cite{zeng2023transformers}. Since the experiment is to evaluate the ability of HIs in prognostics, all HIs share the same backbone for prediction.}
For comparison HIs, the backbone will be trained on the $D_{train}$ and tested on the $D_{test}$. The training epoch, learning rate, and batch size are set as 1000, 0.001, and 64, respectively. The early stopping technique with 15 epochs is adapted in all experiments. 

\textcolor{black}{The failure threshold is essential in machinery prognostics. Since all HIs constructed in section \ref{sec323} range from zero to one, the global failure threshold can be determined from the experimental observations.} To maximize data utilization, a threshold of 0.75 is pre-defined for P-Bearing1\_1 and P-Bearing2\_1, while a threshold of 0.9 is applied for the H-Bearing.

\subsubsection{Evaluation Metrics}
To quantitatively evaluate the prediction ability of HIs, root mean square error (RMSE) and predictability index (Pred) \cite{lei2018machinery, javed2015new} are used as evaluation metrics. 
RMSE is defined as
\begin{equation}
    RMSE = \sqrt{\frac{1}{K}\sum_{k=1}^{K}(\hat{y}_k-y_k)^2},\label{eq17}
\end{equation}
where $\hat{y}_k$ and $y_k$ denote the predicted HI and the ground truth HI at time step $k$.
The Pred index is proposed by Javed et al. \cite{javed2015new, baptista2022relation} for predictable feature selection. \textcolor{black}{Here, Pred is selected to assess whether the HI is more predictable and can effectively contribute to prognostic tasks.} To make it capable of predictability evaluation for HIs, the Pred index is updated and formulated as
\begin{equation}
    Pred(y~|~K,L) = \exp\left (-\ln(2)\frac{\frac{1}{K}\sum_{k=1}^{K}|\hat{y}_k-y_k|}{L} \right ).\label{pred}
\end{equation}
The proposed Pred index maps the mean absolute prediction error onto an interval of $[0,1]$. A higher Pred value indicates better predictability of the HI. In general, the HI is considered predictable when the Pred value is greater than 0.5. $L$ is a desired performance limit that is set to 0.3 via numerous experiments in this paper.

\subsubsection{Comparison Results of Various Methods}
\begin{figure}[!h]
\centering
\subfigure[]{
\begin{minipage}[t]{0.32\linewidth}
    \centering
    \includegraphics[width=2.1in]{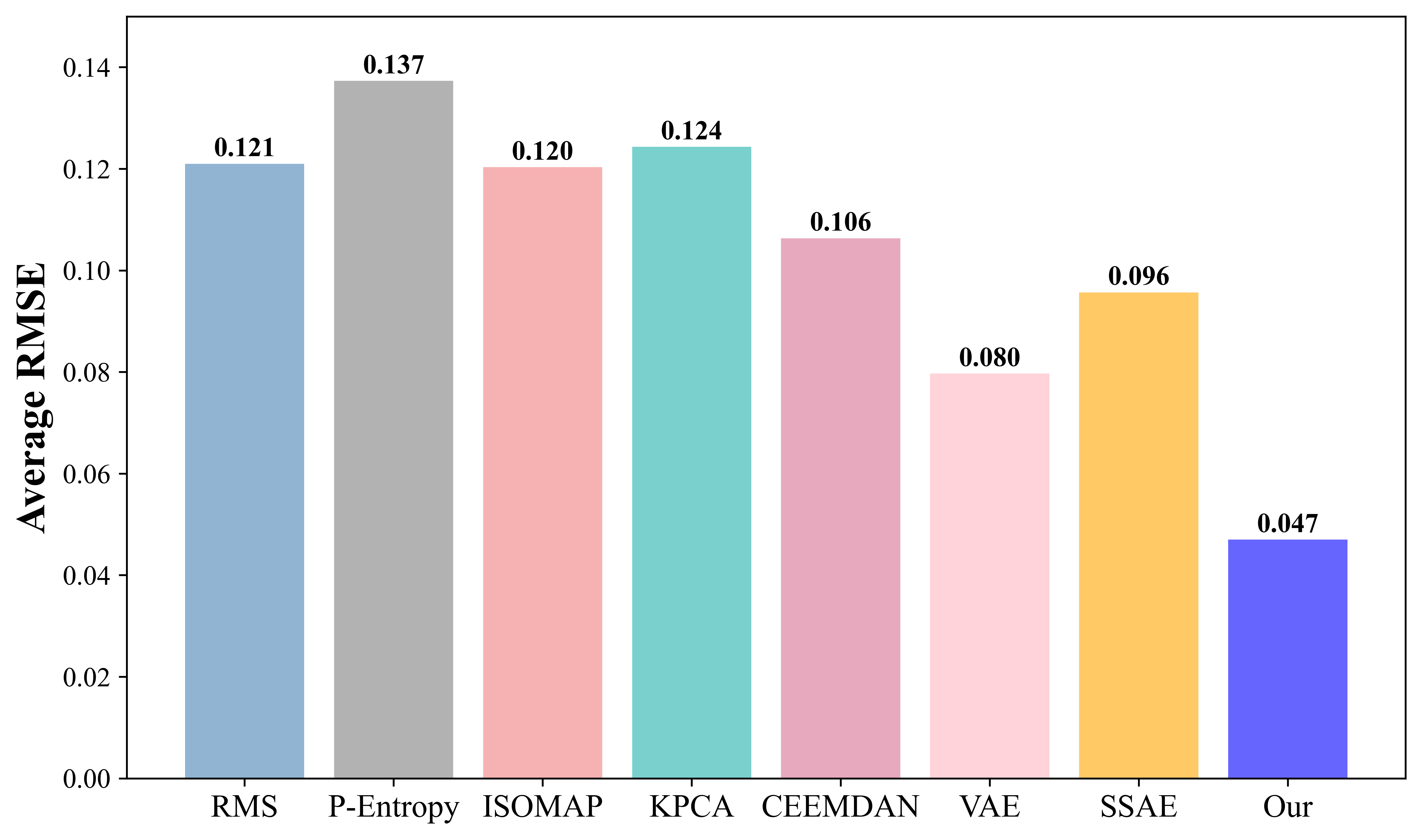}
\end{minipage}%
}
\subfigure[]{
\begin{minipage}[t]{0.32\linewidth}
    \centering
    \includegraphics[width=2.1in]{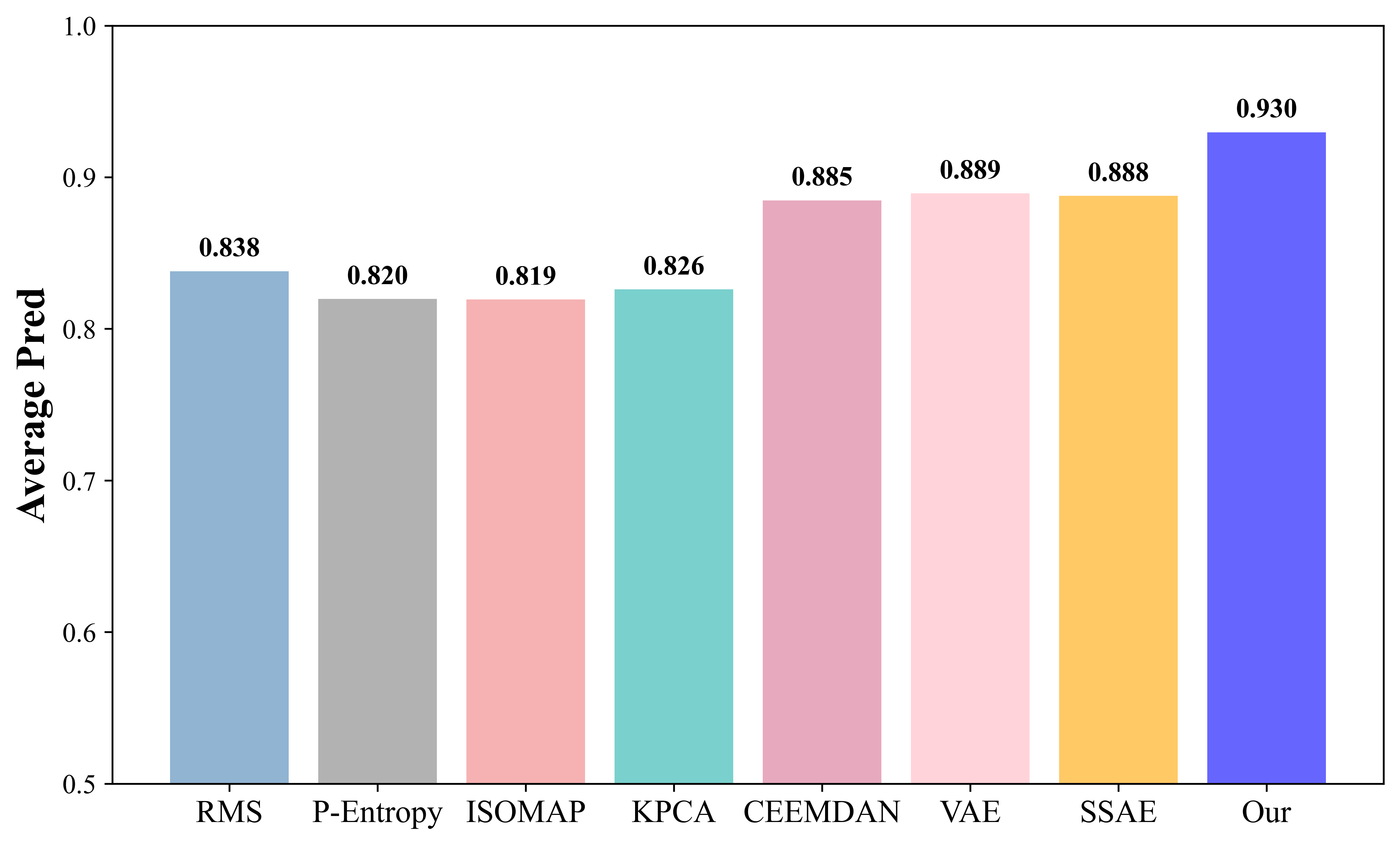}
\end{minipage}%
}
\centering
\vspace{-3mm}
\caption{\textcolor{black}{The average values of various metrics obtained by comparison methods across all tasks. (a) RMSE, (b) Pred.}}
\label{pre_avg}
\end{figure}

\begin{figure}[!h]
\vspace{-3mm}
\centering
\subfigure[]{
\begin{minipage}[t]{0.18\linewidth}
    \centering
    \includegraphics[width=1.2in]{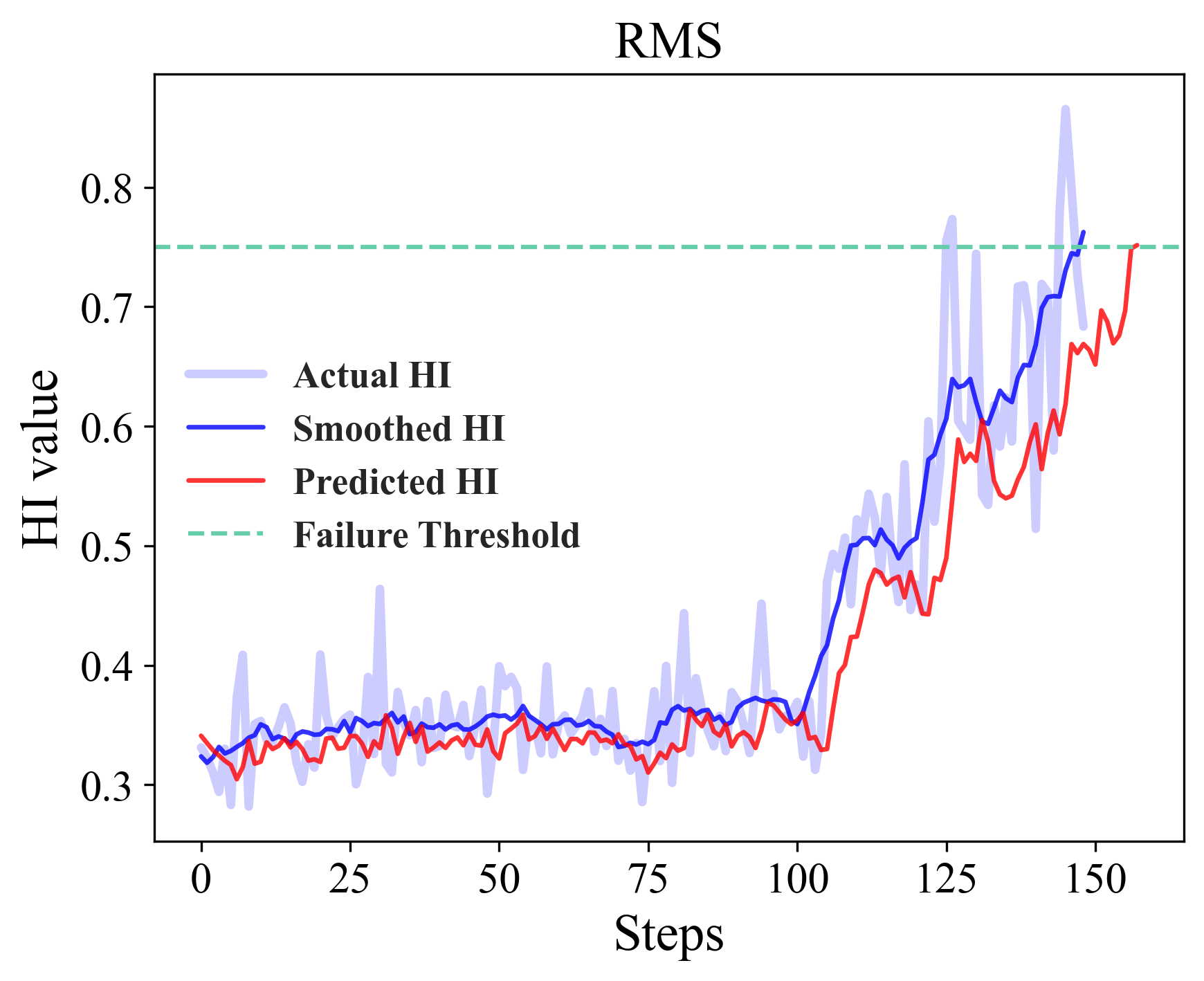}
\end{minipage}%
}%
\subfigure[]{
\begin{minipage}[t]{0.18\linewidth}
    \centering
    \includegraphics[width=1.2in]{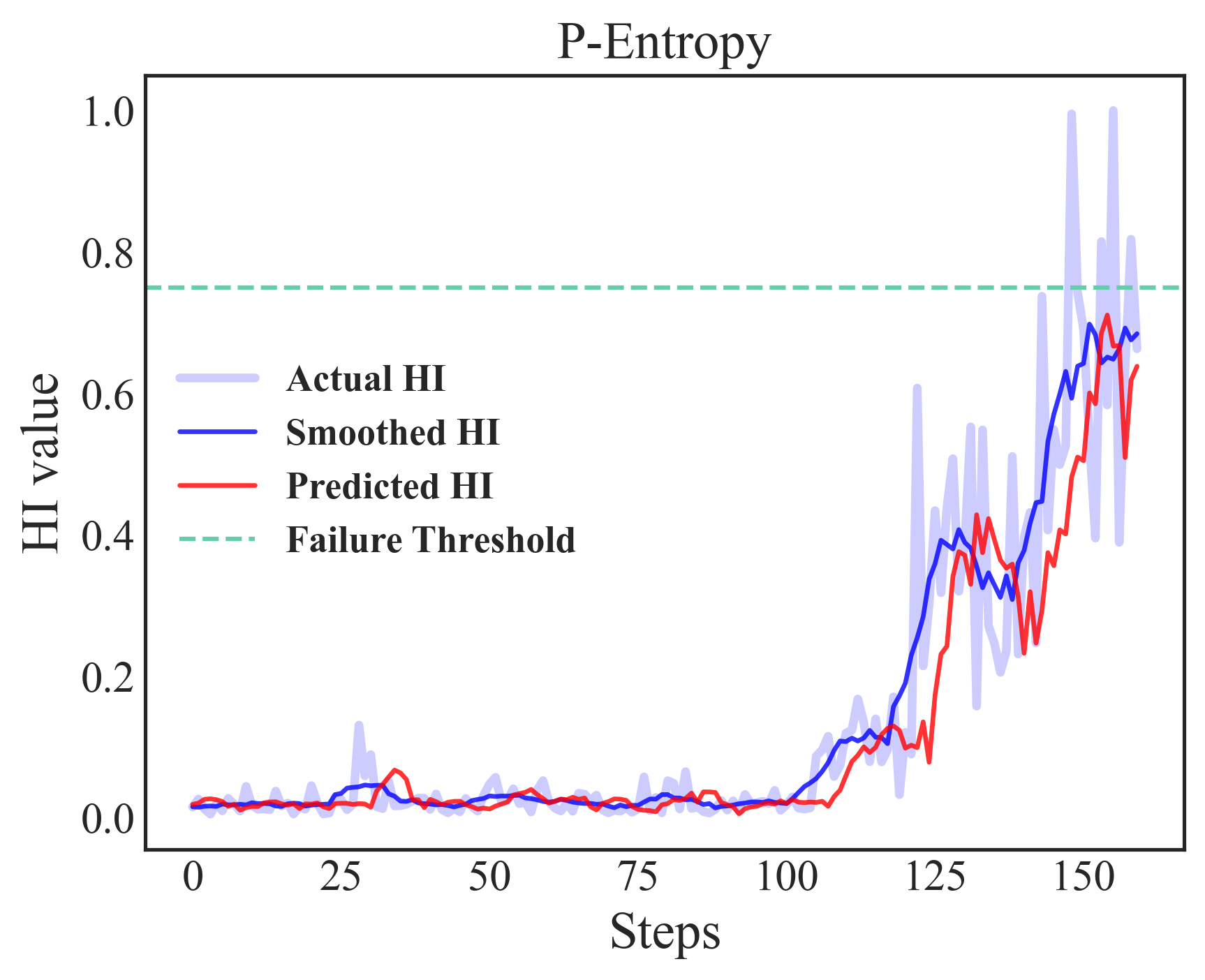}
\end{minipage}%
}
\subfigure[]{
\begin{minipage}[t]{0.18\linewidth}
    \centering
    \includegraphics[width=1.2in]{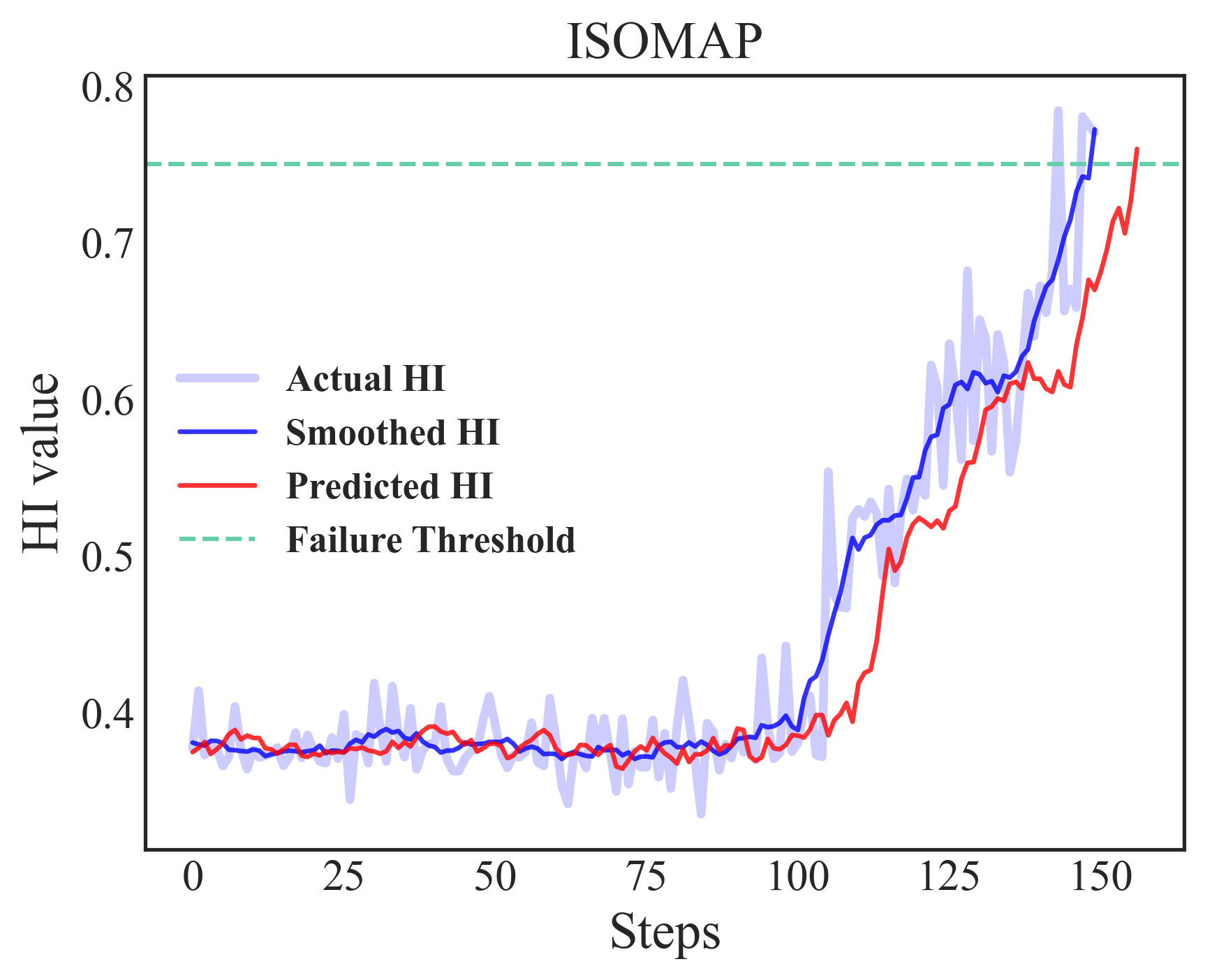}
\end{minipage}%
}
\subfigure[]{
\begin{minipage}[t]{0.18\linewidth}
    \centering
    \includegraphics[width=1.2in]{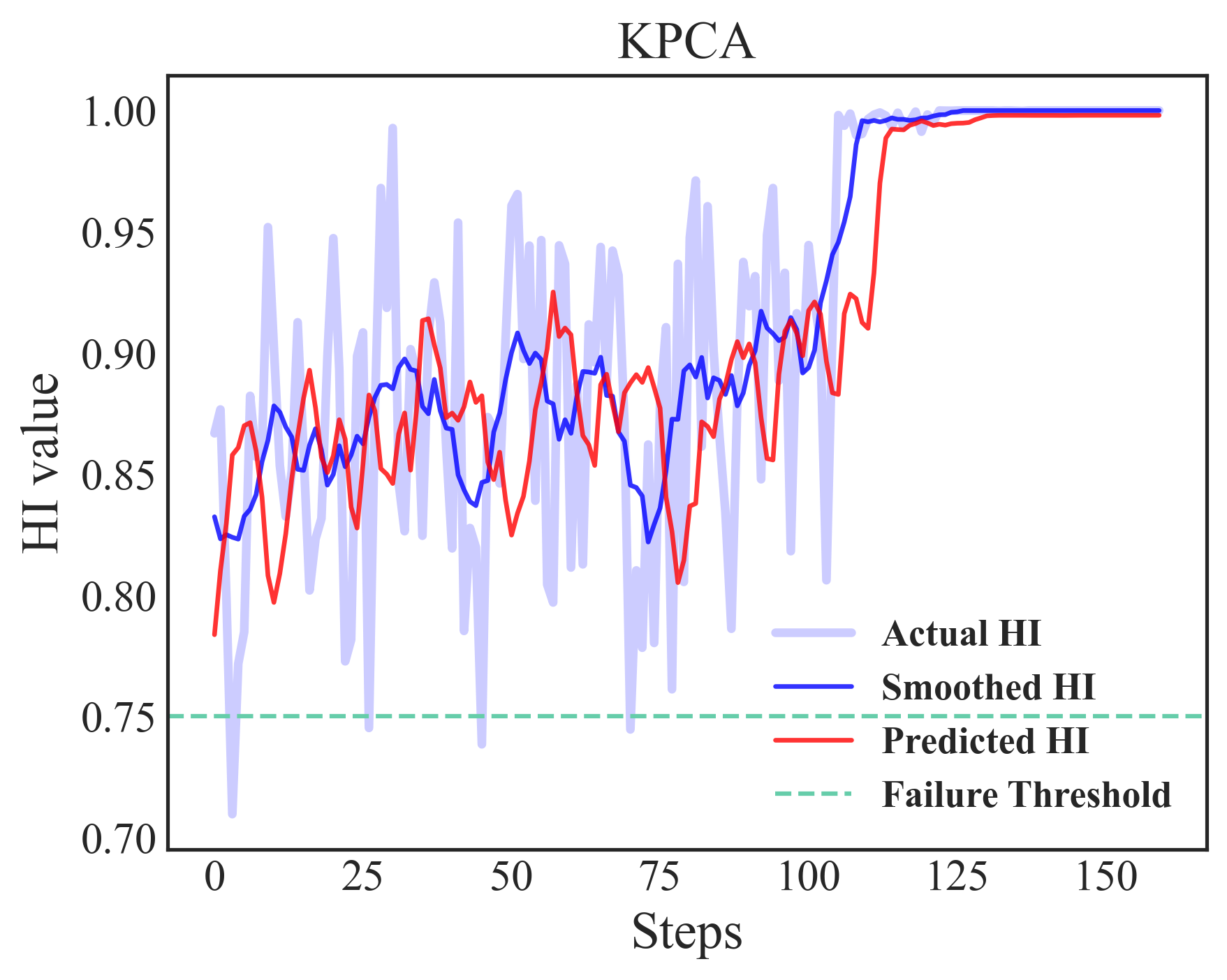}
\end{minipage}%
}\\
\vspace{-3mm}
\subfigure[]{
\begin{minipage}[t]{0.18\linewidth}
    \centering
    \includegraphics[width=1.2in]{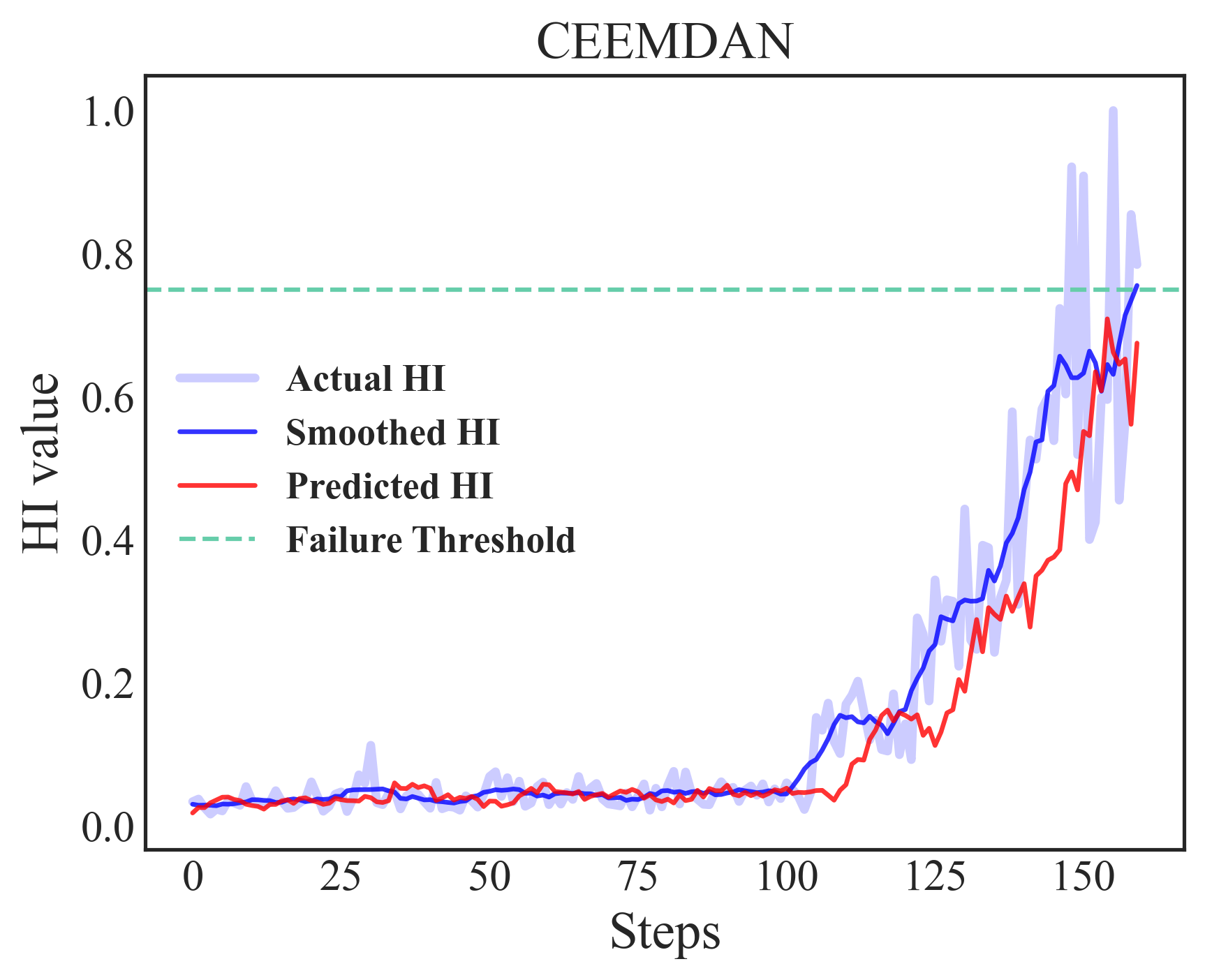}
\end{minipage}%
}%
\subfigure[]{
\begin{minipage}[t]{0.18\linewidth}
    \centering
    \includegraphics[width=1.2in]{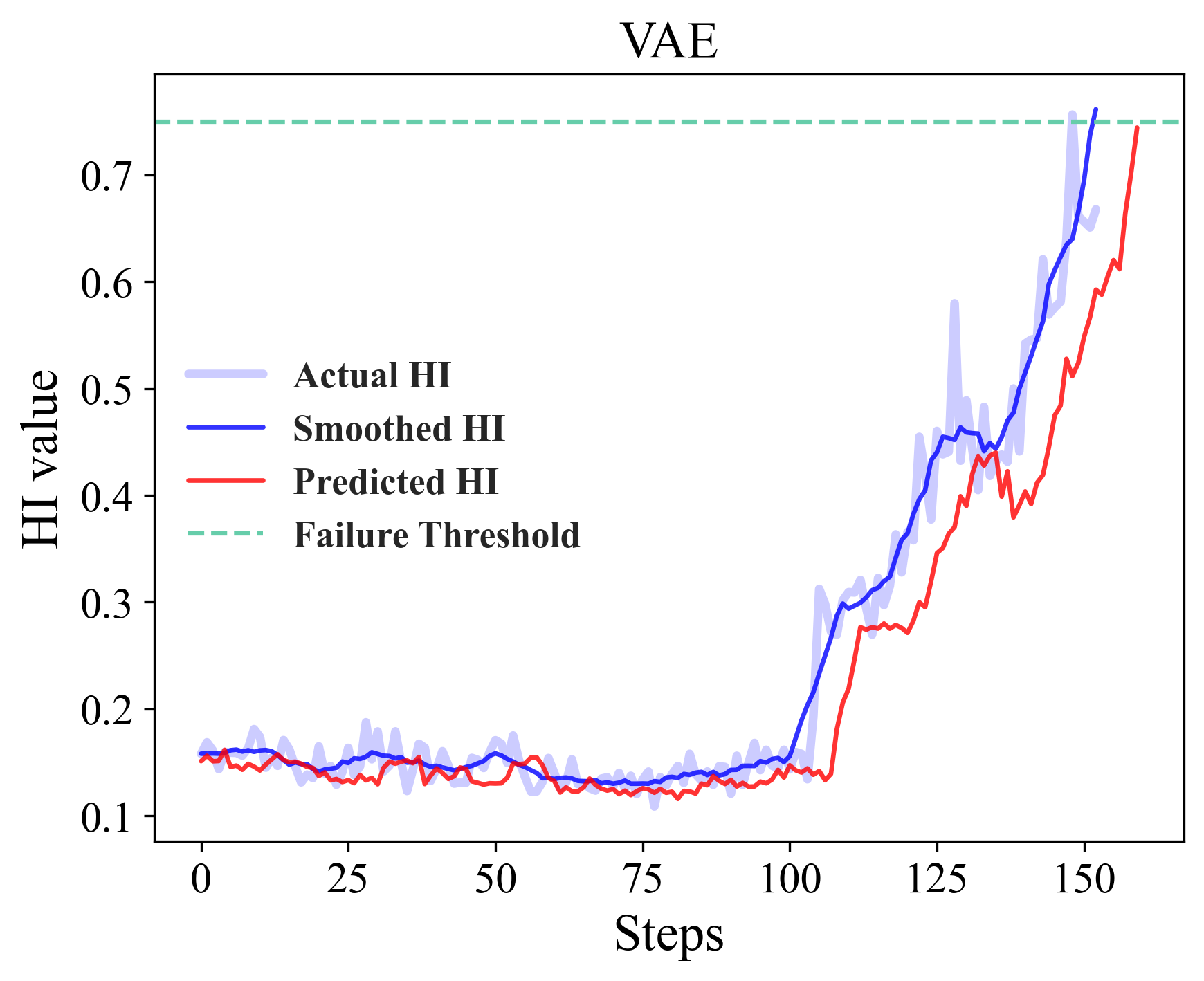}
\end{minipage}%
}
\subfigure[]{
\begin{minipage}[t]{0.18\linewidth}
    \centering
    \includegraphics[width=1.2in]{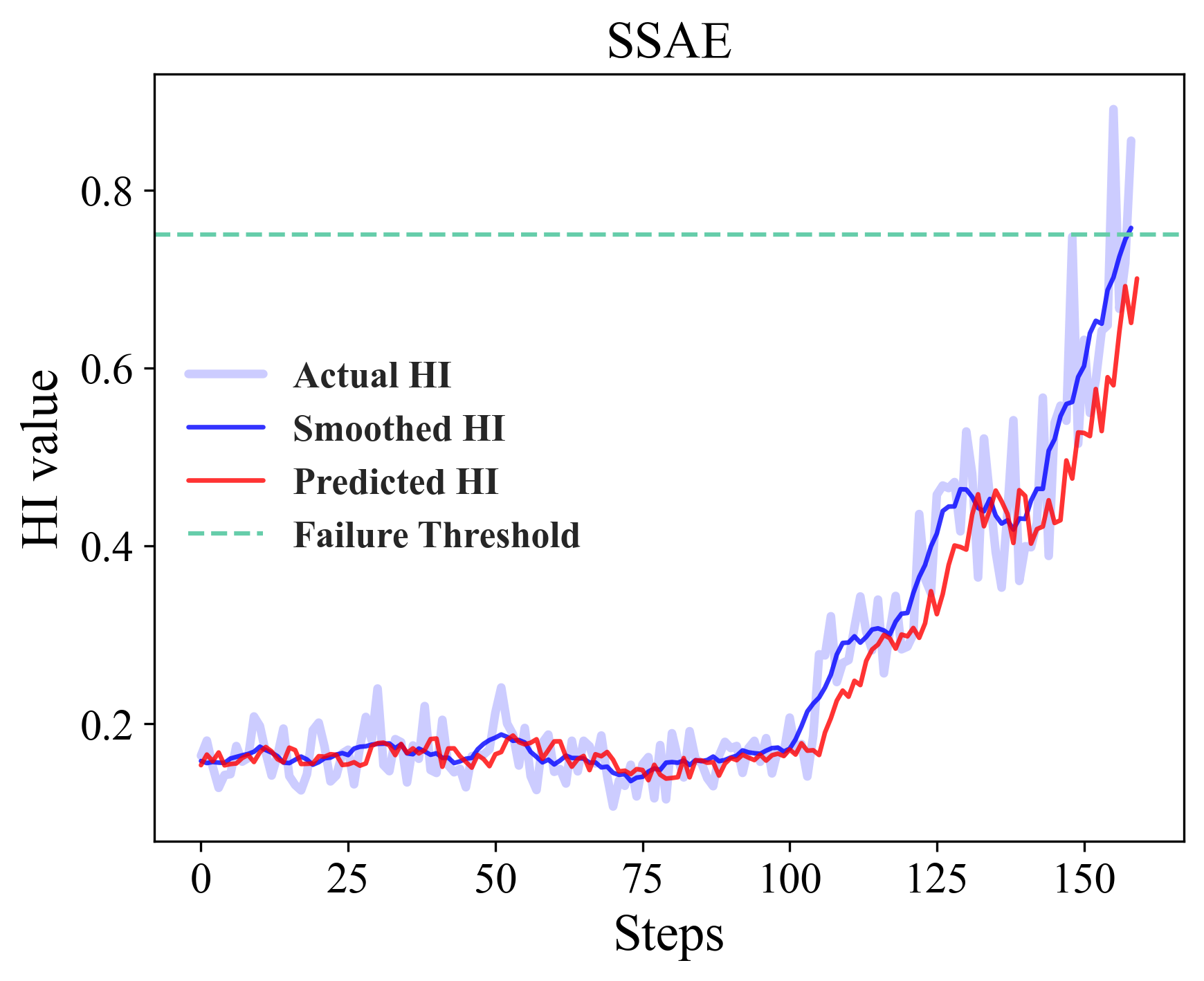}
\end{minipage}%
}
\subfigure[]{
\begin{minipage}[t]{0.18\linewidth}
    \centering
    \includegraphics[width=1.2in]{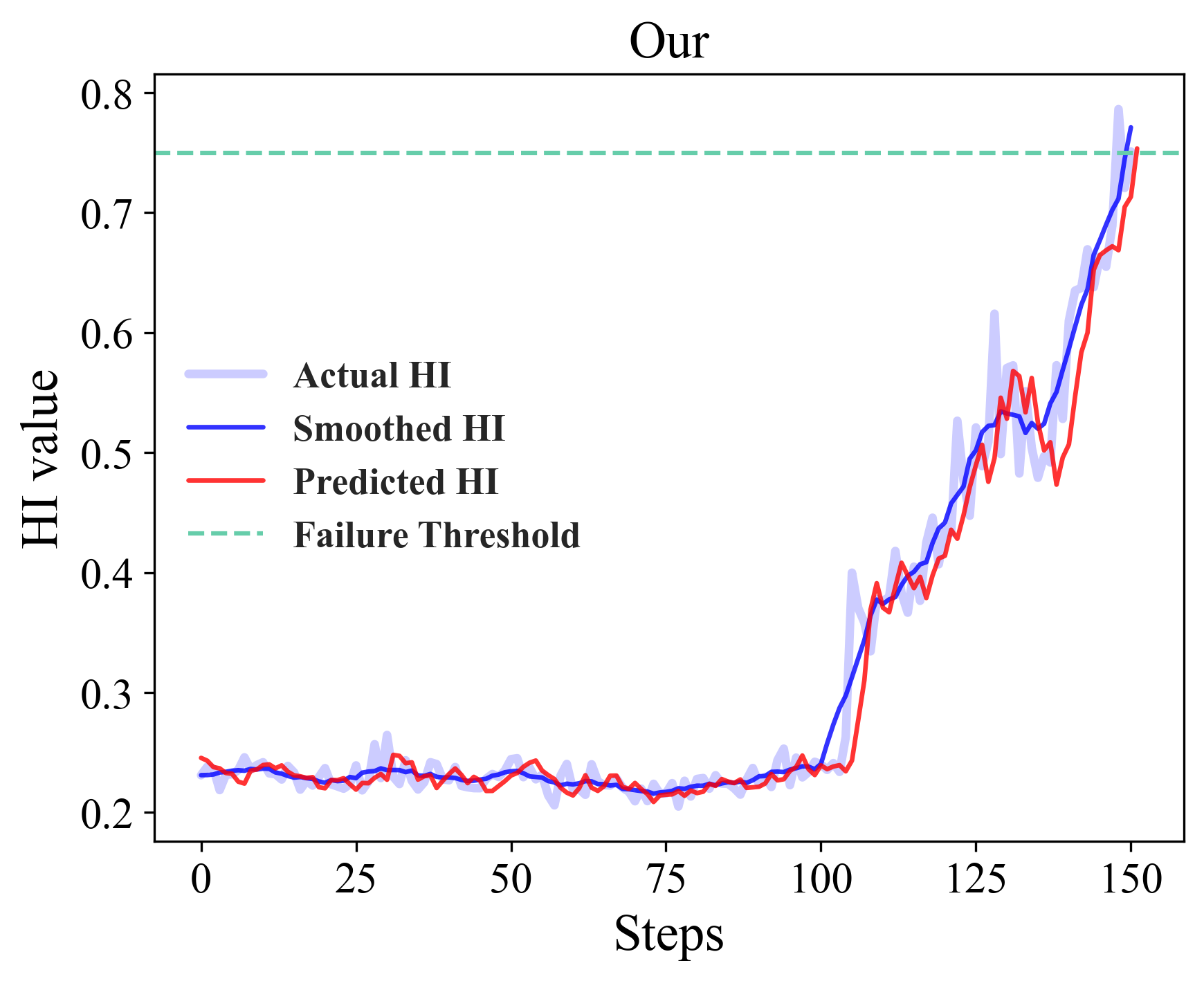}
\end{minipage}%
}
\centering
\vspace{-3mm}
\caption{\textcolor{black}{Prediction results obtained from different HIs for P-Bearing1\_1. (a) RMS, (b) P-Entropy, (c) ISOMAP, (d) KPCA, (e) CEEMDAN, (f) VAE, (g) SSAE, (h) Ours.}}
\label{pretask1}
\end{figure}

\begin{figure}[!h]
\vspace{-3mm}
\centering
\subfigure[]{
\begin{minipage}[t]{0.18\linewidth}
    \centering
    \includegraphics[width=1.2in]{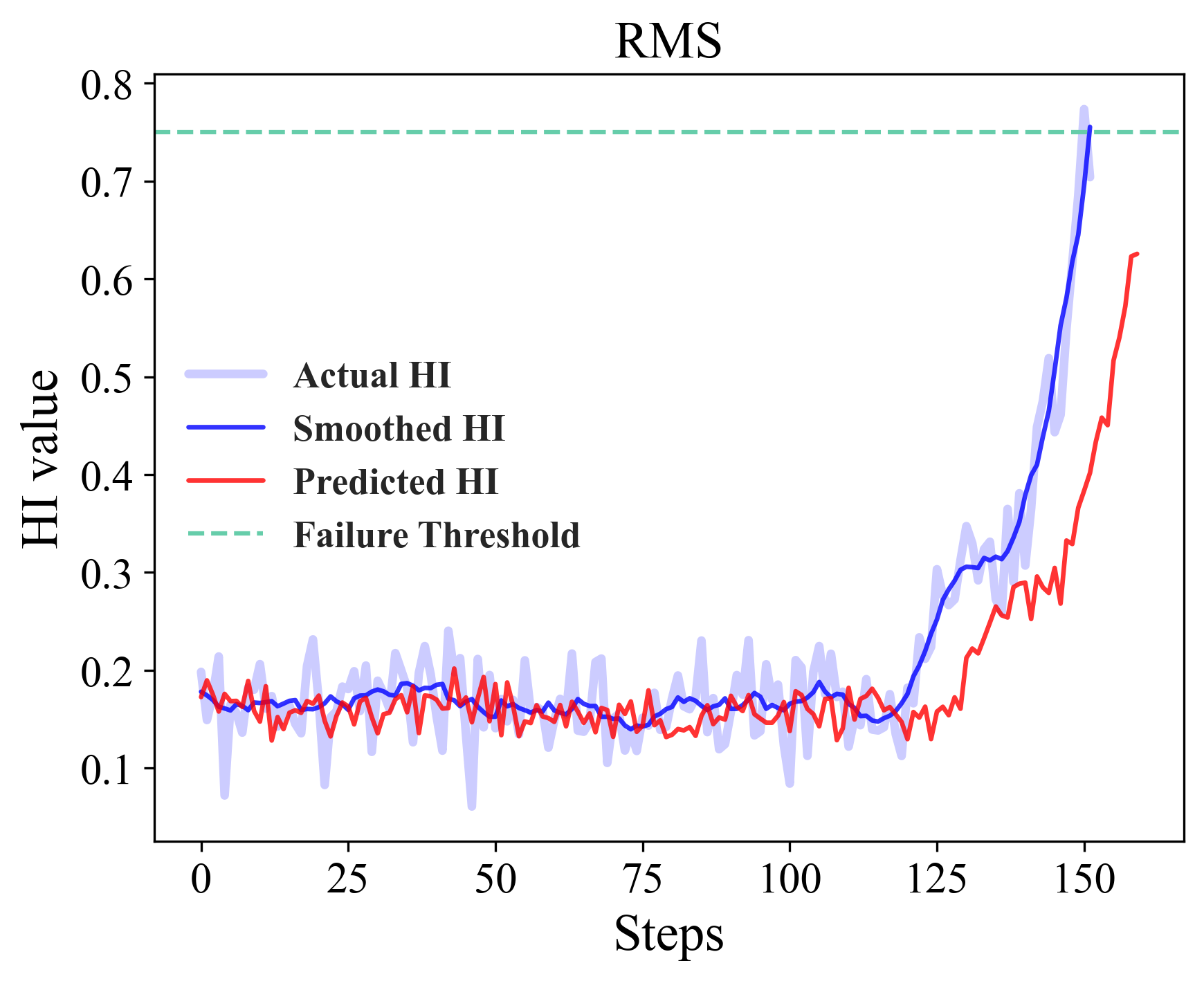}
\end{minipage}%
}%
\subfigure[]{
\begin{minipage}[t]{0.18\linewidth}
    \centering
    \includegraphics[width=1.2in]{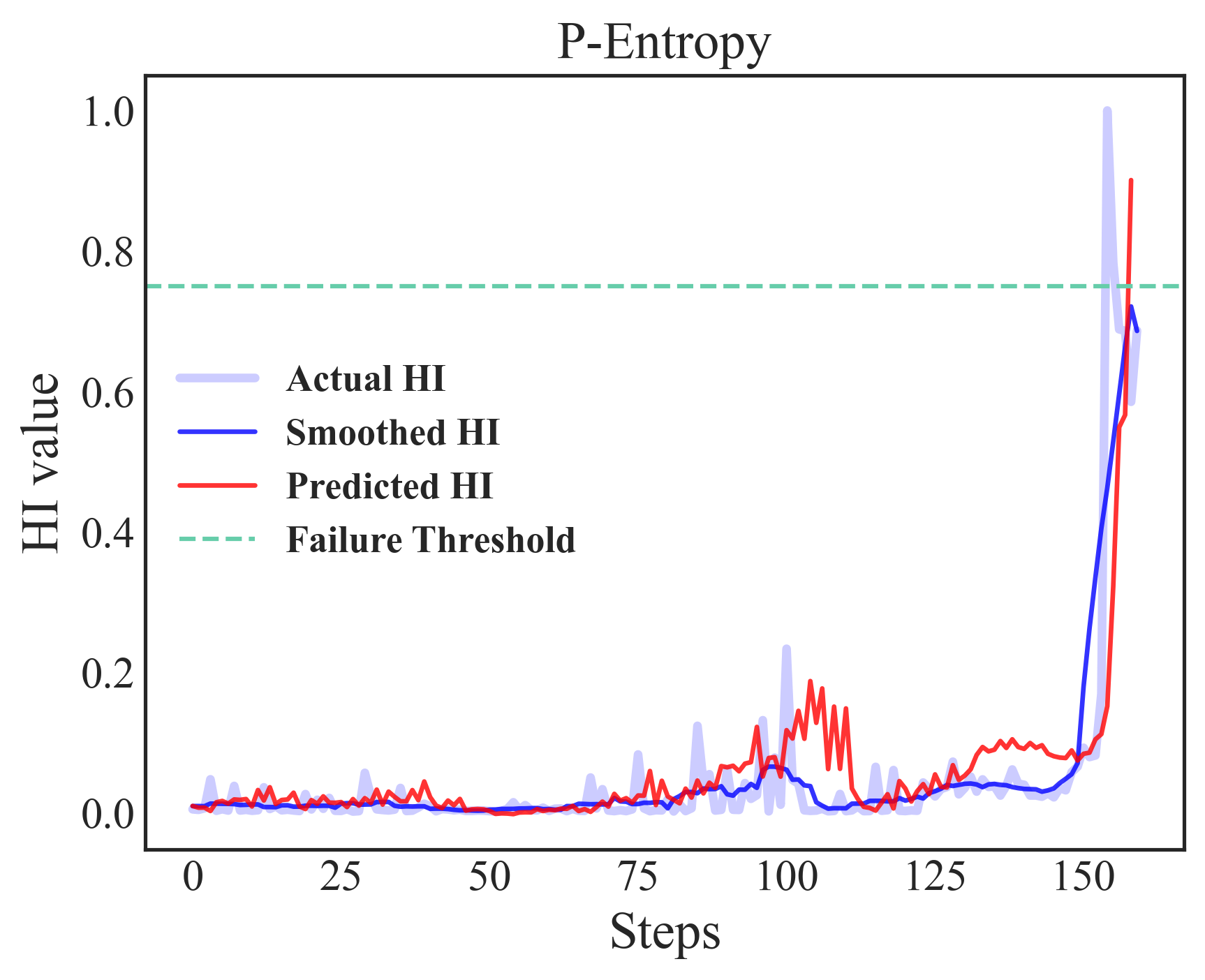}
\end{minipage}%
}
\subfigure[]{
\begin{minipage}[t]{0.18\linewidth}
    \centering
    \includegraphics[width=1.2in]{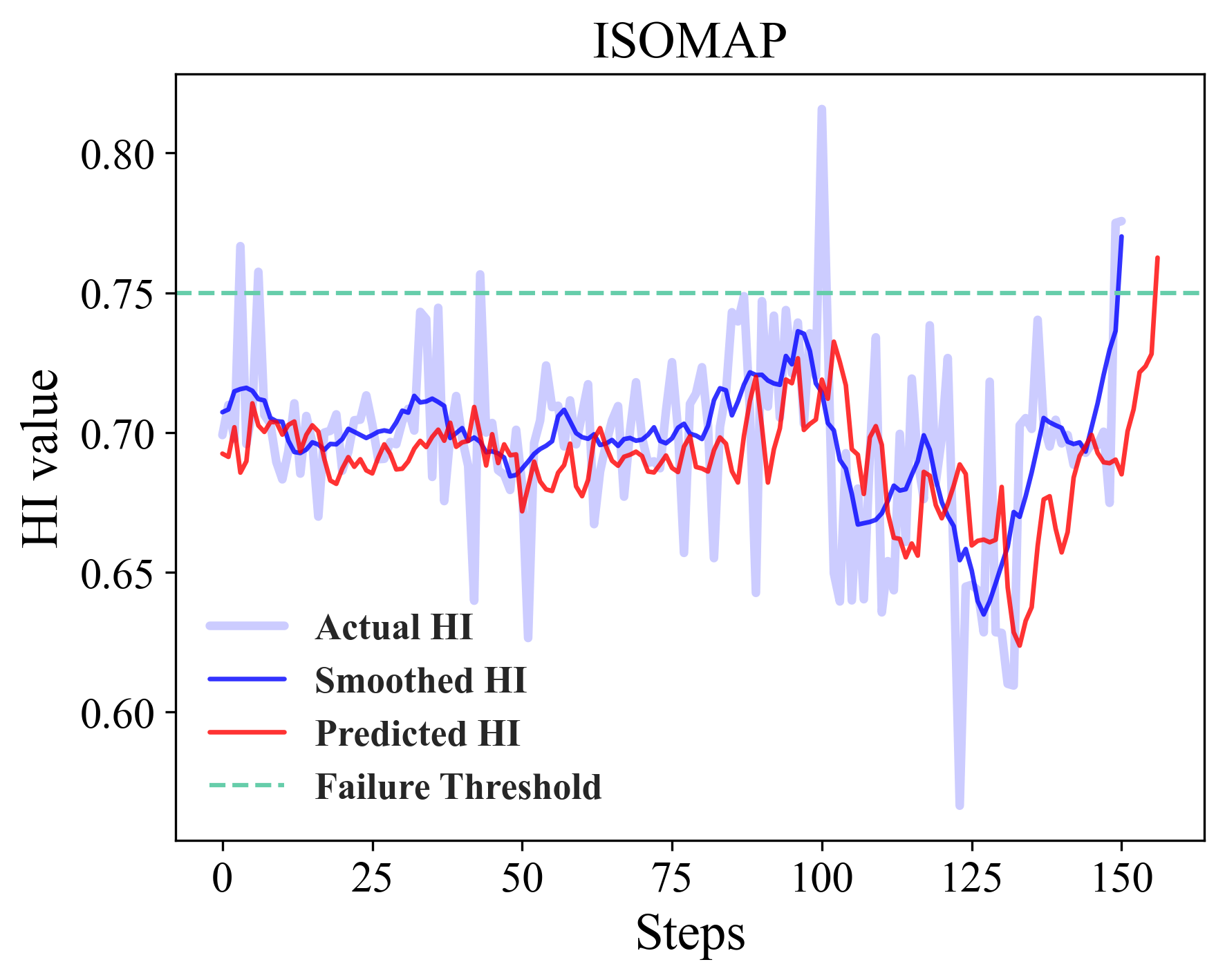}
\end{minipage}%
}
\subfigure[]{
\begin{minipage}[t]{0.18\linewidth}
    \centering
    \includegraphics[width=1.2in]{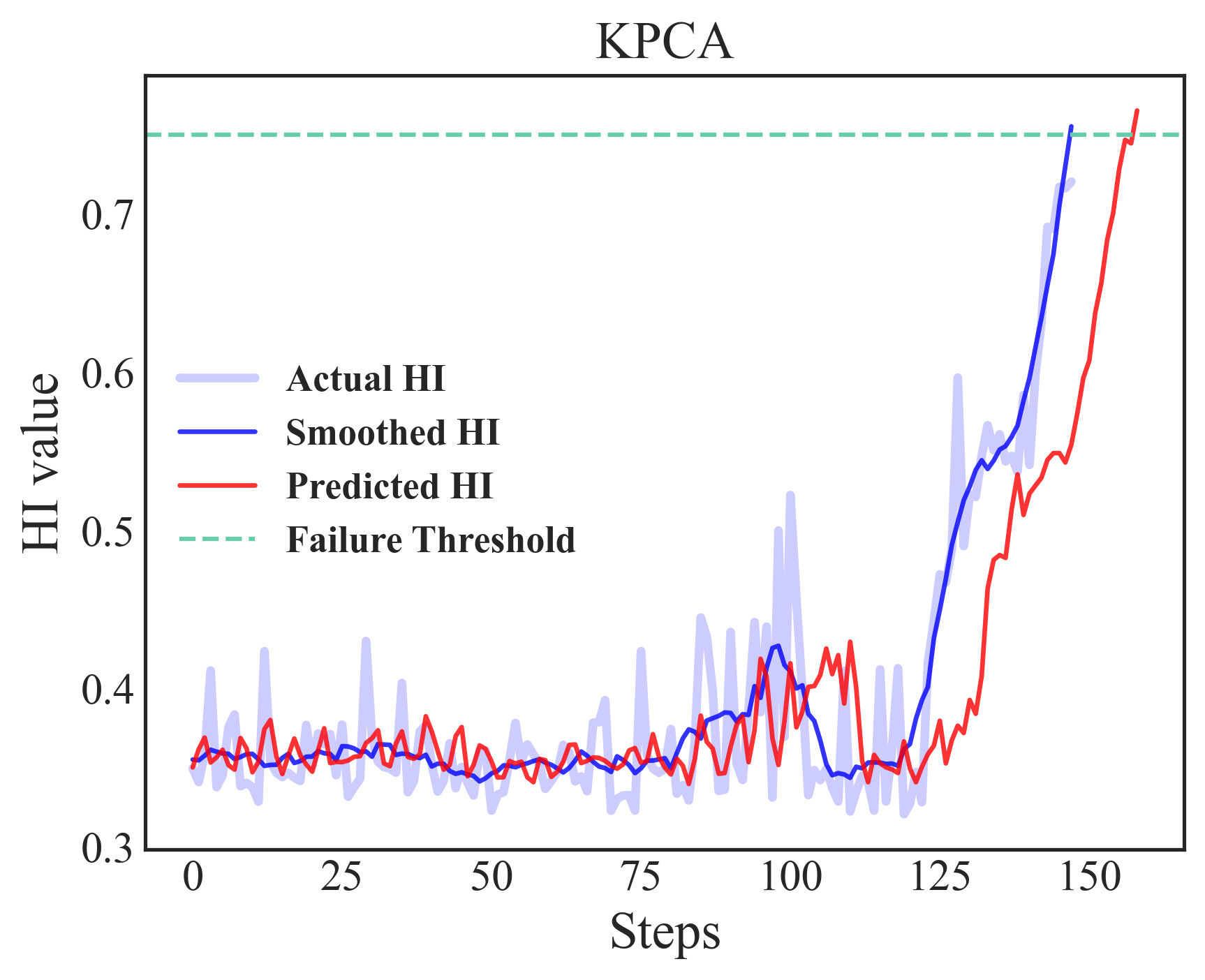}
\end{minipage}%
}\\
\vspace{-3mm}
\subfigure[]{
\begin{minipage}[t]{0.18\linewidth}
    \centering
    \includegraphics[width=1.2in]{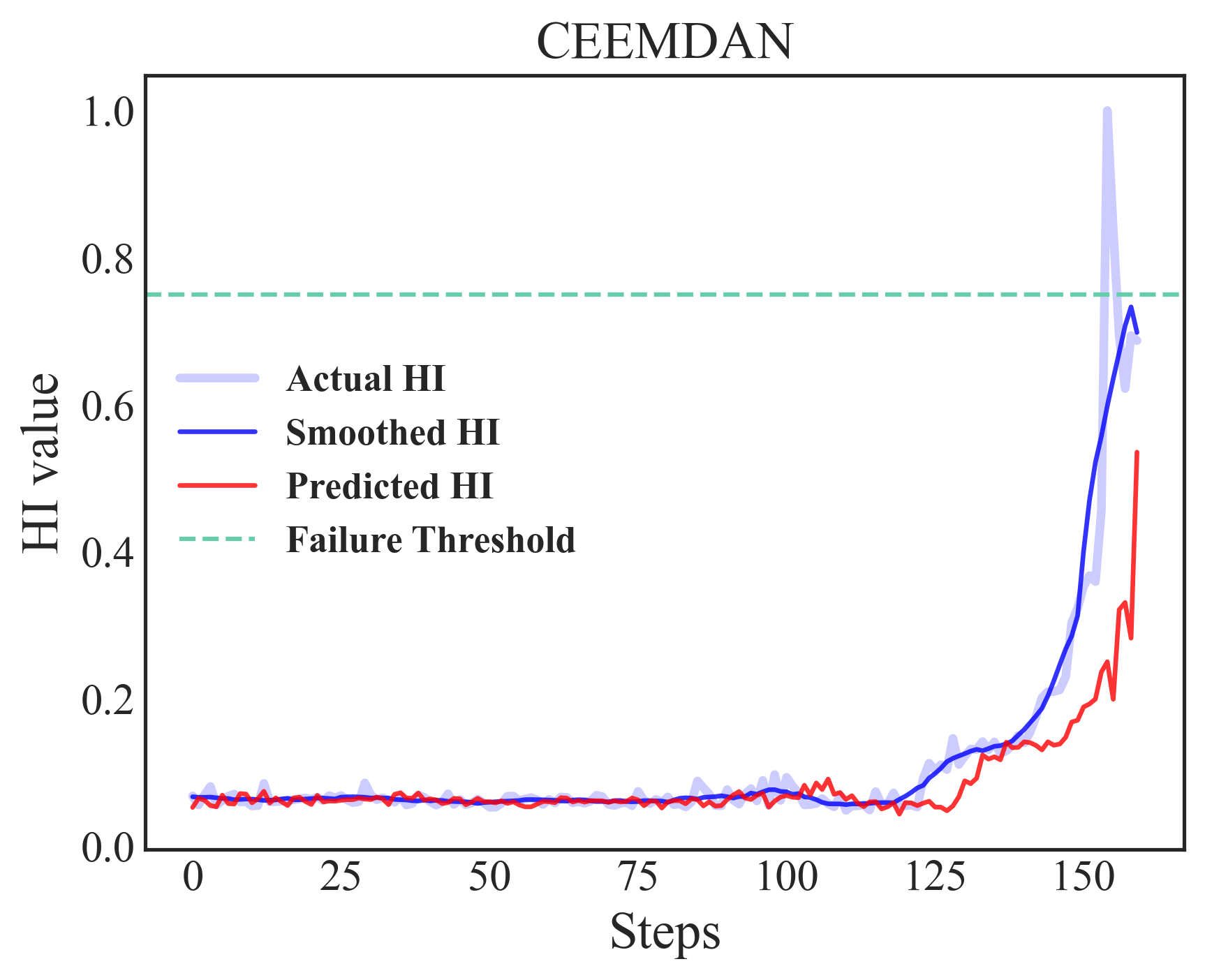}
\end{minipage}%
}%
\subfigure[]{
\begin{minipage}[t]{0.18\linewidth}
    \centering
    \includegraphics[width=1.2in]{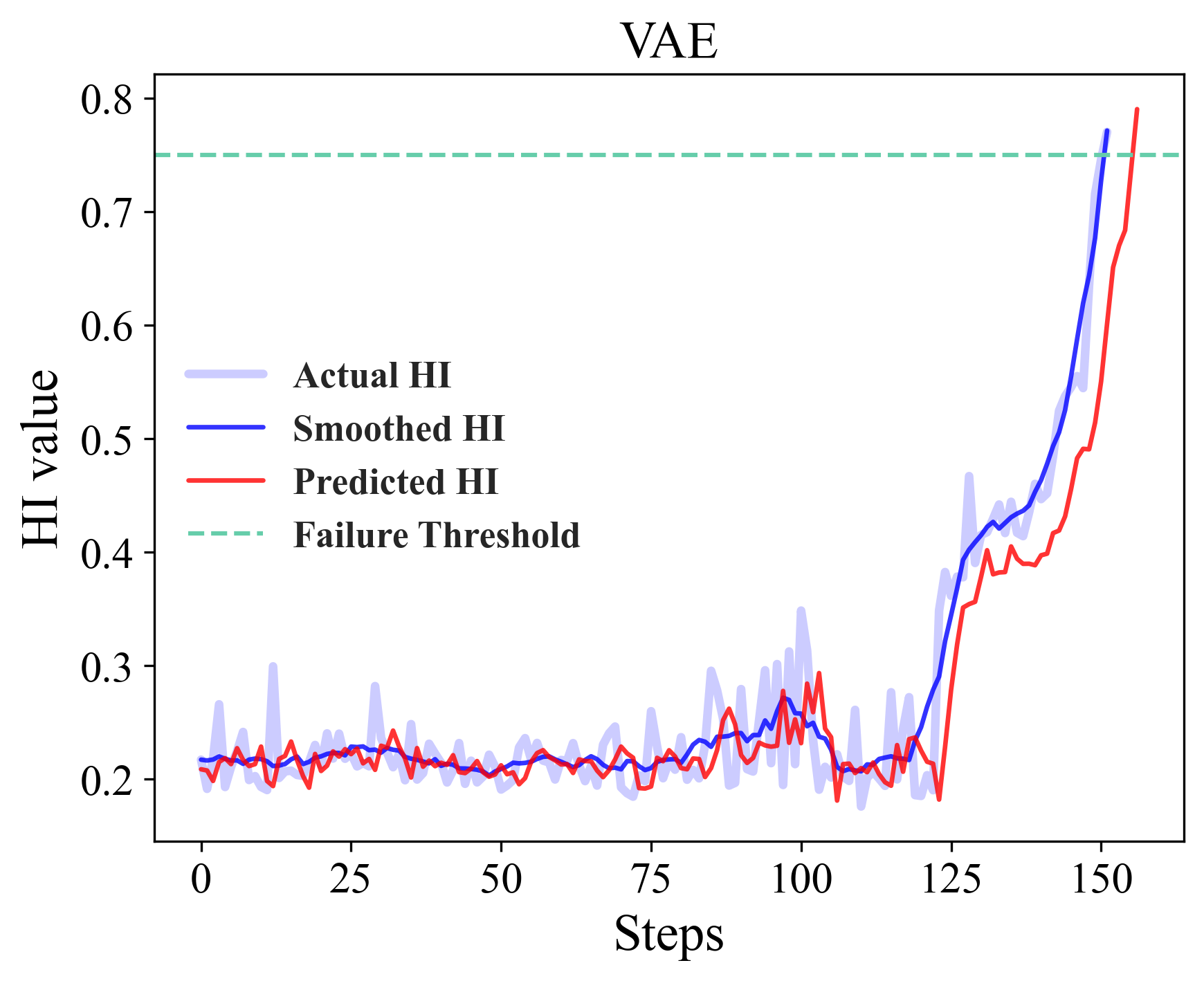}
\end{minipage}%
}
\subfigure[]{
\begin{minipage}[t]{0.18\linewidth}
    \centering
    \includegraphics[width=1.2in]{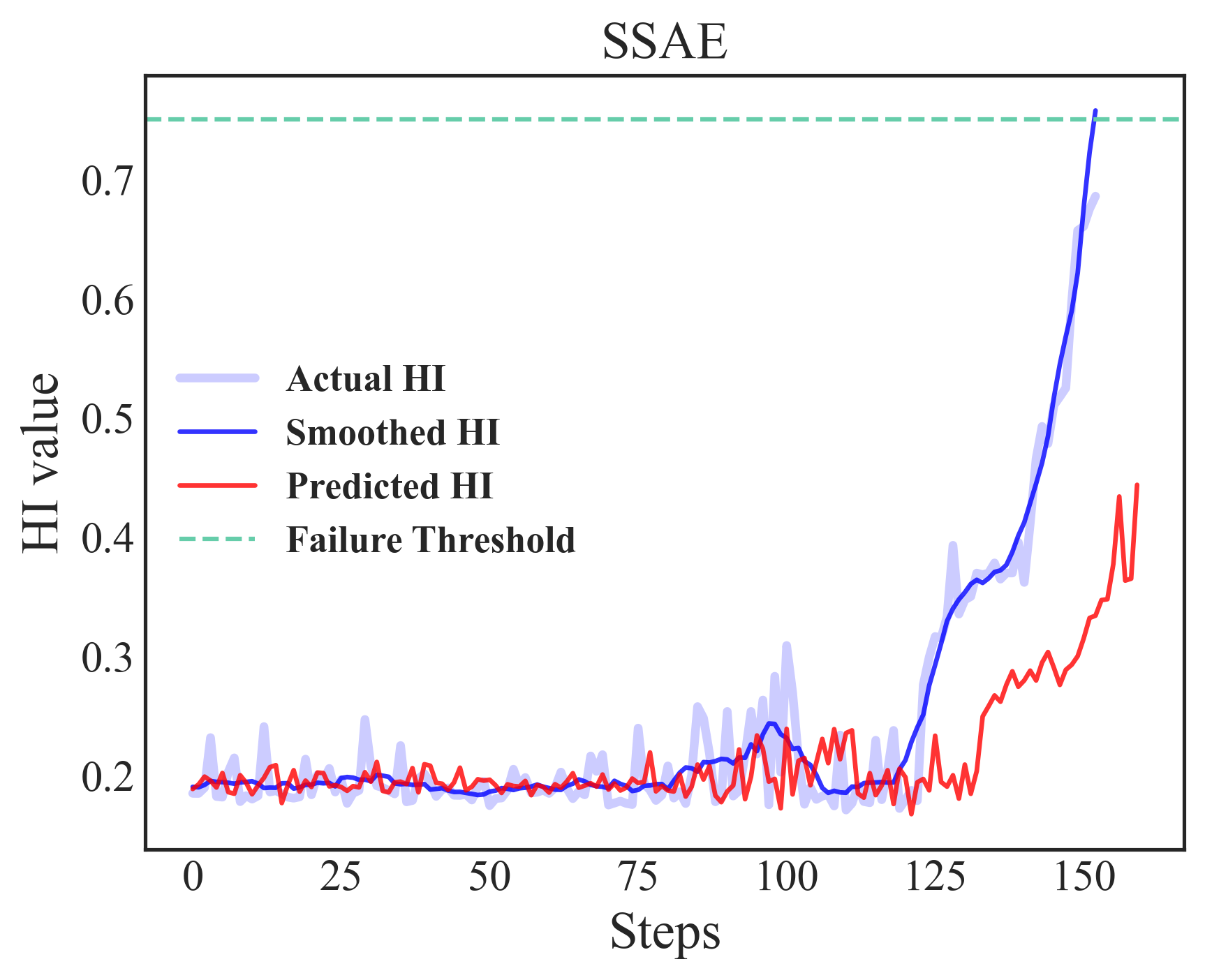}
\end{minipage}%
}
\subfigure[]{
\begin{minipage}[t]{0.18\linewidth}
    \centering
    \includegraphics[width=1.2in]{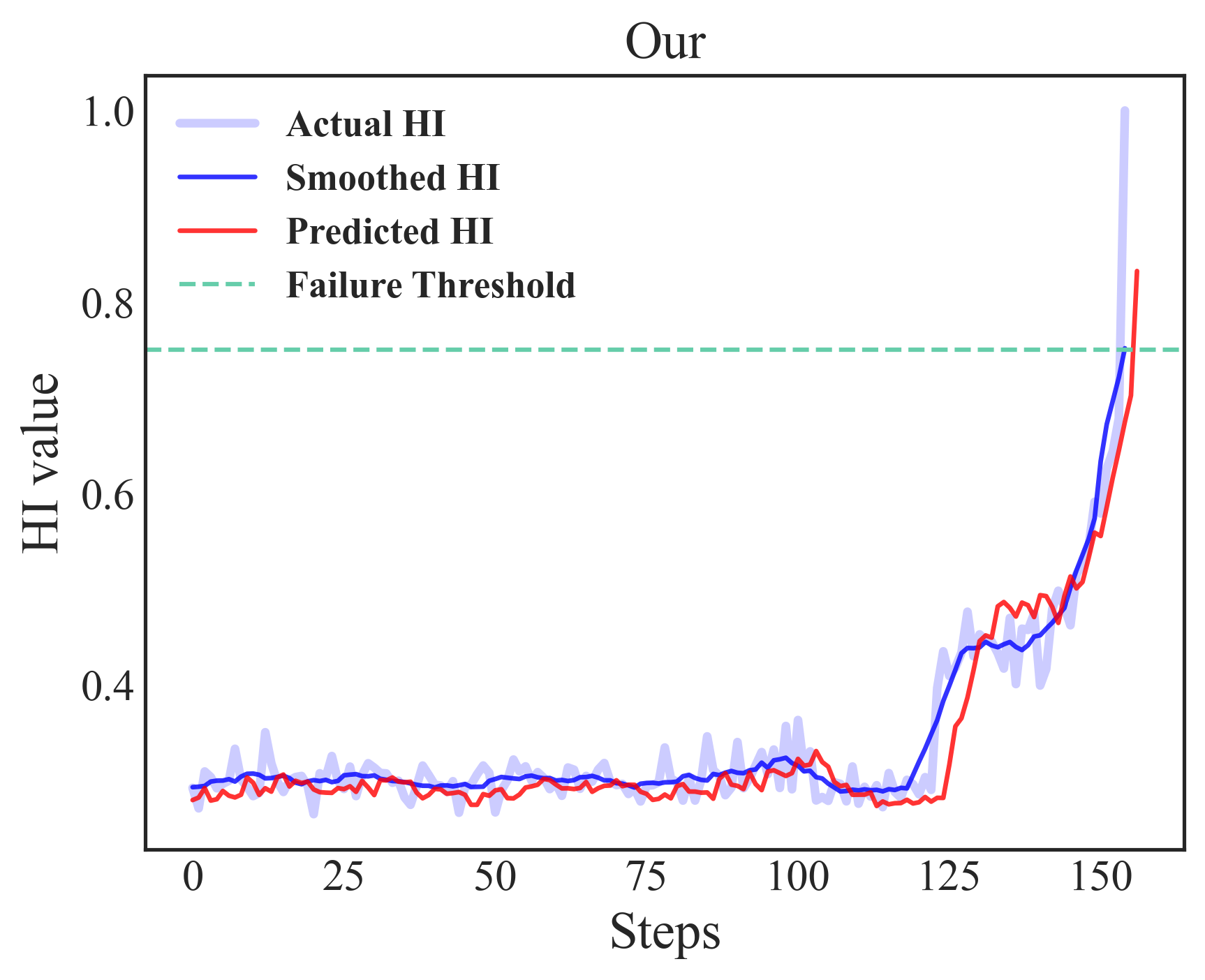}
\end{minipage}%
}
\centering
\vspace{-3mm}
\caption{\textcolor{black}{Prediction results obtained from different HIs for P-Bearing2\_1. (a) RMS, (b) P-Entropy, (c) ISOMAP, (d) KPCA, (e) CEEMDAN, (f) VAE, (g) SSAE, (h) Ours.}}
\label{pretask2}
\end{figure}

\begin{figure}[!h]
\vspace{-3mm}
\centering
\subfigure[]{
\begin{minipage}[t]{0.18\linewidth}
    \centering
    \includegraphics[width=1.2in]{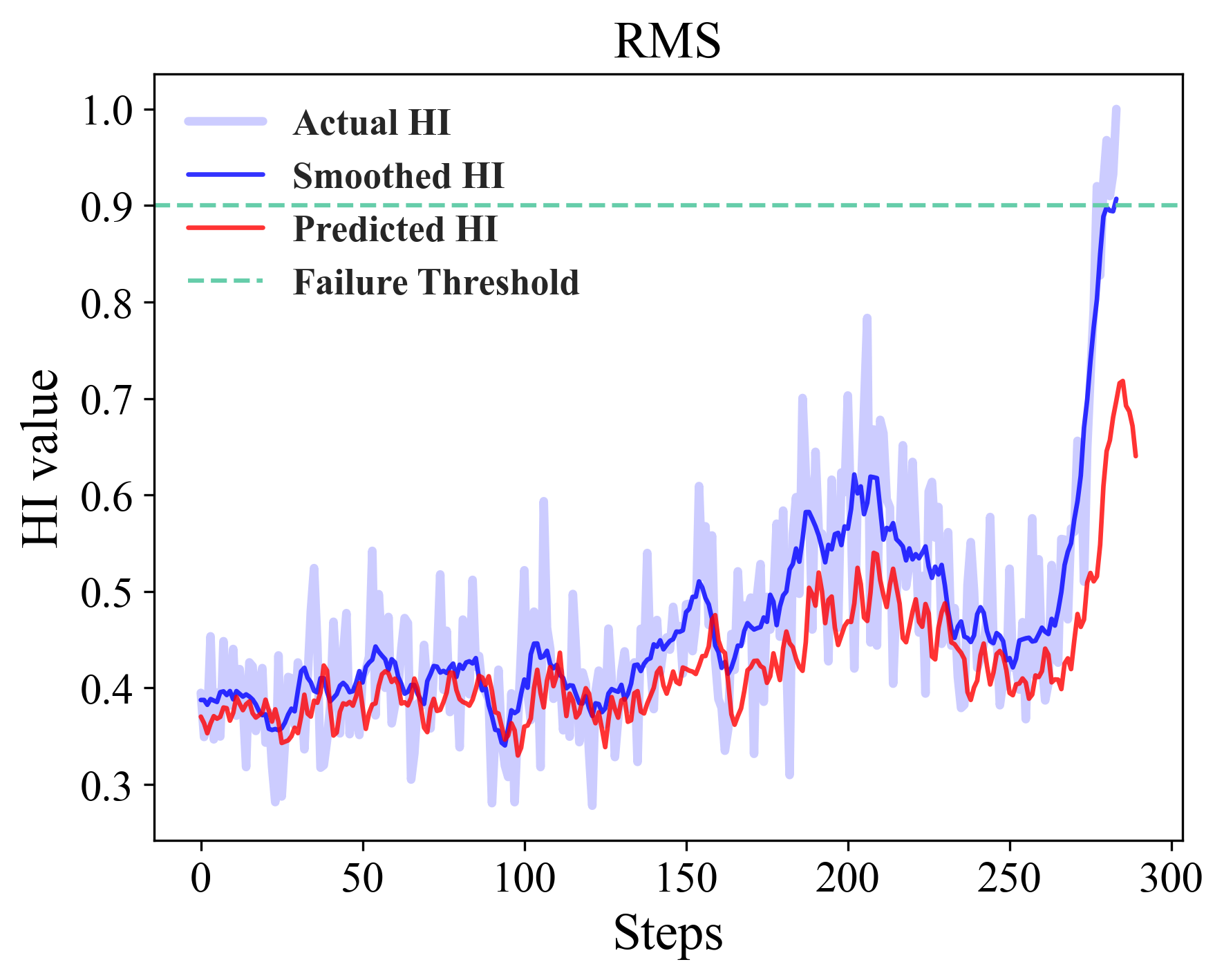}
\end{minipage}%
}%
\subfigure[]{
\begin{minipage}[t]{0.18\linewidth}
    \centering
    \includegraphics[width=1.2in]{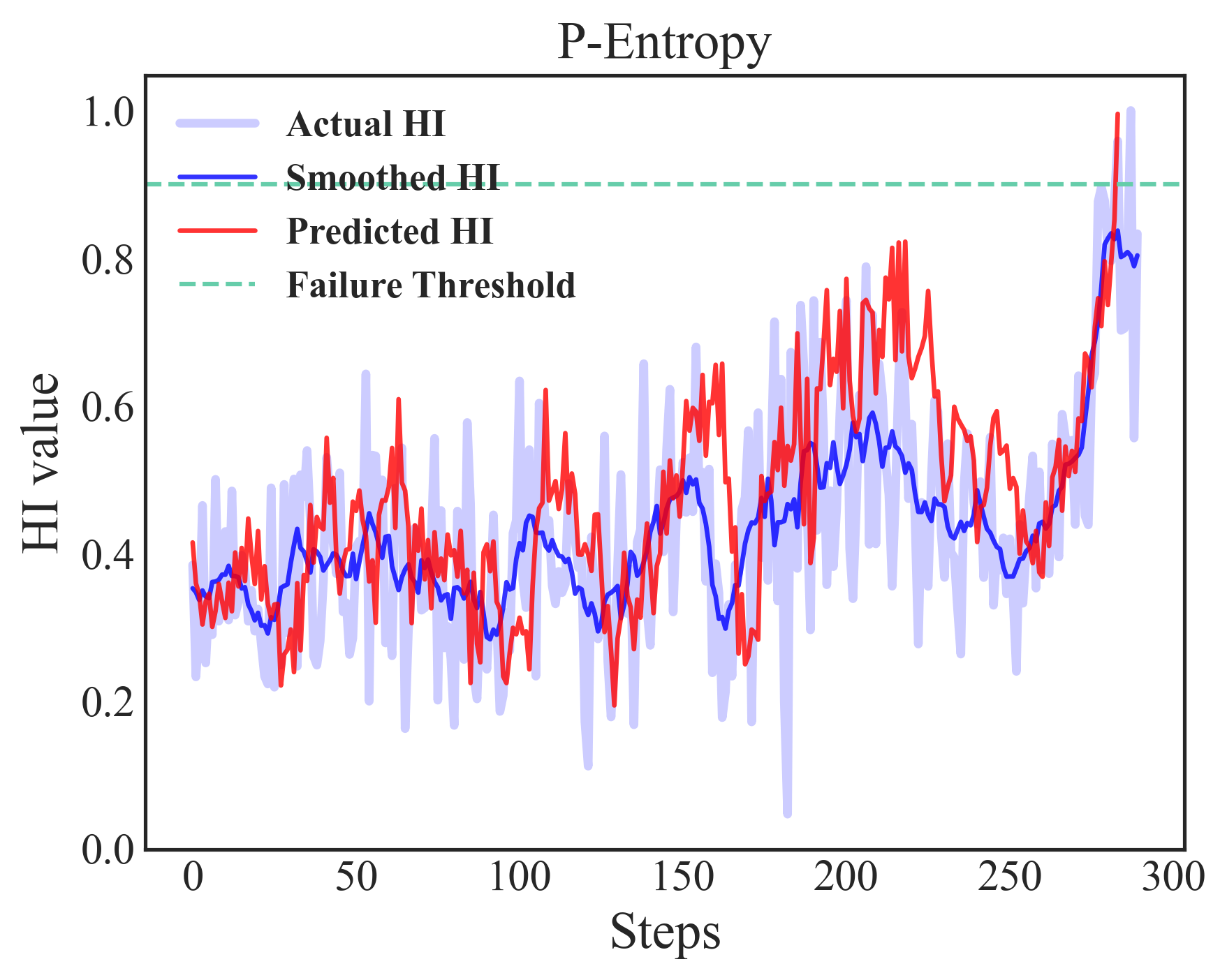}
\end{minipage}%
}
\subfigure[]{
\begin{minipage}[t]{0.18\linewidth}
    \centering
    \includegraphics[width=1.2in]{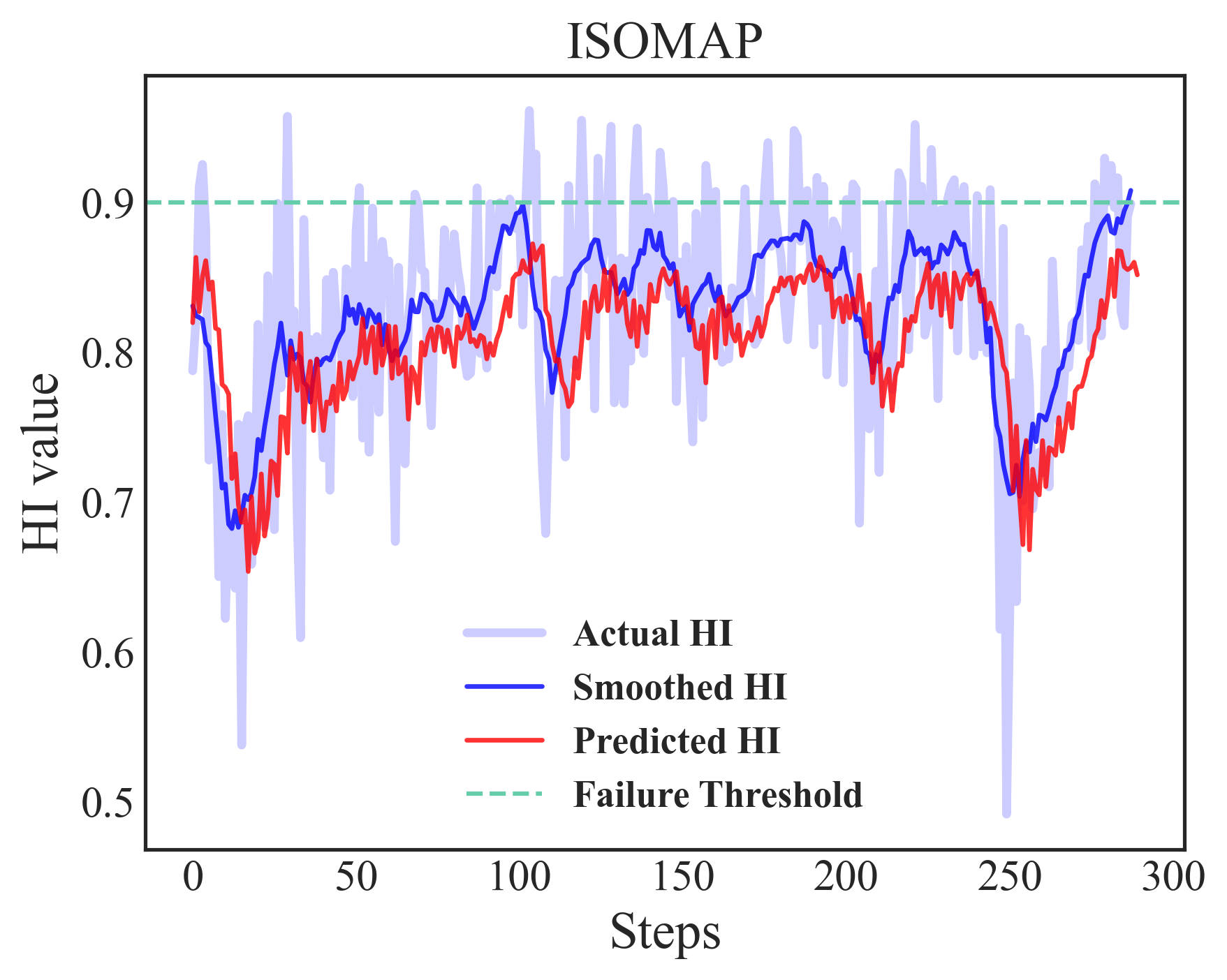}
\end{minipage}%
}
\subfigure[]{
\begin{minipage}[t]{0.18\linewidth}
    \centering
    \includegraphics[width=1.2in]{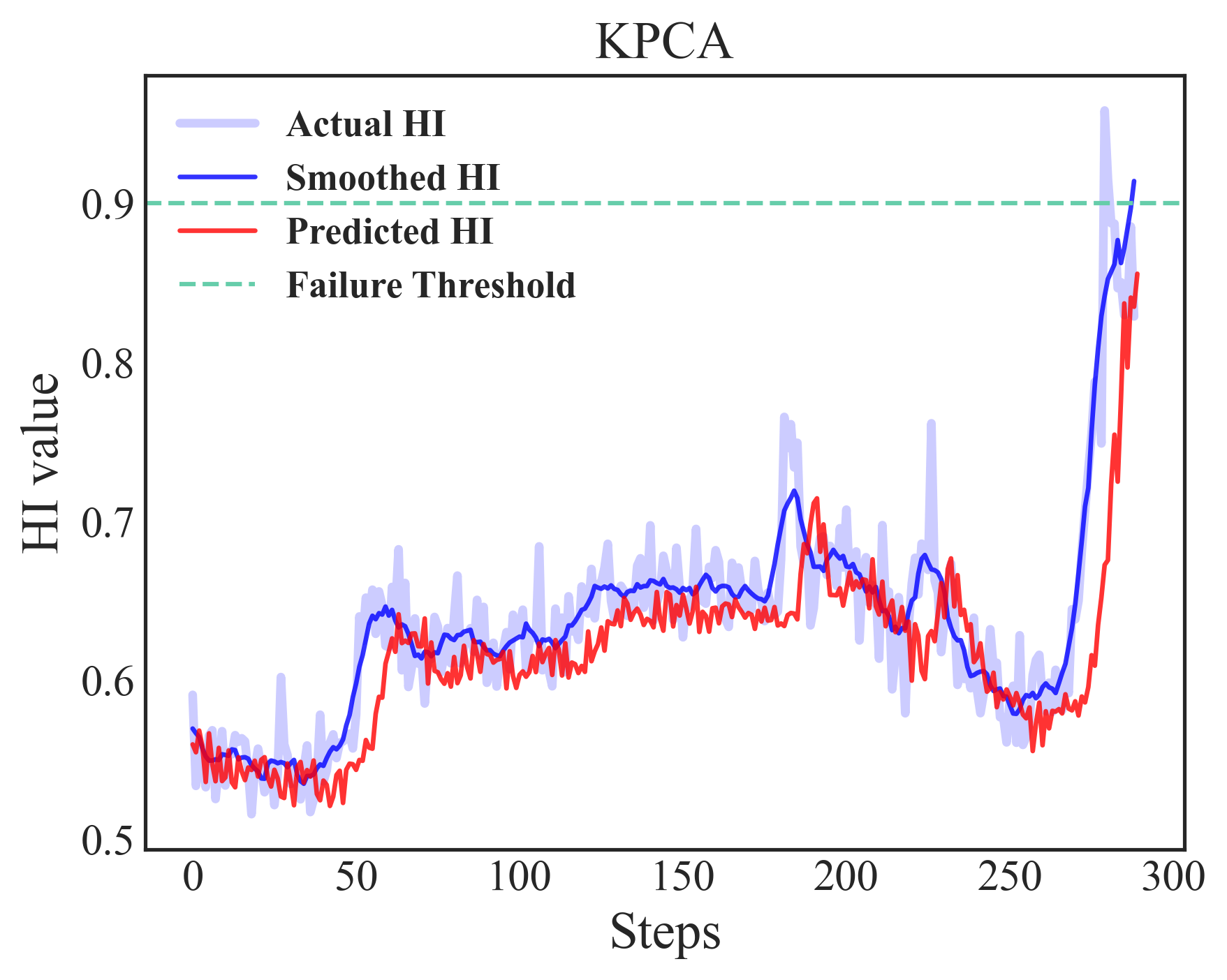}
\end{minipage}%
}\\
\vspace{-3mm}
\subfigure[]{
\begin{minipage}[t]{0.18\linewidth}
    \centering
    \includegraphics[width=1.2in]{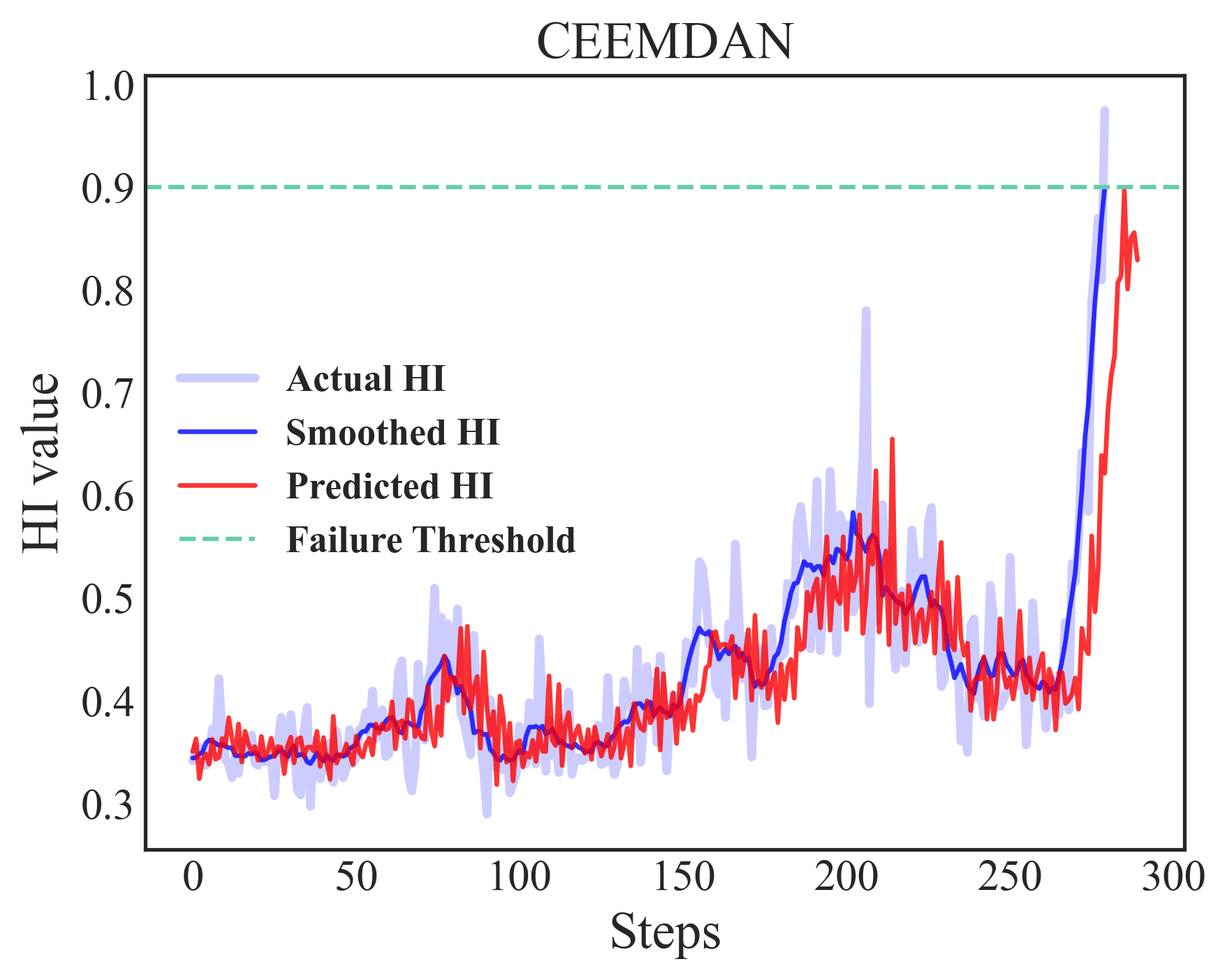}
\end{minipage}%
}%
\subfigure[]{
\begin{minipage}[t]{0.18\linewidth}
    \centering
    \includegraphics[width=1.2in]{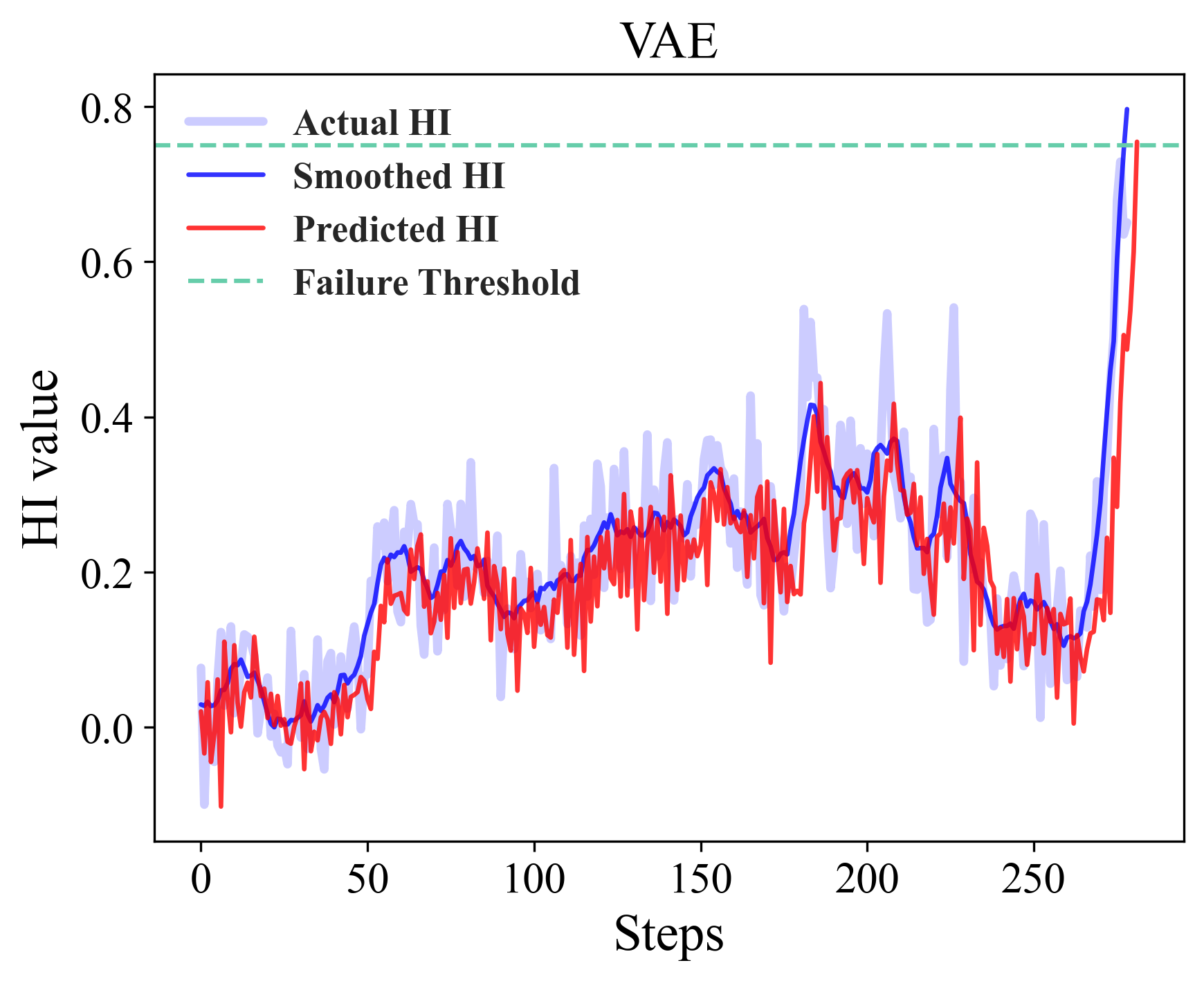}
\end{minipage}%
}
\subfigure[]{
\begin{minipage}[t]{0.18\linewidth}
    \centering
    \includegraphics[width=1.2in]{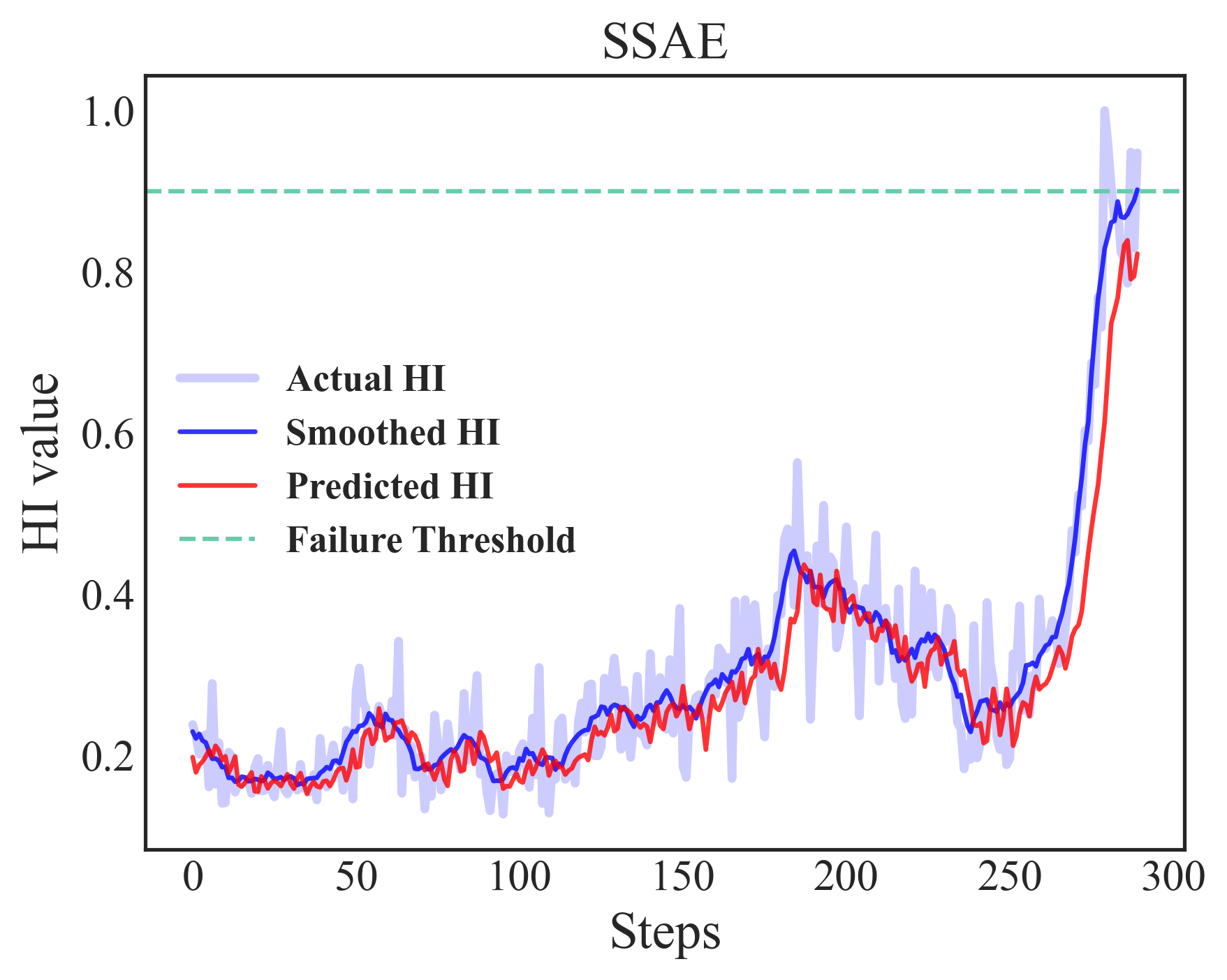}
\end{minipage}%
}
\subfigure[]{
\begin{minipage}[t]{0.18\linewidth}
    \centering
    \includegraphics[width=1.2in]{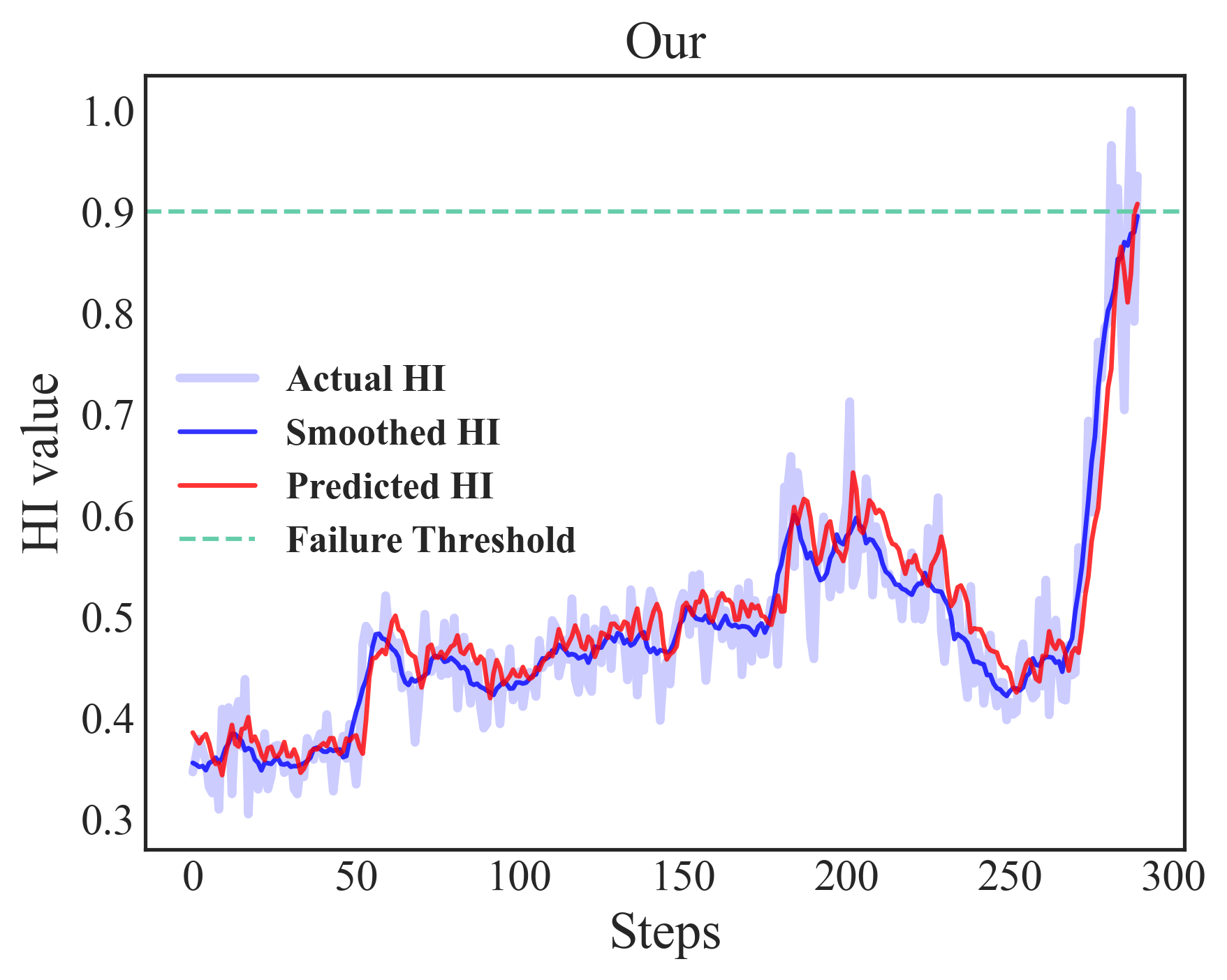}
\end{minipage}%
}
\centering
\vspace{-3mm}
\caption{\textcolor{black}{Prediction results obtained from different HIs for H-Bearing. (a) RMS, (b) P-Entropy, (c) ISOMAP, (d) KPCA, (e) CEEMDAN, (f) VAE, (g) SSAE, (h) Ours.}}
\label{pretask3}
\end{figure}

The prognostic results of comparison methods are presented in Table. \ref{table4} and their average performances are summarized in Fig. \ref{pre_avg}. The proposed dynamic HI achieves the best RMSE and Pred for all test bearings. For Pred, our method stands out as the only one with an average value exceeding 0.9, surpassing the second-highest value (0.889 achieved by VAE) by 0.041. This indicates a significant enhancement in the predictability of dynamic HI during its construction by considering the HI-level temporal dependence. The impact of HIs predictability on prognostics is clearly demonstrated in Fig. \ref{pretask1} to \ref{pretask3}. Across all testing bearings, only the trend of the proposed dynamic HI is well-predicted, demonstrating its superior capability in degradation prognostics. 
It can be noted that certain methodologies, despite achieving high Tred scores, such as KPCA and ISOMAP (shown in Table. \ref{table3}), fail to be predicted exactly. This indicates that only focusing on the trendability in construction may lead to unsatisfactory HI prediction results. 
Additionally, The RUL can be defined as the time interval between the first prediction time and the time when the predicted HI first exceeds the failure threshold \cite{guo2023novel,guo2024hybrid}. As the prediction result of Dynamic HI closely aligns with its smoothed trajectories, as presented in Fig. \ref{pretask1} (h), Fig. \ref{pretask2} (h), Fig. \ref{pretask3} (h), it achieves high RUL estimation accuracies. For instance, for P-Bearing1\_1, the actual failure step is 155 while the estimated step is 156, yielding a prediction accuracy of 99.35\%. However, as the predictability of other methods is insufficient, the HI prediction results are not satisfied, thereby leading to poor prognostic performance.

From the comparison, it can be concluded that the proposed HI construction method can significantly enhance the predictability of the constructed HI compared to other approaches. This improvement not only allows dynamic HI to better represent the degradation trend but also boosts its capability for degradation prognostics. Consequently, dynamic HI facilitates predictive maintenance decisions based on its superior prediction results.

\begin{table*}[!h]\centering
\vspace{-3mm}
\caption{Prediction Performance of Different HIs in the Degradation Trend Prediction of Bearing HI}
\setlength\tabcolsep{3.5pt}
\label{table4}
\scalebox{0.9}{
\begin{tabularx}{0.53\textwidth}{ccccccc}
  \hline
  \multirowcell{2}{Method} & \multicolumn{2}{c}{P-Bearing1\_1} & \multicolumn{2}{c}{P-Bearing2\_1} & \multicolumn{2}{c}{H-Bearing}\\
  \quad & RMSE & Pred & RMSE & Pred & RMSE & Pred \\
  \hline
  RMS & 0.099 & 0.853 & 0.133 & 0.852 & 0.131 & 0.809\\
  P-Entropy & 0.094 & 0.888 & 0.093 & 0.899 & 0.225 & 0.672\\
  ISOMAP & 0.084 & 0.892 & 0.121 & 0.829 & 0.156 & 0.737\\
  KPCA & 0.148 & 0.771 & 0.121 & 0.851 & 0.104 & 0.856\\
  CEEMDAN & 0.091 & 0.889 & 0.130 & 0.893 & 0.098 & 0.872\\
  VAE & 0.089 & 0.883 & 0.070 & 0.910 & 0.080 & 0.875\\
  SSAE & 0.077 & 0.906 & 0.144 & 0.849 & 0.066 & 0.908\\
  Ours & $\mathbf{0.032}$ & $\mathbf{0.956}$ & $\mathbf{0.058}$ & $\mathbf{0.916}$ & $\mathbf{0.051}$ & $\mathbf{0.917}$ \\
  \hline
\end{tabularx}}
\end{table*}

\subsubsection{Ablation Studies and Discussion}
\label{sec3.3.4}
To verify the effectiveness of the proposed modules and block on bearing degradation trend prediction, \textcolor{black}{ablation studies are conducted on HIs constructed by No SkipAE, No inner HI-prediction block, No SkipAE and HI-prediction block, No HI-generating module, and No SkipAE and HI-generating module, which have been illustrated in section \ref{sec3.2.4}. }

\begin{figure}[!h]
\centering
\subfigure[]{
\begin{minipage}[t]{0.28\linewidth}
    \centering
    \includegraphics[width=1.8in]{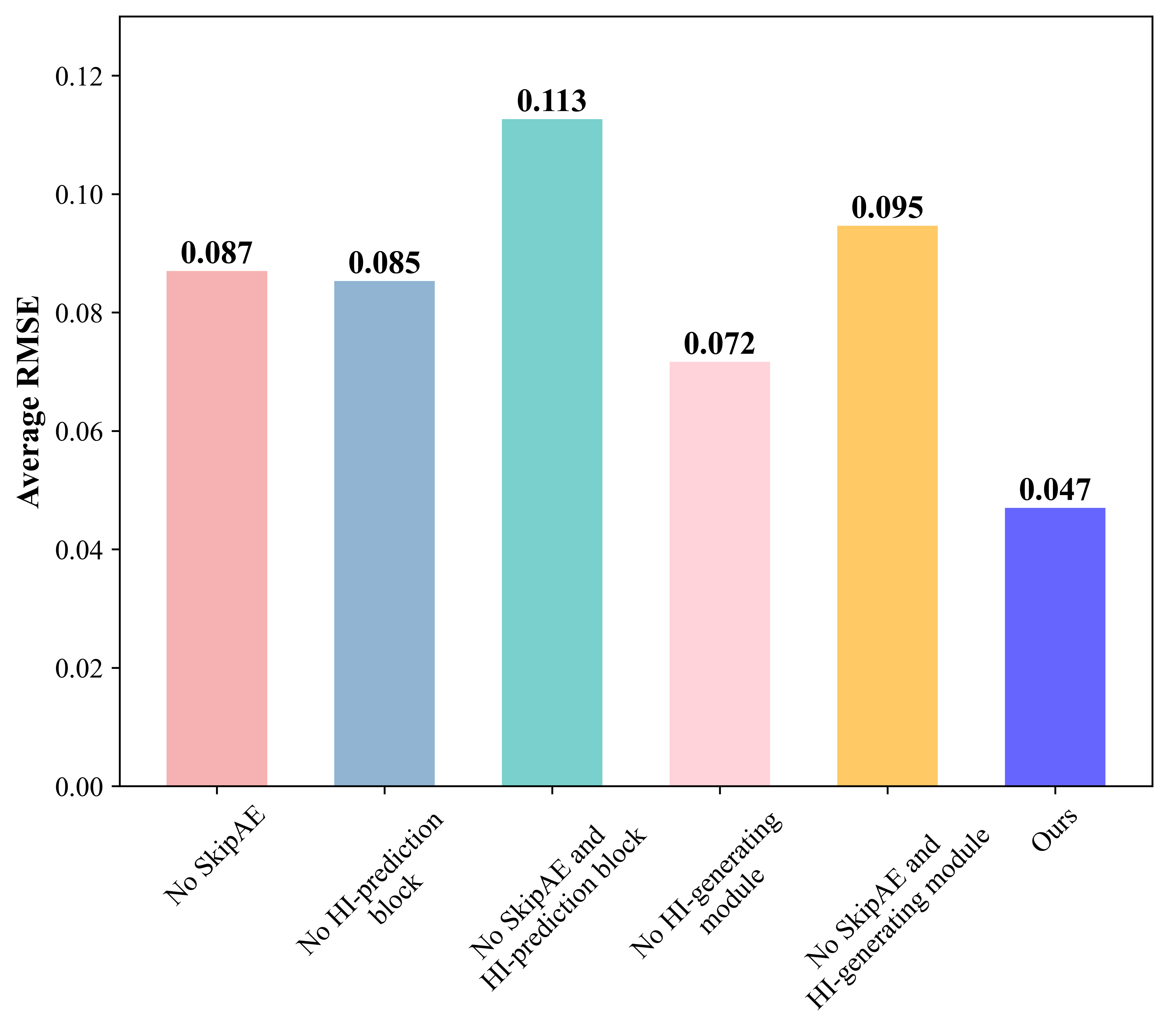}
\end{minipage}%
}
\subfigure[]{
\begin{minipage}[t]{0.28\linewidth}
    \centering
    \includegraphics[width=1.8in]{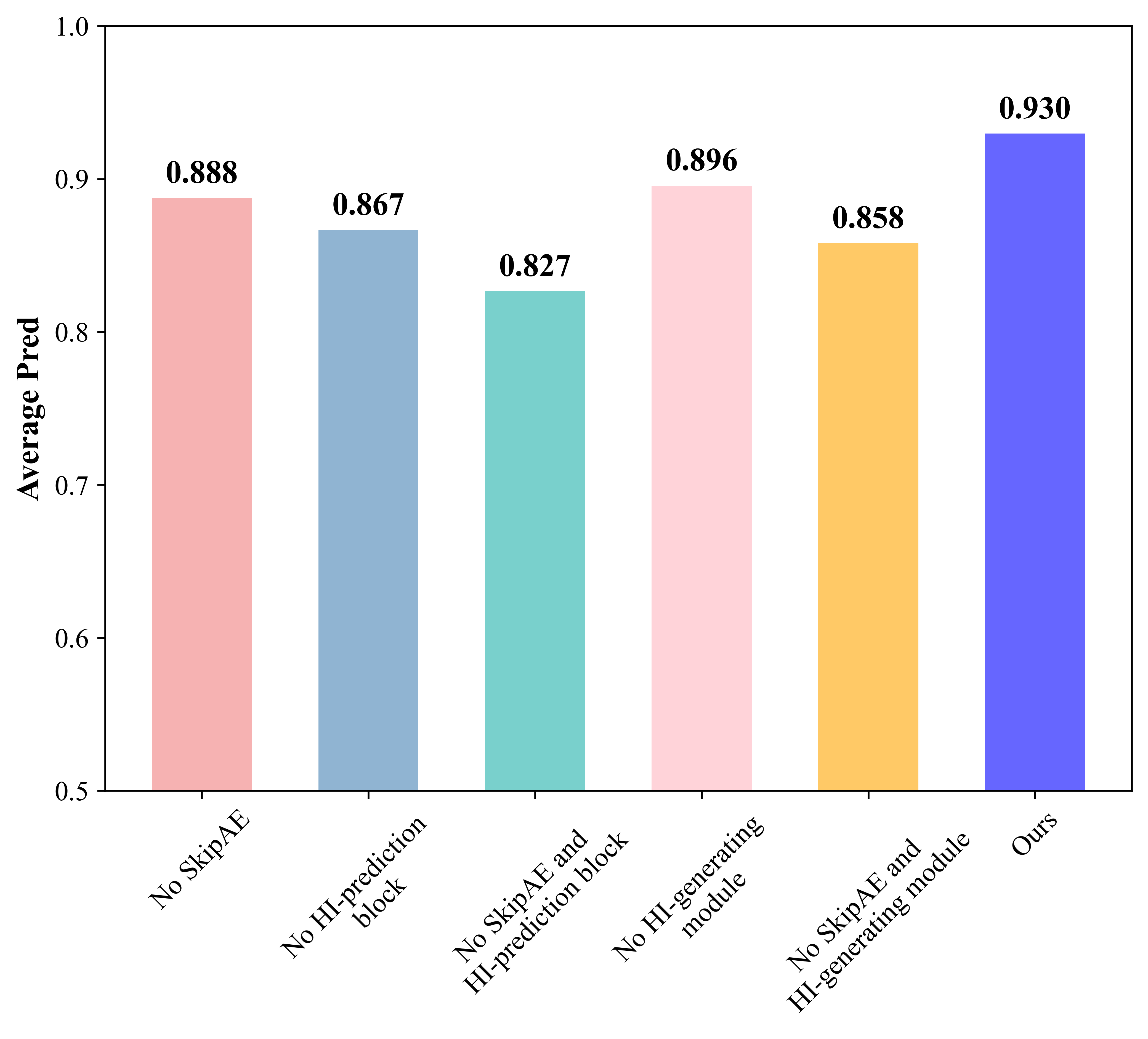}
\end{minipage}%
}
\centering
\vspace{-3mm}
\caption{\textcolor{black}{The average values of various metrics from ablation studies on degradation trend prediction across all tasks are presented. (a) RMSE, (b) Pred.}}
\label{avg_pre_ab}
\end{figure}

\textcolor{black}{Table. \ref{hi_pre_ablation} details the experimental results, showing that the removal of different components results in varying degrees of performance decline across all metrics. As shown in Fig. \ref{avg_pre_ab} (a), the most significant performance drop is observed with the No SkipAE and HI-prediction block, where the average RMSE increases from 0.047 to 0.113 and the average Pred decreases from 0.930 to 0.827. This indicates that skip connections and HI-level temporal dependence jointly enhance the ability of the HI in degradation prediction. Compared to methods that remove only SkipAE (No SkipAE) or the inner HI-prediction block (No HI-prediction block), the more significant performance degradation demonstrates that the proposed framework effectively integrates these two components to enhance the performance of HI in bearing prognostics.}

\begin{figure}[!t]
\centering
\subfigure[]{
\begin{minipage}[t]{0.22\linewidth}
    \centering
    \includegraphics[width=1.4in]{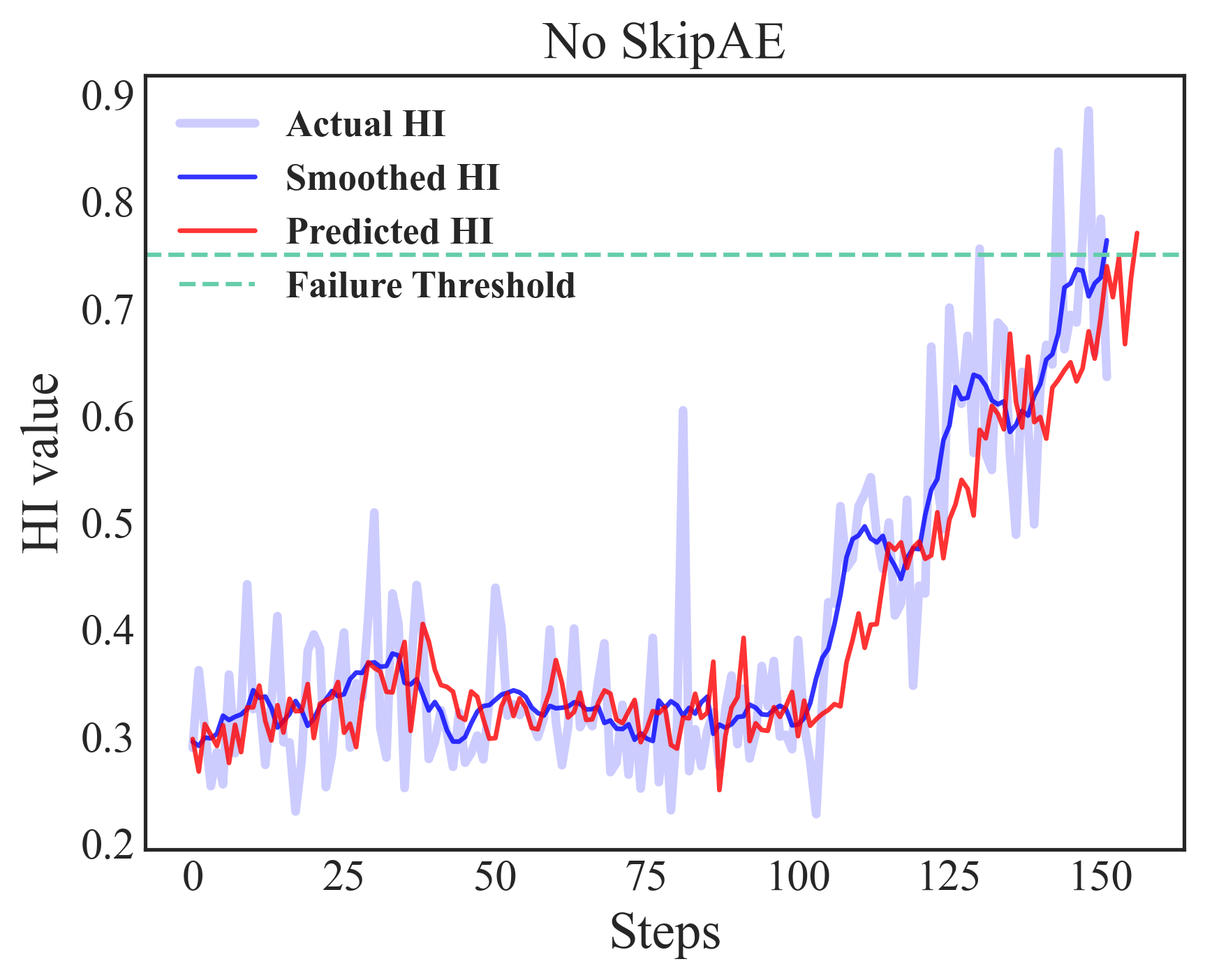}
\end{minipage}%
}
\subfigure[]{
\begin{minipage}[t]{0.22\linewidth}
    \centering
    \includegraphics[width=1.4in]{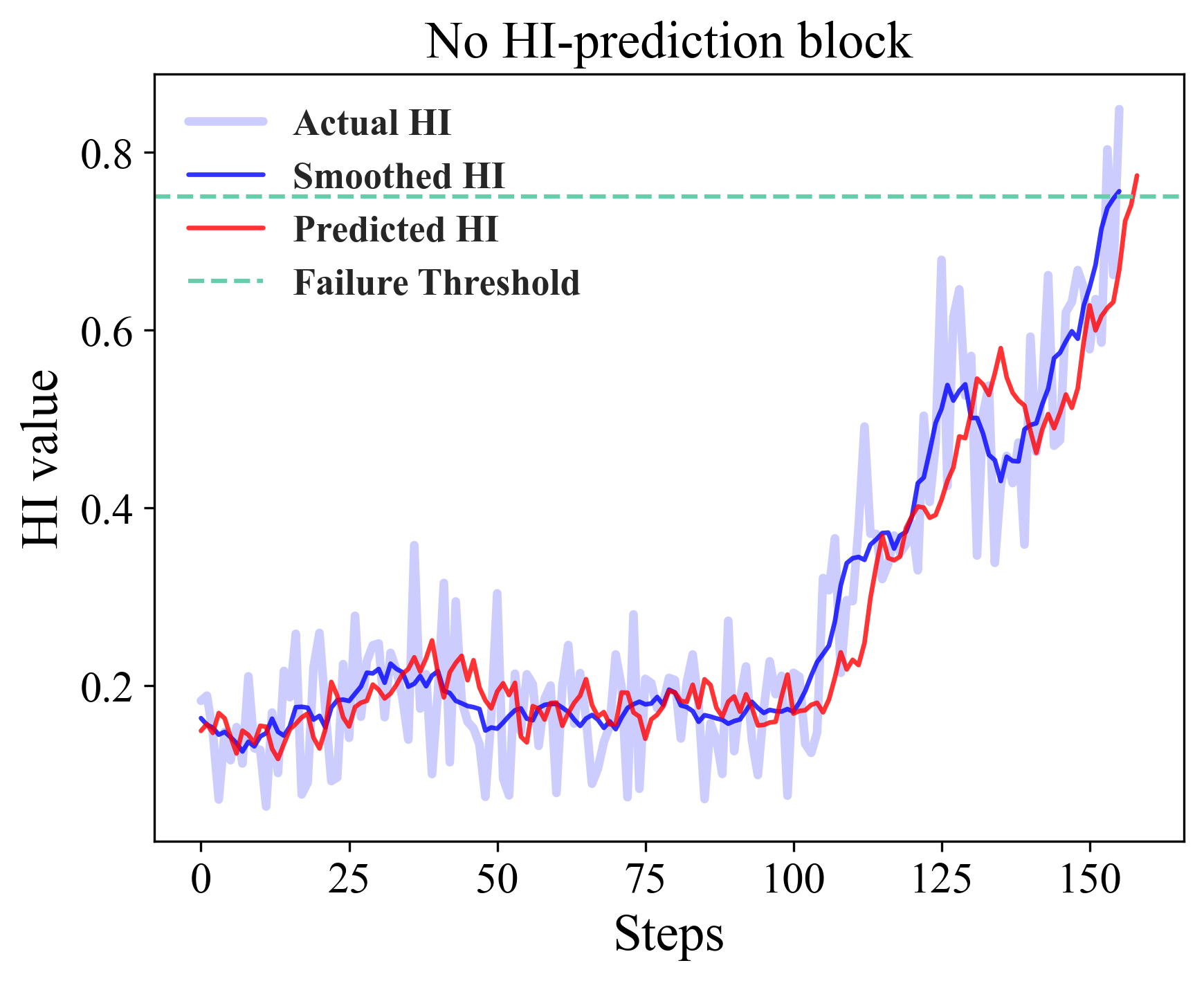}
\end{minipage}%
}
\subfigure[]{
\begin{minipage}[t]{0.22\linewidth}
    \centering
    \includegraphics[width=1.4in]{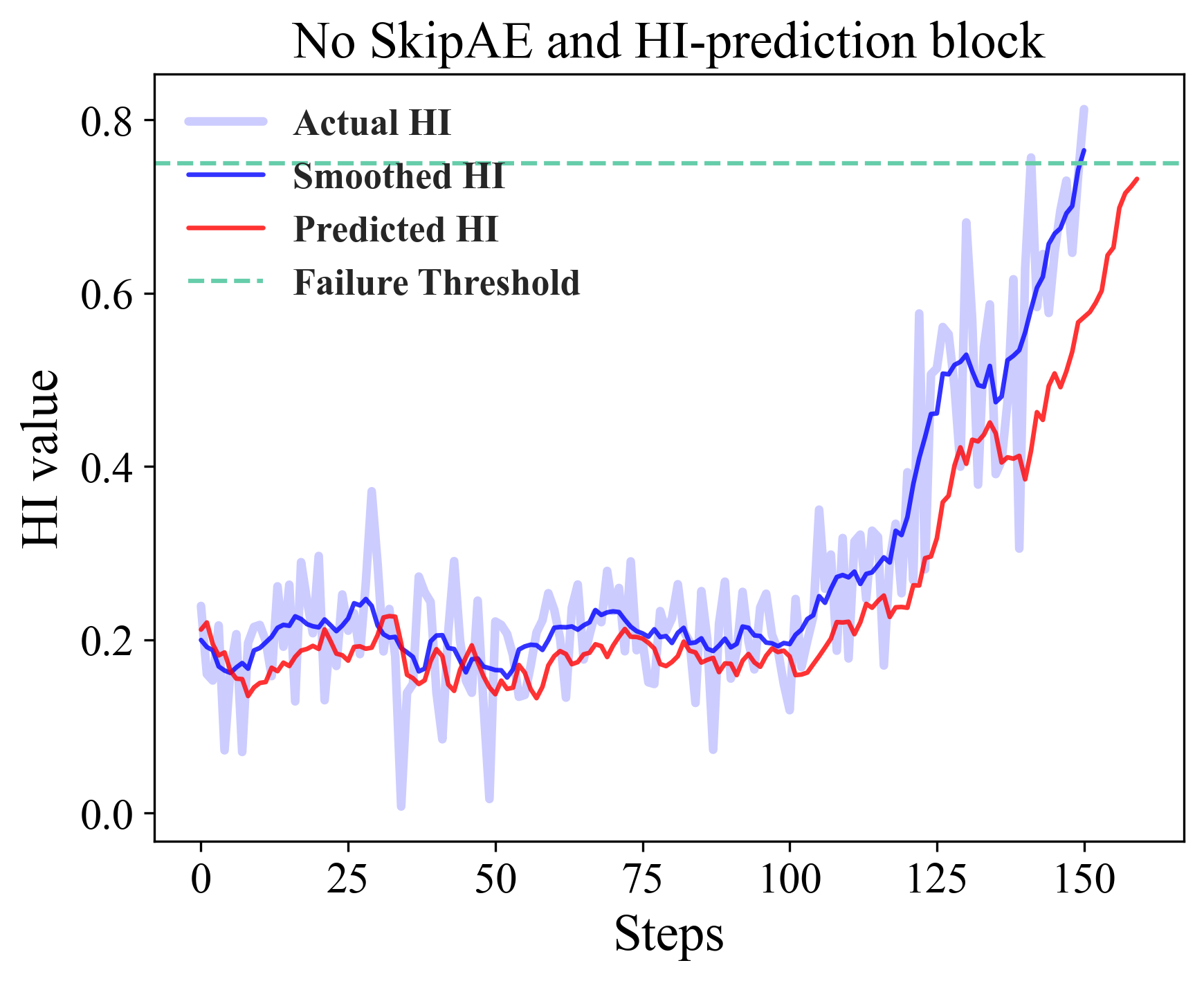}
\end{minipage}%
}\\
\vspace{-3mm}
\subfigure[]{
\begin{minipage}[t]{0.22\linewidth}
    \centering
    \includegraphics[width=1.4in]{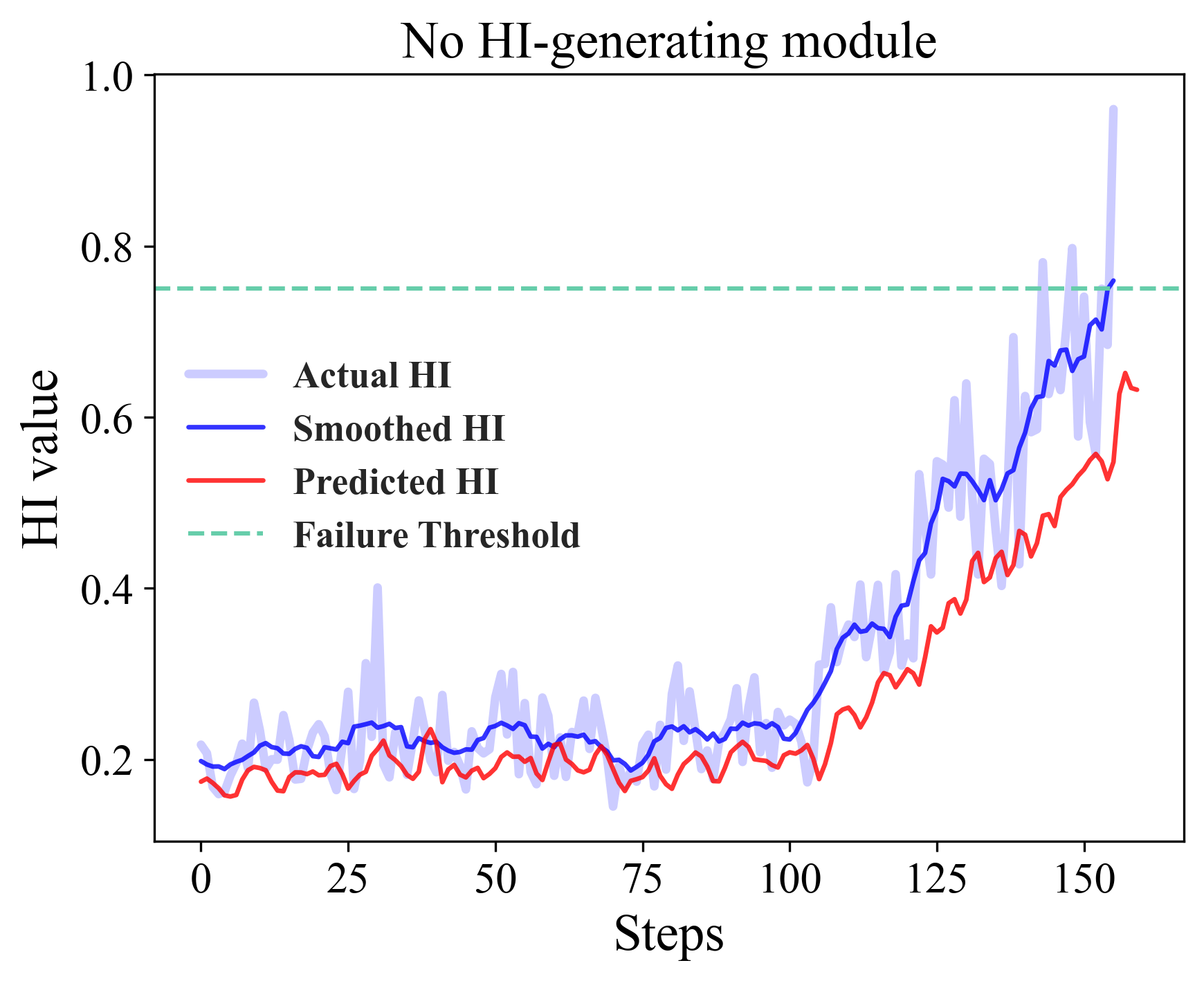}
\end{minipage}%
}
\subfigure[]{
\begin{minipage}[t]{0.22\linewidth}
    \centering
    \includegraphics[width=1.4in]{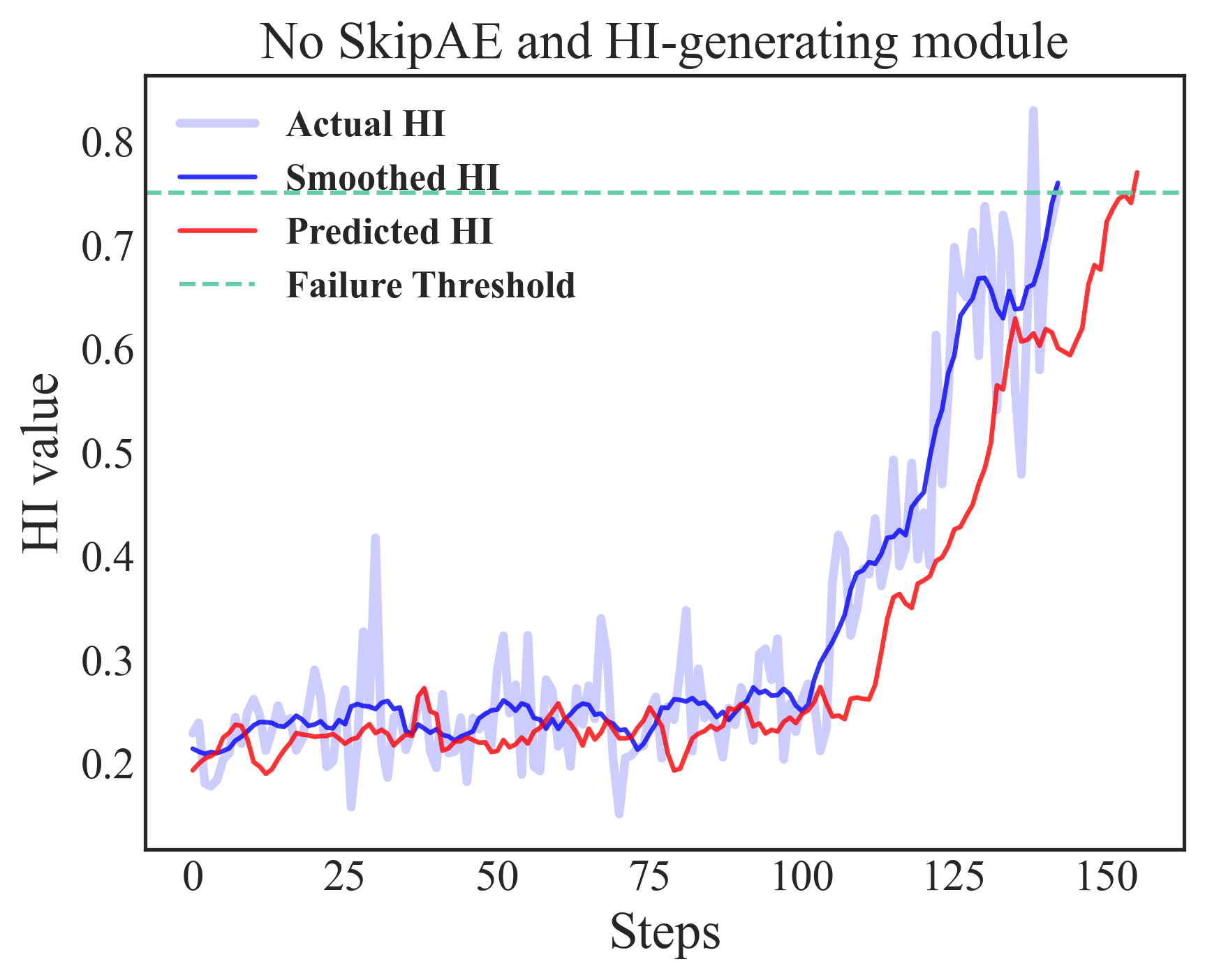}
\end{minipage}%
}
\subfigure[]{
\begin{minipage}[t]{0.22\linewidth}
    \centering
    \includegraphics[width=1.4in]{prebearing11_Our.png}
\end{minipage}%
}
\centering
\vspace{-3mm}
\caption{\textcolor{black}{Prediction results obtained from various ablation studies on the P-Bearing1\_1. (a) No SkipAE, (b) No inner HI-prediction block, (c) No SkipAE and HI-prediction block, (d) No HI-generating module, (e) No SkipAE and HI-generating module, (f) Ours.}}
\label{trend_prediction_ab}
\end{figure}

\textcolor{black}{The effectiveness of SkipAE is demonstrated through a comparison with the No SkipAE method. As illustrated in Table. \ref{hi_pre_ablation}, all metrics for No SkipAE show a decline across all research tasks compared to our method, indicateing that learning more representative degradation features enhances HI predictability. This effectiveness is visually demonstrated in Fig. \ref{trend_prediction_ab} (a) and (f), where our HI exhibits superior robustness, monotonicity, and predictability with skip connections.}

\textcolor{black}{
The effectiveness of the inner HI-prediction block is shown in Fig. \ref{avg_pre_ab} (b), where removing this block decreases the average predictability of HI by 6.77\% (from 0.930 to 0.867). Fig. \ref{trend_prediction_ab} (b) illustrates that this removal results in increased fluctuation of the HI, reducing its trendability and monotonicity, and ultimately impacting its predictability. These results underscore the importance of HI-temporal dependence in bearing prognostics and highlight the necessity of considering it during HI construction.
}

\textcolor{black}{Fig. \ref{avg_pre_ab} illustrates that removing the HI-generating module leads to an increase in average RMSE from 0.047 to 0.072 and a decrease in average Pred from 0.930 to 0.896, demonstrating the effectiveness of the HI-generating module. As the inner HI-prediction block is integrated into this module, the experiment results demonstrate the integrity of the proposed framework and further highlight the contribution of HI-level temporal dependence in degradation trend prediction.} 

\textcolor{black}{Additionally, the effectiveness of SkipAE and the dynamic HI-generating module is validated. As shown in Table. \ref{hi_pre_ablation}, removing these components (No SkipAE and HI-generating module) results in a decline across all metrics, indicating their joint contribution to HI prediction. Compared to removing only the HI-generating module (No HI-generating module), the average performance further declines, as shown in Fig. \ref{avg_pre_ab}, underscoring the effectiveness of SkipAE in bearing prognostics. }

Based on the ablation study, it can be concluded that \textcolor{black}{the proposed SkipAE, inner HI-prediction block, and the HI-generating module} jointly improve the ability of HI for reliable and accurate prognostics.

\begin{table*}[!t]\centering
\caption{\textcolor{black}{Results of the Ablation Study in the Bearing Degradation Trend Prediction.}}
\setlength\tabcolsep{3.5pt}
\label{hi_pre_ablation}
\scalebox{0.8}{
\begin{tabularx}{0.75\textwidth}{ccccccc}
  \hline
  \multirowcell{2}{Method} & \multicolumn{2}{c}{P-Bearing1\_1} & \multicolumn{2}{c}{P-Bearing2\_1} & \multicolumn{2}{c}{H-Bearing}\\
  \quad & RMSE & Pred & RMSE & Pred & RMSE & Pred \\
  \hline
  No SkipAE & 0.071 & 0.887 & 0.111 & 0.891 & 0.079 & 0.885\\
  \textcolor{black}{No HI-prediction block} & 0.083 & 0.863 & 0.088 & 0.884 & 0.085 & 0.853\\
  \textcolor{black}{No SkipAE and HI-prediction block} & \textcolor{black}{0.105} & \textcolor{black}{0.842} & \textcolor{black}{0.123} & \textcolor{black}{0.834} & \textcolor{black}{0.110} & \textcolor{black}{0.804} \\
  \textcolor{black}{No HI-generating module} & \textcolor{black}{0.078} & \textcolor{black}{0.882} & \textcolor{black}{0.072} & \textcolor{black}{0.901} & \textcolor{black}{0.065} & \textcolor{black}{0.904} \\
  \textcolor{black}{No SkipAE and HI-generating module} & \textcolor{black}{0.081} & \textcolor{black}{0.871} & \textcolor{black}{0.117} & \textcolor{black}{0.848} & \textcolor{black}{0.086} & \textcolor{black}{0.855} \\
  Ours & $\mathbf{0.032}$ & $\mathbf{0.956}$ & $\mathbf{0.058}$ & $\mathbf{0.916}$ & $\mathbf{0.051}$ & $\mathbf{0.917}$ \\
  \hline
\end{tabularx}}
\end{table*}

\section{Conclusions}
\label{sec:Conclusions}
In this paper, an unsupervised framework is proposed to construct dynamic HI, which is effective for both degradation trend characterization and prediction of rolling bearings. A SkipAE is first developed to learn representative degradation features from raw signals without expert knowledge. \textcolor{black}{By introducing skip connections, multi-level and multi-scale information is preserved in the learned features to facilitate the HI construction.} Subsequently, a dynamic HI-generating module is proposed to construct HI from the learned features, where an inner HI-prediction block is integrated to explicitly guarantee the HI-level temporal dependence during the HI construction. \textcolor{black}{The HI-level temporal dependence captures the dynamic information between past and future HI values, improving the representability and the predictability of the dynamic HI for bearing degradation prognostics.} The comparison experiments are conducted on IEEE PHM 2012 and HIT-B datasets. Compared to current popular machine learning and deep learning-based HI construction methods, the proposed approach constructs dynamic HI with superior trendability, monotonicity, robustness, and predictability. \textcolor{black}{Notably, degradation trend prediction tests with representative HIs reveal that the predictability of dynamic HIs is significantly improved, highlighting its promising potential in degradation prognostics for rolling bearings.}

\bibliography{myrefs.bib}

\end{document}